%% file: main.tex
\documentclass[acmtog,nonacm]{acmart}

\usepackage{booktabs} %
\usepackage{cleveref}
\usepackage{array}

\setcopyright{none}
\usepackage{tikz}
\usetikzlibrary{matrix,fit,backgrounds,positioning,calc}

\usepackage{arydshln}
\usepackage{multirow}

\usepackage[ruled]{algorithm2e} %

\SetAlFnt{\small}
\SetAlCapFnt{\small}
\SetAlCapNameFnt{\small}
\SetAlCapHSkip{0pt}

\makeatletter
\let\@authorsaddresses\@empty
\makeatother

\renewcommand\footnotetextcopyrightpermission[1]{}

\def\eg{\emph{e.g.}}

\begin{document}

\title{IP-Composer: Semantic Composition of Visual Concepts}

\author{Sara Dorfman}
\affiliation{
\institution{Tel Aviv University}
\country{Israel}
}
\author{Dana Cohen-Bar}
\affiliation{
\institution{Tel Aviv University}
\country{Israel}
}
\author{Rinon Gal}
\affiliation{
\institution{NVIDIA}
\country{Israel}
}
\author{Daniel Cohen-Or}
\affiliation{
\institution{Tel Aviv University}
\country{Israel}
}

\input{0_abstract}

\begin{teaserfigure}
    \centering
    \input{figures/teaser}
    \label{fig:teaser}
\end{teaserfigure}

\maketitle

\input{1_intro}

\input{2_related}
\input{3_method}
\input{4_experiments}

\input{5_conclusions}

\bibliographystyle{ACM-Reference-Format}
\bibliography{main}

\input{figures/additional_qualitative_results_1}
\input{figures/additional_qualitative_results_2}

\end{document}

%% file: 0_abstract.tex
\begin{abstract}

    Content creators often draw inspiration from multiple visual sources, combining distinct elements to craft new compositions. Modern computational approaches now aim to emulate this fundamental creative process. Although recent diffusion models excel at text-guided compositional synthesis, text as a medium often lacks precise control over visual details. Image-based composition approaches can capture more nuanced features, but existing methods are typically limited in the range of concepts they can capture, and require expensive training procedures or specialized data. We present IP-Composer, a novel training-free approach for compositional image generation that leverages multiple image references simultaneously, while using natural language to describe the concept to be extracted from each image. 
    Our method builds on IP-Adapter, which synthesizes novel images conditioned on an input image's CLIP embedding. We extend this approach to multiple visual inputs by crafting composite embeddings, stitched from the projections of multiple input images onto concept-specific CLIP-subspaces identified through text. Through comprehensive evaluation, we show that our approach enables more precise control over a larger range of visual concept compositions.

\end{abstract}

%% file: figures/teaser.tex
\centering

\includegraphics[width=0.85\textwidth]{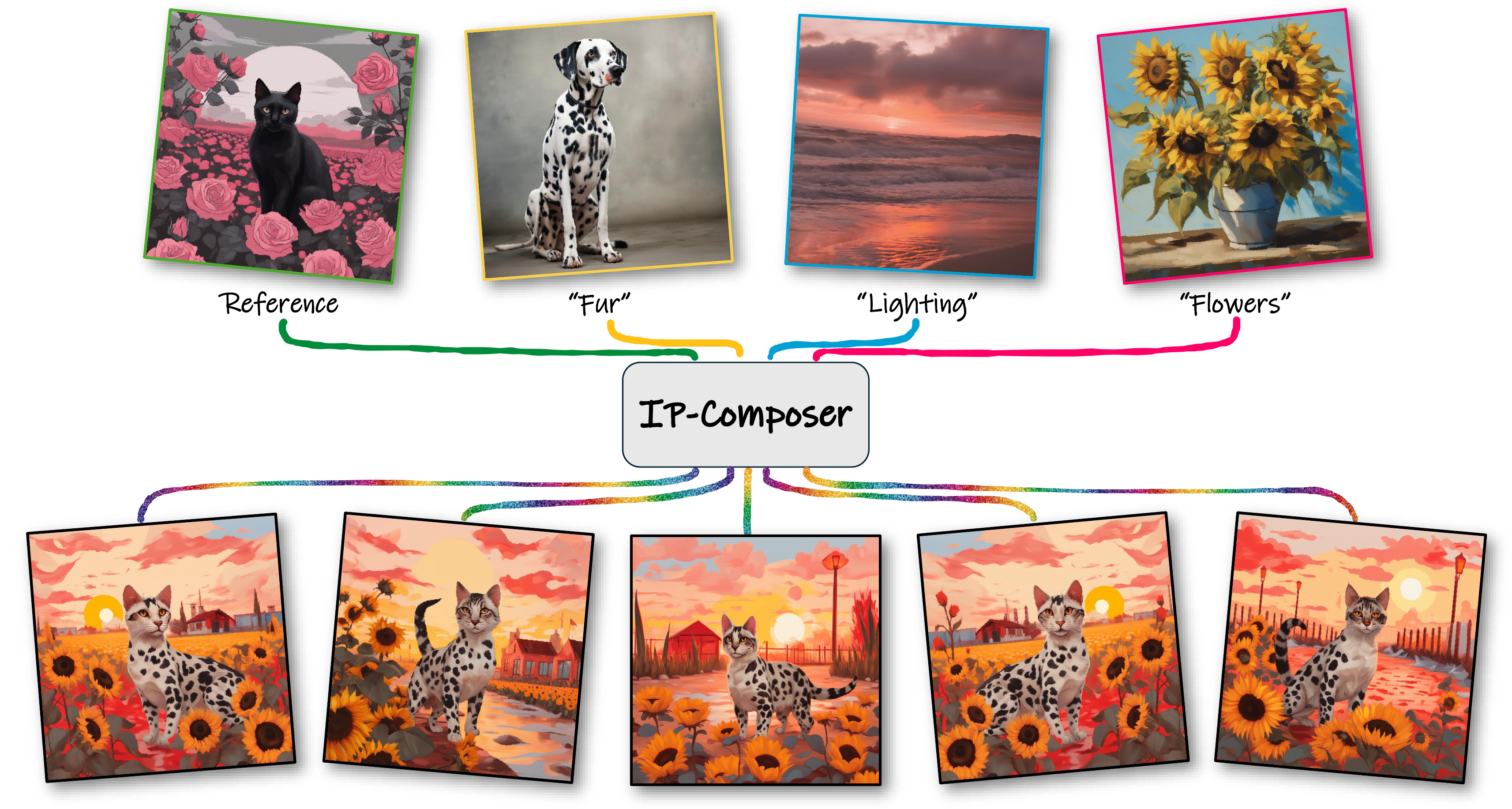}
\caption{
IP-Composer enables compositional generation from a set of visual concepts. These are portrayed through a set of input images, along with a prompt describing the desired concept to be extracted from each.
}

%% file: 1_intro.tex
\vspace{-10pt}
\section{Introduction}
The ability to fuse visual concepts from different sources into a single cohesive composition is a fundamental aspect of developing novel, creative content. This approach mirrors our natural creative processes: selecting specific attributes, objects, and elements from various inspirations to craft something new and unique. 

Extensive research has been conducted to enable such compositional image generation capabilities. The common approach relies on the unprecedented ability of recent diffusion models to synthesize images conditioned on natural language prompts. Since language is inherently composable, one can easily combine unrelated concepts through simple prompts. However, text-based methods often lack the detailed control and precision that is frequently required for more fine-grained applications~\cite{Zhang_2023_ICCV,gal2022textual,ruiz2022dreambooth}. To address these shortcomings, a more recent line of work focuses on manipulation and composition techniques based on image references~\cite{richardson2024popsphotoinspireddiffusionoperators,lee2024languageinformedvisualconceptlearning,zhang2023prospectpromptspectrumattributeaware,vinker2023concept}. As images are inherently more expressive and precise, these techniques are able to capture complex visual details that textual descriptions often fail to convey. Although powerful, these methods are frequently constrained by a limited range of concepts they can handle, or require computationally expensive per-concept training and fine-tuning that reduce their practicality and scalability. 

Our work, IP-Composer, aims to address these limitations by introducing a highly flexible, training-free approach for compositional image generation that combines several concepts drawn from multiple visual sources.
We build on IP-Adapter, an encoder-based approach that augments an existing text-to-image diffusion model (e.g., SDXL) with a new image-condition input, allowing users to generate novel variations of the content shown in the conditioning image. Importantly, IP-Adapter employs CLIP as a feature extraction backbone. 

Recently, it has been shown~\citep{gandelsman2024interpretingclipsimagerepresentation} that CLIP's attention heads span semantic subspaces of the CLIP embedding space. These subspaces are then characterized by finding textual descriptions whose CLIP embeddings span the same space. In their work, \citet{gandelsman2024interpretingclipsimagerepresentation} demonstrate that this property can be used to improve the accuracy of CLIP-based classification, by simply subtracting projections to subspaces linked to background properties such as ``snow" or ``water", which lead to spurious correlations. 
Our hypothesis is that if it is possible to remove a concept (``water background") without harming the semantics of the rest of the embedding, then it may also be possible to replace it with a different instance from the same concept category, drawn from a different image.

To achieve this, we propose to first identify the CLIP-subspaces that are tied to the textually described concepts that we want to extract from each conditioning image. This is done by asking an LLM to generate a list of texts describing possible variations of each concept, encoding them into CLIP-space, and finding the subspace which they span. Then, we encode each image and project its CLIP-embedding onto its relevant concept subspace, extracting an isolated concept embedding. These extracted concept representations can then be recombined to create new composite embeddings that preserve the semantic meaning of each component. Finally, this computed embedding replaces the standard image encoding in the IP-Adapter pipeline, enabling the synthesis of novel images containing the composition of concepts (see \cref{fig:teaser}). 

Notably, our approach bridges the gap between high-level conceptual control and fine-grained visual detail, using text to describe and select broad concepts, but specifying the unique instance of each concept through visual examples.

We compare our method with prior training-based approaches, and demonstrate that it not only allows for more general concept selection, but also competes favorably even in scenarios where training data is available. Compared to existing CLIP-based methods that rely on embeddings interpolation or concatenation, our approach achieves higher accuracy and robustness, enabling better control over a broader range of concepts with minimal attribute leakage.

All in all, our approach offers an intuitive, text-based and training-free method to generate images inspired by multiple visual concepts, opening new possibilities for creative content generation and visual exploration.

%% file: 2_related.tex
\section{Related work}
\paragraph{\textbf{Controllable diffusion models}}
Text-to-image diffusion models \citep{nichol2021glide, balaji2022ediffi,rombach2021highresolution, 
ramesh2022hierarchical,
saharia2022photorealistic,ho2020denoising} have emerged as a powerful paradigm for high-quality image generation, demonstrating remarkable capabilities in translating natural language descriptions into detailed images. As the technology matured, researchers explored various control mechanisms beyond text, including spatial controls such as segmentation masks \citep{couairon2022diffedit}, sketches \citep{voynov2023sketch}, depth maps \citep{zhang2023adding, mou2024t2i, bhat2023loosecontrolliftingcontrolnetgeneralized} and layout \citep{dahary2024yourselfboundedattentionmultisubject, avrahami2023spatext,zheng2023layoutdiffusion, li2023gligenopensetgroundedtexttoimage}. While text and spatial controls offer structural guidance, they often fall short in precisely controlling style and appearance. 

This limitation motivated the development of image-guided generation methods. 
One approach involves personalization through per-image optimization, either by fine-tuning token embeddings \citep{gal2022textual} or the model itself \citep{ruiz2022dreambooth}. More efficient encoder-based approaches~\citep{gal2023encoder,arar2023domain,ruiz2023hyperdreambooth,Wei_2023_ICCV,mou2024t2i,ye2023ipadapter} have also emerged. Of these, IP-Adapter \citep{ye2023ipadapter} employs a decoupled cross-attention mechanism to inject image features into the generation process. Our approach leverages a pre-trained IP-Adapter model to similarly inject image features into the generative process. However, we extend it to handle compositional generation, where multiple input images are used to describe an array of visual concepts that should appear in the generated outputs.

\paragraph{\textbf{CLIP directions for image editing}}
The discovery of semantically meaningful directions in latent spaces was first demonstrated in GANs \citep{goodfellow2014generative, karras2019style} where moving along these trajectories enables controlled image editing operations~\citep{shen2020interpreting}. Early unsupervised methods discovered these editing directions through various approaches: \citep{voynov2020unsupervised} learned directions by predicting identifiable image transformations, GANSpace \citep{harkonen2020ganspace} employed PCA to find dominant directions in latent codes, and SeFa \citep{shen2021closedformfactorizationlatentsemantics} analyzed generator weights directly to identify principal editing directions. 

The emergence of CLIP \citep{radford2021learning}, bridging visual and textual representations in a shared embedding space, revolutionized image editing by enabling text-guided manipulation. StyleClip \citep{patashnik2021styleclip} leveraged this capability by finding traversal directions that align images with textual descriptions. \citet{abdal2021clip2stylegan} proposed methods for discovering interpretable editing directions in CLIP space with automatic natural language descriptions. StyleGAN-NADA \citep{gal2021stylegannadaclipguideddomainadaptation} took a different approach, using CLIP-space directions to enable zero-shot domain adaptation. 
Recent works \citep{baumann2024continuoussubjectspecificattributecontrol, zhuang2024magnetknowtexttoimagediffusion} have demonstrated image editing capabilities in Stable Diffusion by manipulating CLIP text embedding.
Lastly, \citet{Guerrero_Viu_2024} leveraged domain diffusion prior~\citep{ramesh2022hierarchical, aggarwal2023controlledconditionaltextimage} to create clusters of image embeddings for source and target prompts, enabling the discovery of disentangled directions specifically for texture image editing.

Our approach also explores CLIP's embedding space. However, rather than finding directions of movement in CLIP space, we identify subspaces which encode specific semantic concepts. We then stitch new embeddings from the projections of different images on different concept spaces.

\paragraph{\textbf{Compositional image generation}}
Recent works have explored various approaches to enable multi-condition control in image generation. For text-based control, \citep{liu2023compositionalvisualgenerationcomposable} improved multi-object generation by introducing methods to compose multiple text prompts coherently. For spatial and global controls, Composer \citep{huang2023composercreativecontrollableimage} trained a diffusion model that accepts multiple conditions at test-time, while Uni-ControlNet \citep{zhao2023unicontrolnetallinonecontroltexttoimage} achieved similar capabilities with significantly reduced training costs by training two small adapters. Several works have focused on compositing elements from different images: ProSpect \citep{zhang2023prospectpromptspectrumattributeaware} introduced a step-aware prompt space to learn decomposed attributes from images for new compositions, while \citet{lee2024languageinformedvisualconceptlearning} proposed learning disentangled concept encoders aligned with language-specified axes, enabling composition through concept remixing. Finally, pOps~\citep{richardson2024popsphotoinspireddiffusionoperators} tunes a diffusion prior model~\citep{ramesh2022hierarchical} to learn semantic operators for element composition, though it requires training each operator on a suitable dataset, limiting its practical applications.

In contrast, our method enables compositional image generation using an off-the-shelf IP-Adapter model. Our approach leverages the ease of language-based controls to identify concept-specific CLIP subspaces, but uses image inputs to convey more specific details.

%% file: 3_method.tex
\section{Method}
We begin by describing our method for the simple case of creating a composition of two images.
Given a reference image \( I_{\text{ref}} \) (typically one describing the background or scene layout) and a concept image \( I_c \), we would like to output a composition depicting the concept \( c \) from \( I_c \) while obtaining the rest of the attributes from \( I_{\text{ref}} \). 

At the core of our method lies the ability to isolate and extract the ``$c$ component'' from a CLIP image embedding. Motivated by recent findings on the existence of different semantic subspaces in CLIP, we aim to construct a projection matrix \( P_c \), which will be used to project CLIP image embeddings to obtain the encoding of the specific concept ``$c$''.

\paragraph{\textbf{Constructing The Projection Matrix}} To construct a projection matrix for a concept $c$, we first gather a set of texts $t_1, \dots, t_n$, each describing an instance of the concept, with the aim of conceptually spanning its domain. To do so, we query a large language model (LLM) and simply ask it to create texts that span the concept's attribute space.
Next, using the CLIP text encoder \( CLIP_t \), we obtain embeddings for the collected texts: \( CLIP_t(t_1), \dots, CLIP_t(t_n) \). To extract the most relevant directions of this subspace, we apply Singular Value Decomposition (SVD) to the matrix of text embeddings. Let the combined embedding matrix be represented as:
\begin{equation}
E = \left[CLIP_t(t_1), \dots, CLIP_t(t_n)\right]^T,
\end{equation}
where \( E \in \mathbb{R}^{n \times d} \), \( n \) is the number of texts and \( d \) the embedding dimension. The SVD of \( E \) can be expressed as:
\begin{equation}
E = U \Sigma V^T,
\end{equation}
where \( U \in \mathbb{R}^{n \times n} \), \( \Sigma \in \mathbb{R}^{n \times d} \), and \( V \in \mathbb{R}^{d \times d} \). The rows of \( V \), also referred to as the right singular vectors, represent directions in the embedding space. While it is a common practice to normalize the embeddings before constructing the matrix \( E \), we observe improved performance when working with the unnormalized embeddings which also preserve the natural variation in the data.

Finally, we select the top \( r \) singular vectors (corresponding to the \( r \) largest singular values) from \( V \). These vectors capture the most significant variations in the subspace corresponding to concept \( c \). The projection matrix \( P_c \in \mathbb{R}^{d \times d} \) is then computed as:
\begin{equation}
P_c = V_r^T V_r,
\end{equation}
where \( V_r \in \mathbb{R}^{r \times d} \) contains the top \( r \) singular vectors. The value of \( r \) is selected empirically, and depends on the nature of the concept. In practice, the same \( r \) can often be used for most concepts, but broader concepts like ``animals" commonly benefit from utilizing more directions than specific concepts like ``dog breeds".

\input{figures/overview}

\paragraph{\textbf{Image Composition}} We aim to create a composite embedding that jointly encodes the concept \( c \) from \( I_c \) while preserving the remaining attributes of \( I_{\text{ref}} \). To achieve this, we simply replace the concept-space projection of \( I_{\text{ref}} \) with the projection of \( I_c \). More concretely, the composite embedding is given by:
\begin{equation}
\mathbf{e}_{\text{comp}} = \mathbf{e}_{\text{ref}} - P_c \mathbf{e}_{\text{ref}} + P_c \mathbf{e}_c,
\end{equation}
where \( \mathbf{e}_{\text{ref}} \) and \( \mathbf{e}_c \) are the CLIP embeddings of \( I_{\text{ref}} \) and \( I_c \), respectively. This composite embedding \( \mathbf{e}_{\text{comp}} \) is then passed to the IP-Adapter to generate the final composed image \( I_{\text{comp}} \), combining the attributes of \( I_{\text{ref}} \) with the concept instance extracted from \( I_c \).

\paragraph{\textbf{Generalization to Multiple Concepts}} The same approach can be extended to \( K \) concepts, \( \{c_1, c_2, \dots, c_K\} \), with corresponding projection matrices \( \{P_{c_1}, P_{c_2}, \dots, P_{c_K}\} \) and concept images \( \{I_{c_1}, I_{c_2}, \dots, I_{c_K}\} \). Here, the composed embedding is constructed by subtracting the concept-space projections of the reference embedding and adding the matching concept embedding from each source image:
\begin{equation}
\mathbf{e}_{\text{comp}} = \mathbf{e}_{\text{ref}} - \sum_{k=1}^K P_{c_k} \mathbf{e}_{\text{ref}} + \sum_{k=1}^K P_{c_k} \mathbf{e}_{c_k},
\end{equation}

where \( \mathbf{e}_{c_k} \) represents the embedding of the concept image \( I_{c_k} \). Note that we do not subtract the projection of each concept on the subspaces of the other concepts as we find empirically that this makes the compositions more sensitive to the choice of the number of singular vectors $r_c$ used for each concept.

\input{figures/qualitative_results}

\paragraph{\textbf{Implementation Details}} We implement our method on top of a pre-trained SDXL~\cite{podell2024sdxl} model using an IP-Adapter~\cite{ye2023ipadapter} encoder based on OpenCLIP-ViT-H-14~\cite{ilharco_gabriel_2021_5143773} (ip-adapter\_sdxl\_vit-h). To generate concept variation descriptions, we used GPT-4o~\cite{OpenAI2022ChatGPT} and asked for $150$ prompts. In cases that require higher subspace dimensions (\eg object insertion) we instead generated $500$ prompts.

%% file: figures/overview.tex
\begin{figure}
    \centering
    \setlength{\belowcaptionskip}{-5pt}
    \setlength{\abovecaptionskip}{4pt}
    
    \includegraphics[width=1.0\linewidth]{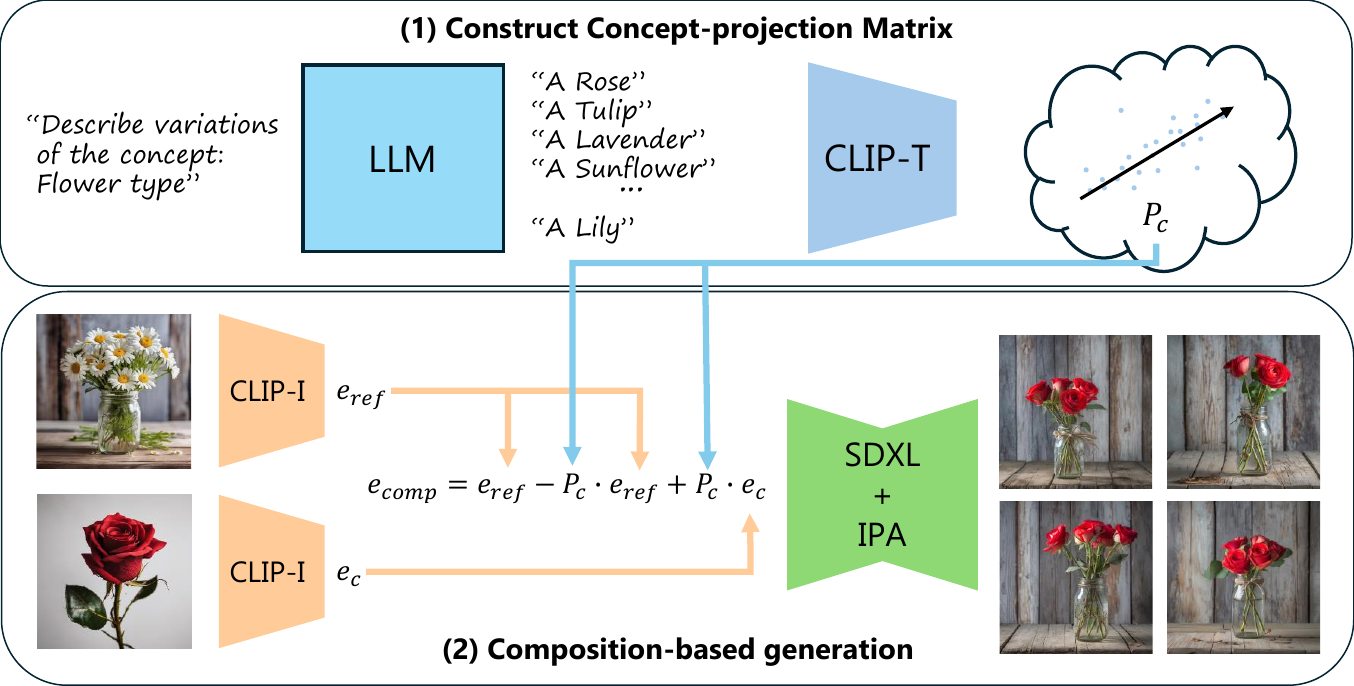}
    \caption{
    Method overview for a 2-image composition scenario. (top) We use an LLM to generate texts describing possible variations of a concept we want to extract from the concept-image. We encode the responses using CLIP, and find the embedding-subspace that they span.
    (bottom) We generate a composite CLIP-embedding by replacing the projection of the reference image on this embedding-subspace with the matching projection of the concept-image. The composite embedding can be used by an off-the-shelf IP-Adapter to generate images combining the reference and the visual concept. The same approach can be applied with additional concept images.}
\label{fig:method}
\end{figure}

%% file: figures/qualitative_results.tex
\begin{figure*}[!htb]
    \centering
    
    \begin{tikzpicture}
        \matrix (m1) [matrix of nodes,
            nodes={draw, minimum width=2cm, minimum height=1cm, inner sep=0pt, line width=1.5pt},
            row sep=0.45cm,
            column sep=0.443cm
        ] {
            \includegraphics[width=2.2cm,height=2.2cm]{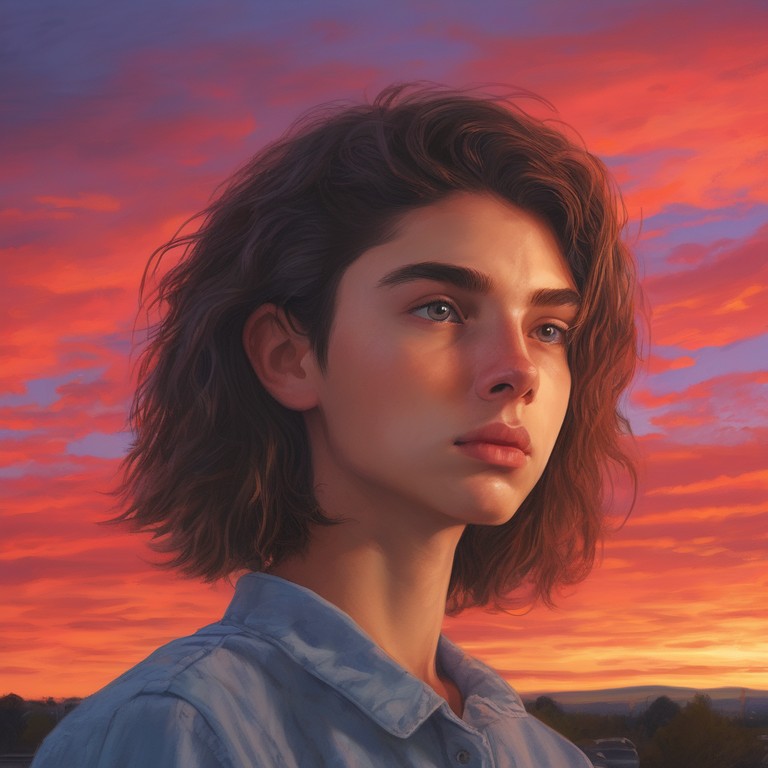} & \includegraphics[width=2.2cm,height=2.2cm]{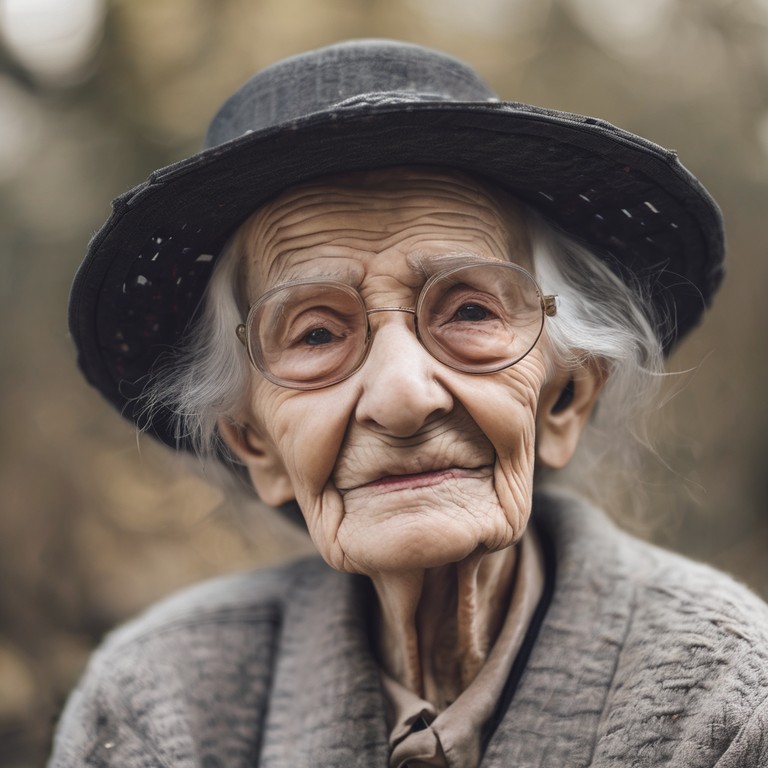} & \includegraphics[width=2.2cm,height=2.2cm]{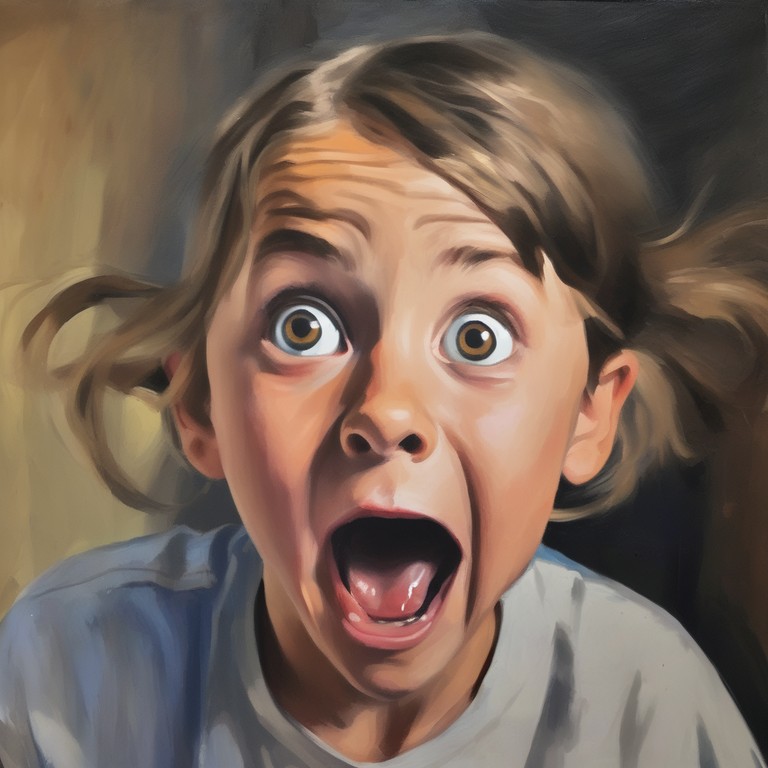} & \includegraphics[width=2.2cm,height=2.2cm]{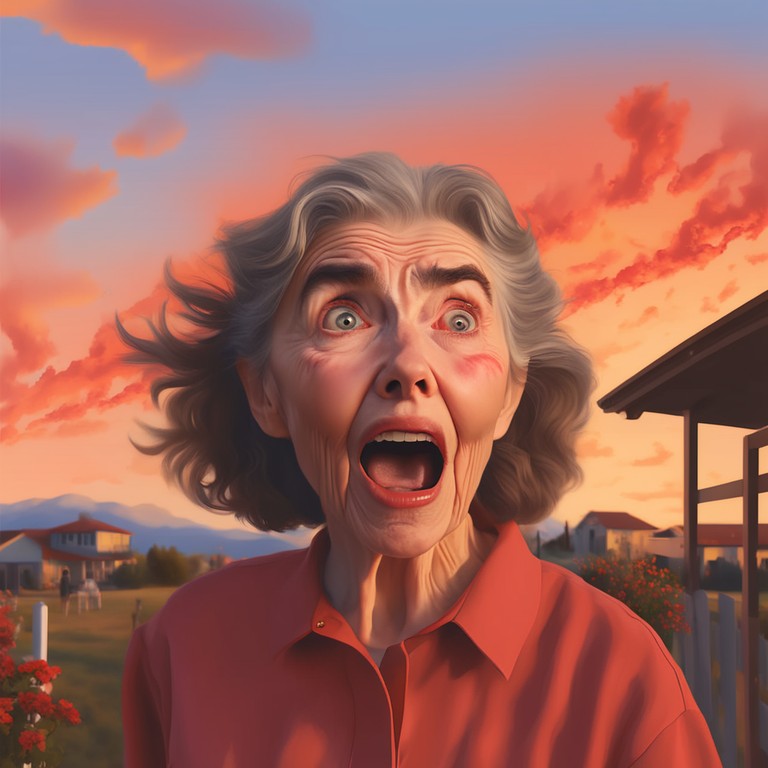} & \includegraphics[width=2.2cm,height=2.2cm]{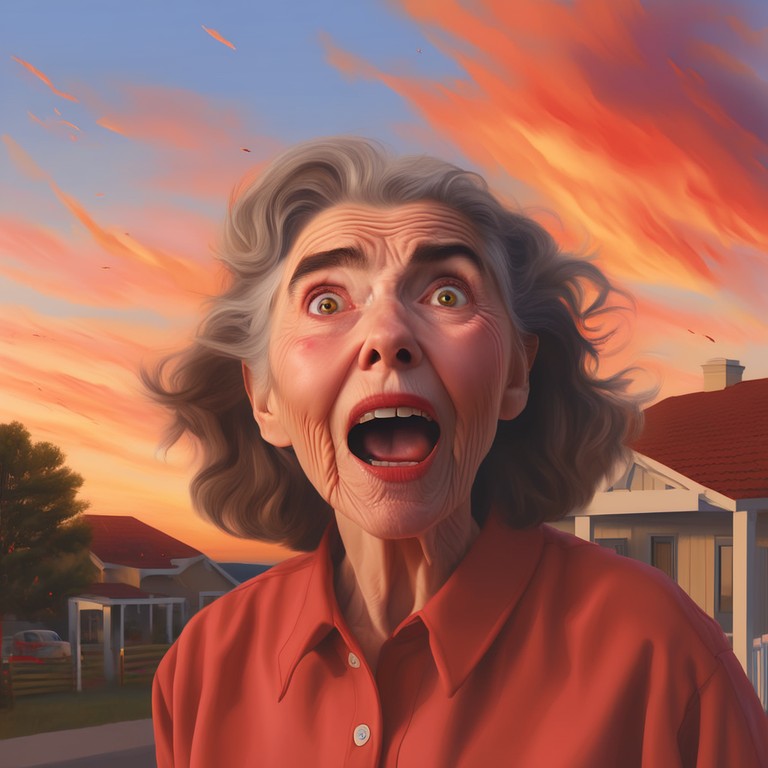} & \includegraphics[width=2.2cm,height=2.2cm]{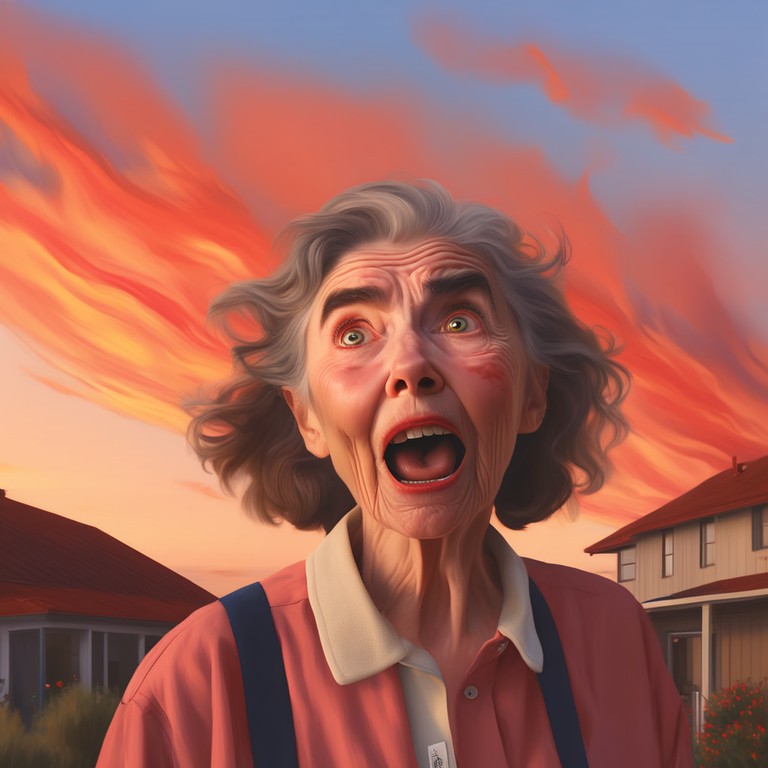} \\
            \includegraphics[width=2.2cm,height=2.2cm]{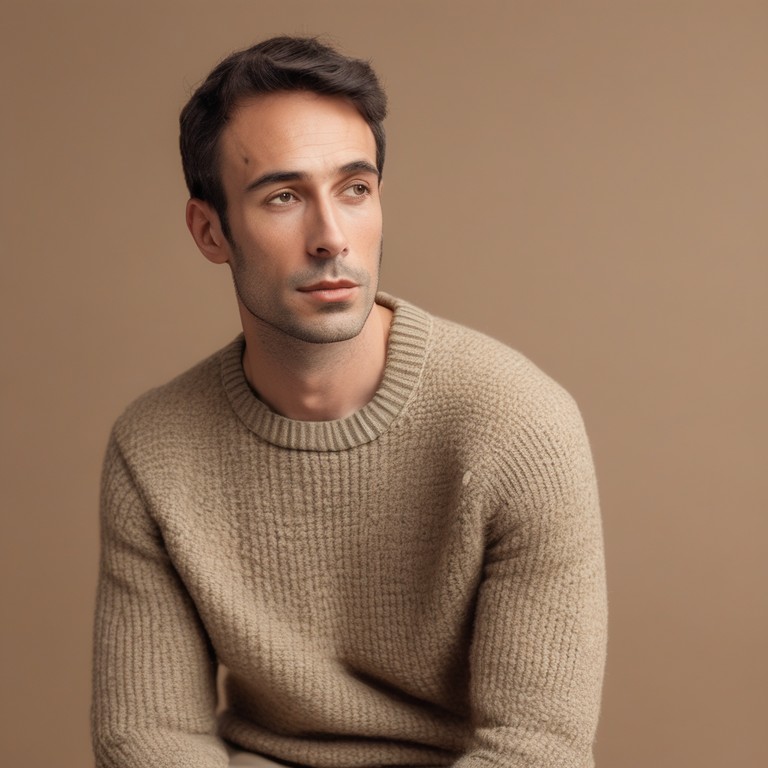} & \includegraphics[width=2.2cm,height=2.2cm]{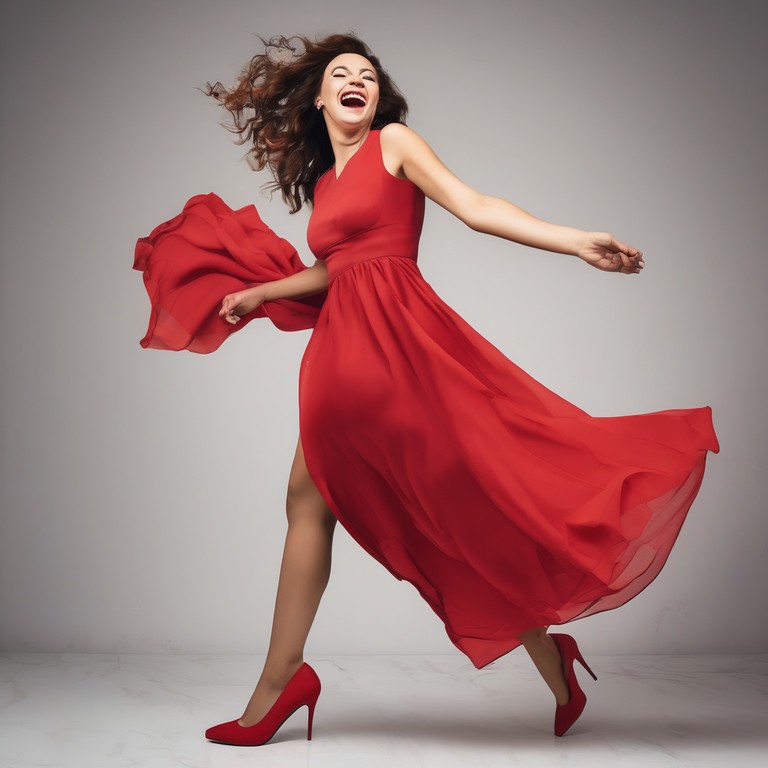} & \includegraphics[width=2.2cm,height=2.2cm]{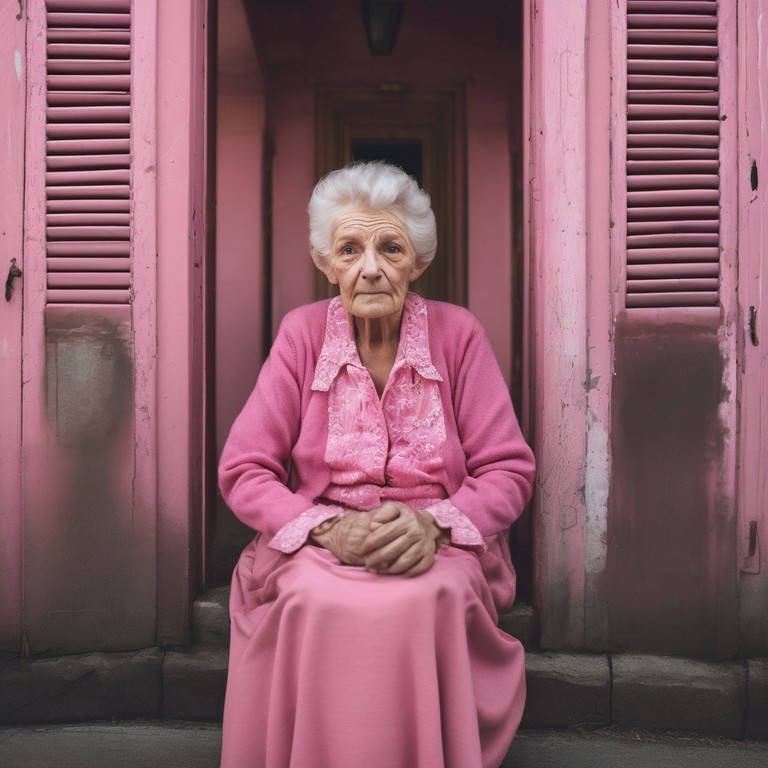} & \includegraphics[width=2.2cm,height=2.2cm]{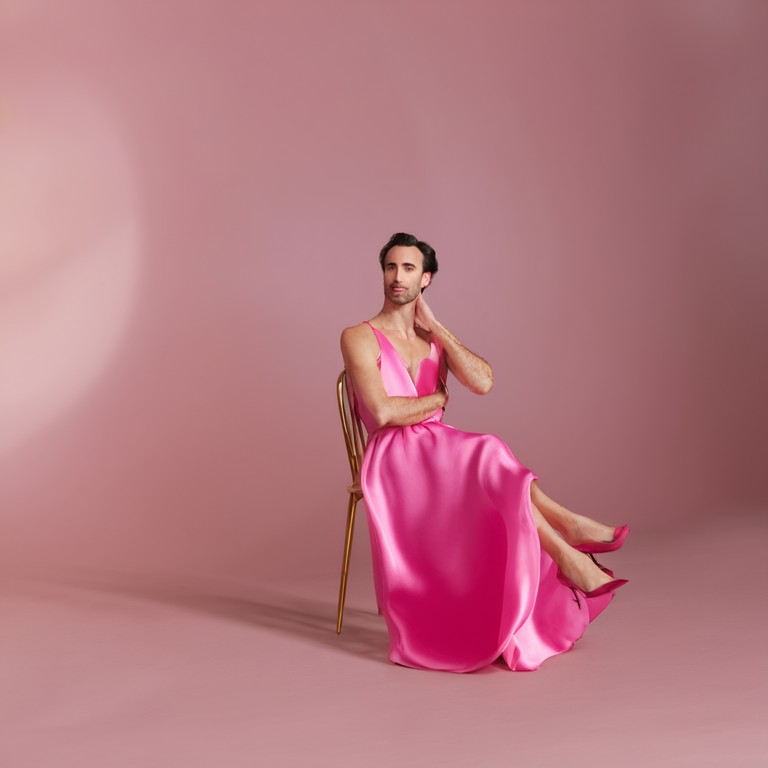} & \includegraphics[width=2.2cm,height=2.2cm]{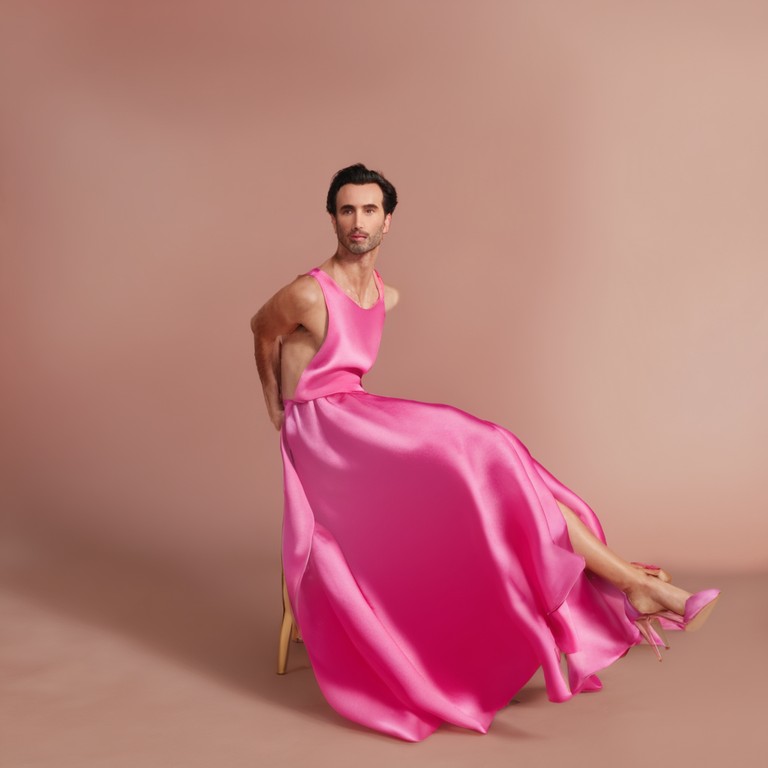} & \includegraphics[width=2.2cm,height=2.2cm]{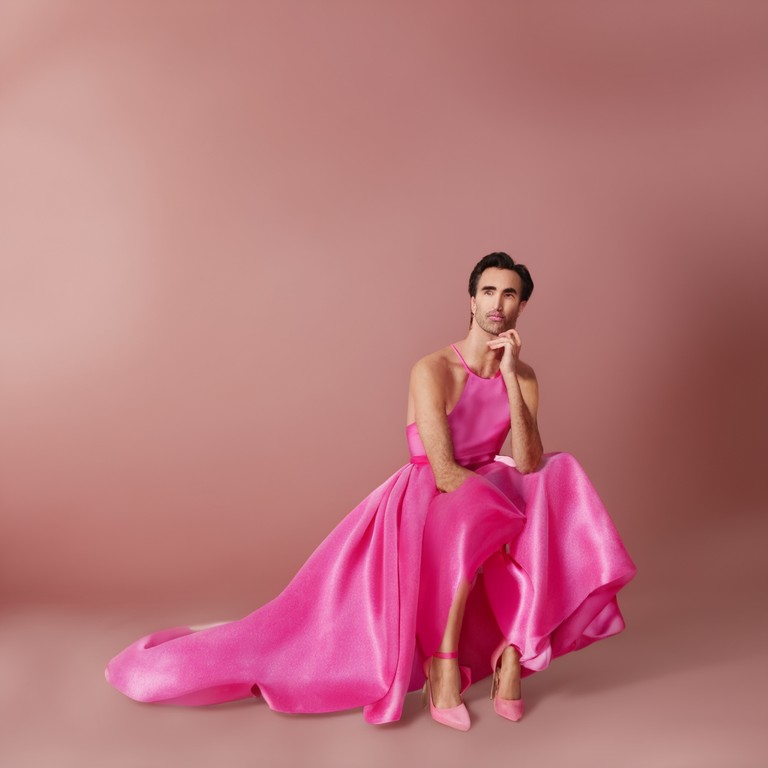} \\
        };

        \path (m1-1-1) -- (m1-1-2) node[midway] {\Large $+$};
        \path (m1-1-2) -- (m1-1-3) node[midway] {\Large $+$};
        \path (m1-1-3) -- (m1-1-4) node[midway] {\Large $=$};

        \path (m1-2-1) -- (m1-2-2) node[midway] {\Large $+$};
        \path (m1-2-2) -- (m1-2-3) node[midway] {\Large $+$};
        \path (m1-2-3) -- (m1-2-4) node[midway] {\Large $=$};

        \node[below=-0.05cm of m1-1-1] {\small ``Person''};
        \node[below=-0.05cm of m1-1-2] {\small ``Age''};
        \node[below=-0.05cm of m1-1-3] {\small ``Expression''};
        \node[below=-0.05cm of m1-1-5] {\small Results};

        \node[below=-0.05cm of m1-2-1] {\small ``Person''};
        \node[below=-0.05cm of m1-2-2] {\small ``Outfit''};
        \node[below=-0.05cm of m1-2-3] {\small ``Color''};
        \node[below=-0.05cm of m1-2-5] {\small Results};
        
        \node[below=0.6cm of m1] (bottom-container) {
            \begin{tikzpicture}
                \matrix (m2) [
                    matrix of nodes,
                    nodes={draw, minimum width=2cm, minimum height=1cm, inner sep=0pt, line width=1.5pt},
                    row sep=0.3cm,
                    column sep=0.3cm
                ] {
                    \includegraphics[width=2.2cm,height=2.2cm]{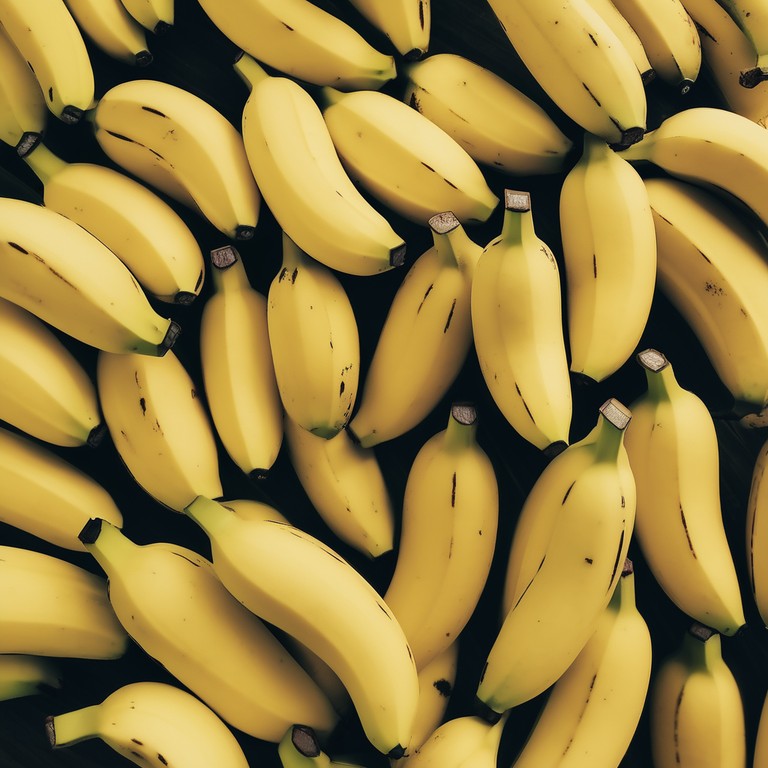} & \includegraphics[width=2.2cm,height=2.2cm]{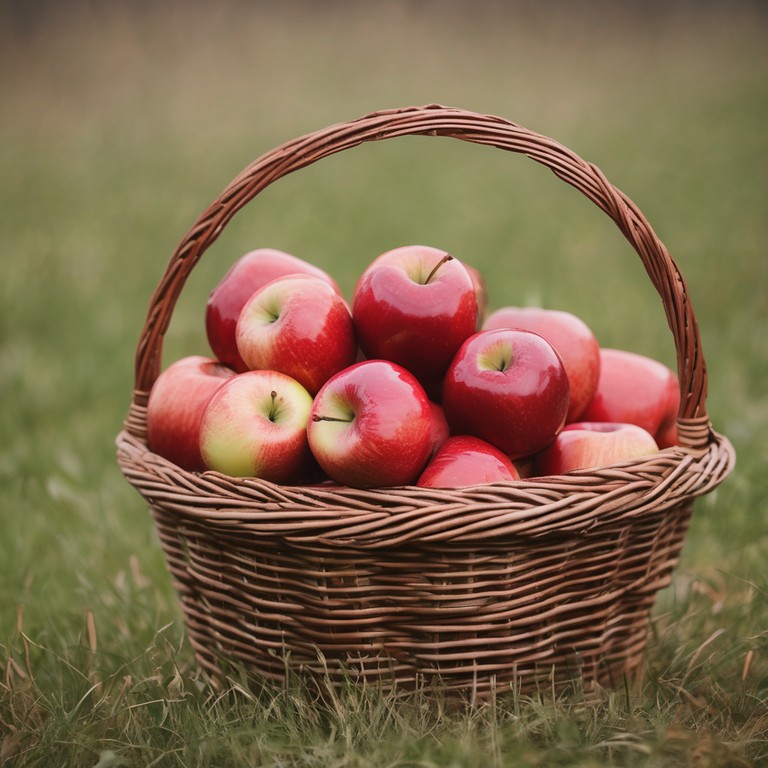} & \includegraphics[width=2.2cm,height=2.2cm]{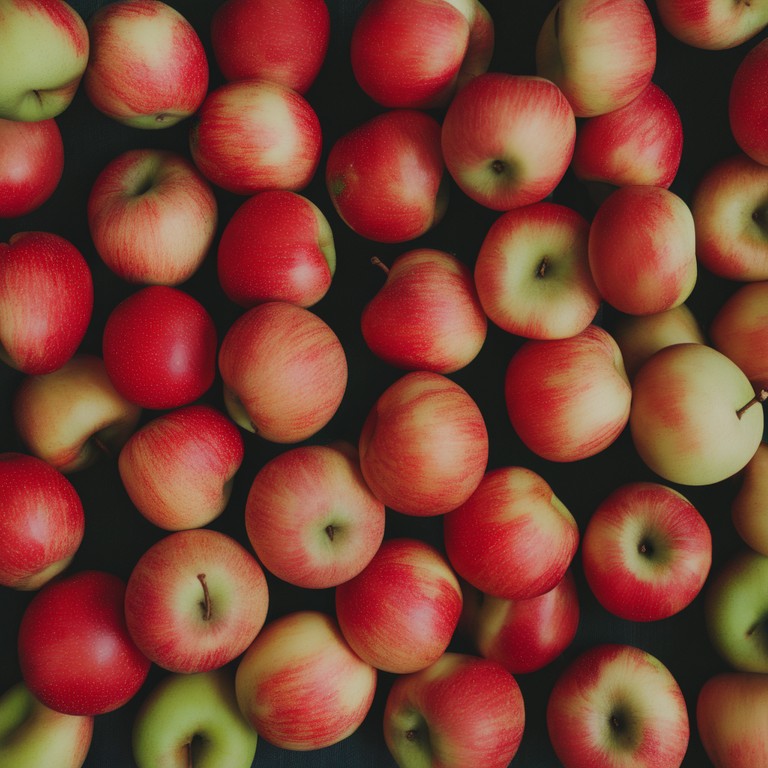} \\
                    \includegraphics[width=2.2cm,height=2.2cm]{images/qualitative_results/2_way/fruit_arrangement/scene.jpg} & \includegraphics[width=2.2cm,height=2.2cm]{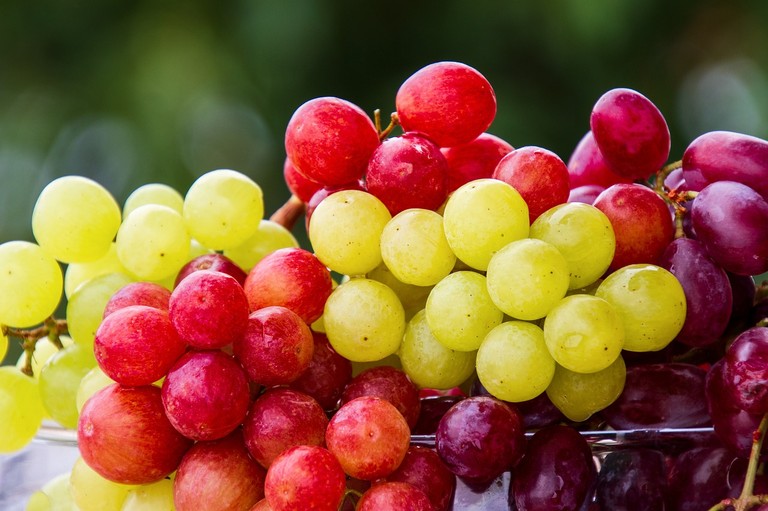} & \includegraphics[width=2.2cm,height=2.2cm]{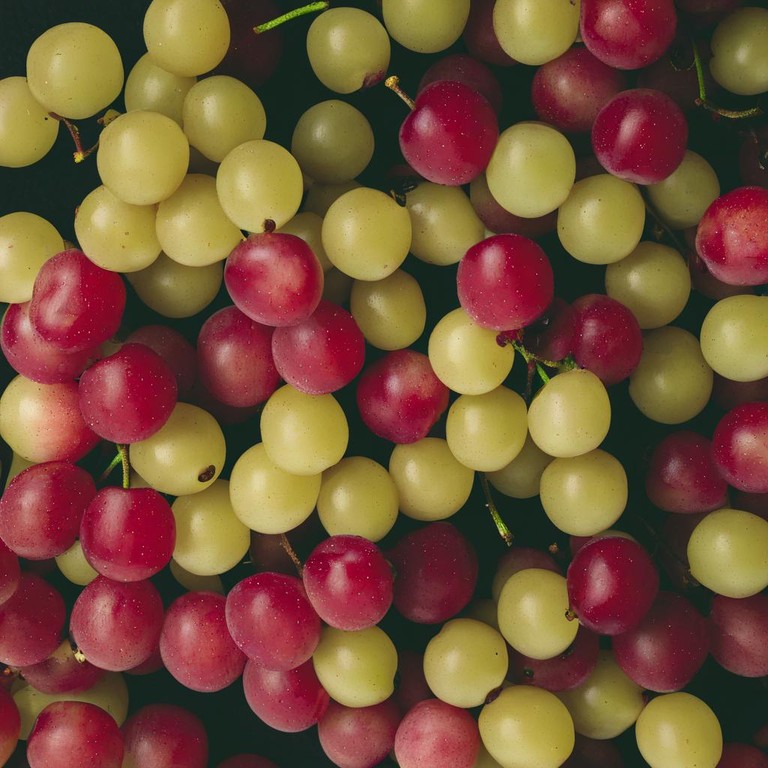} \\
                };

                \node[below=-0.05cm of m2-2-1] {\small ``Layout''};
                \node[below=-0.05cm of m2-2-2] {\small ``Fruit''};
                \node[below=-0.05cm of m2-2-3] {\small Result};

                \path (m2-1-1) -- (m2-1-2) node[midway, yshift=5.0pt] {\Large $+$};
                \path (m2-1-2) -- (m2-1-3) node[midway, yshift=5.0pt] {\Large $=$};
                \path (m2-2-1) -- (m2-2-2) node[midway, yshift=5.0pt] {\Large $+$};
                \path (m2-2-2) -- (m2-2-3) node[midway, yshift=5.0pt] {\Large $=$};
                
                \matrix (m3) [
                    matrix of nodes,
                    nodes={draw, minimum width=2cm, minimum height=1cm, inner sep=0pt, line width=1.5pt},
                    row sep=0.3cm,
                    column sep=0.3cm,
                    right=0.8cm of m2
                ] {
                    \includegraphics[width=2.2cm,height=2.2cm]{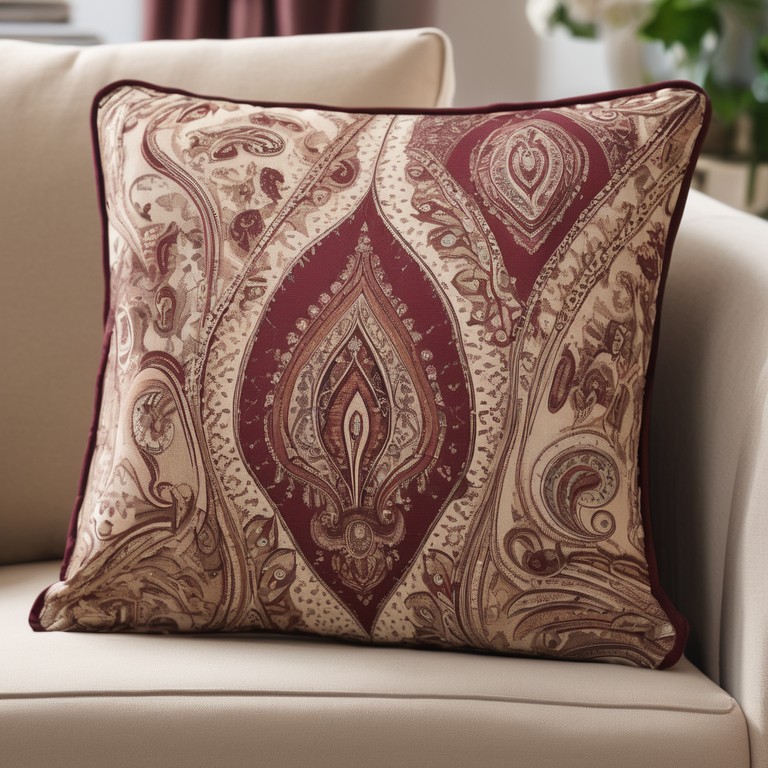} & \includegraphics[width=2.2cm,height=2.2cm]{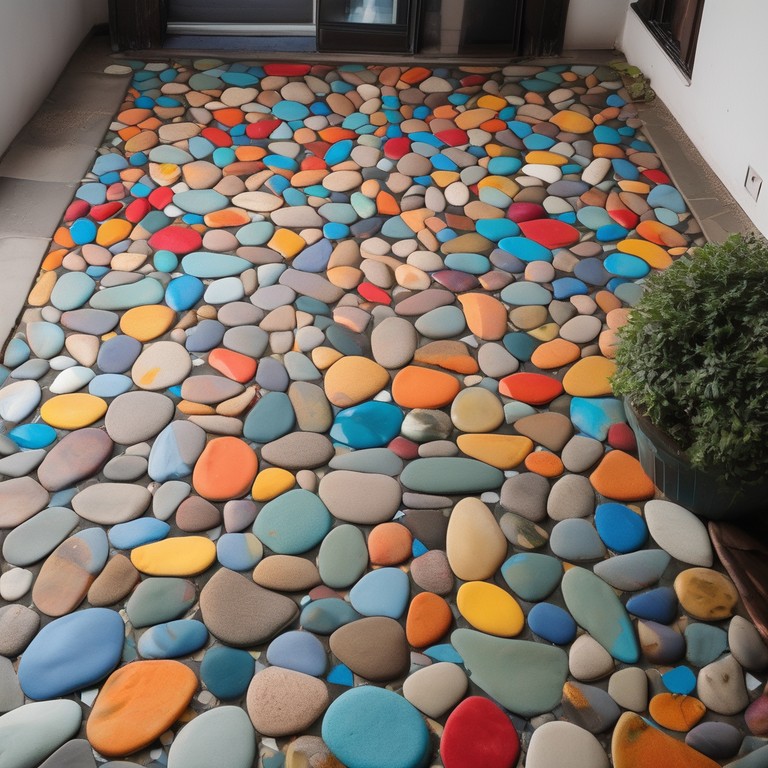} & \includegraphics[width=2.2cm,height=2.2cm]{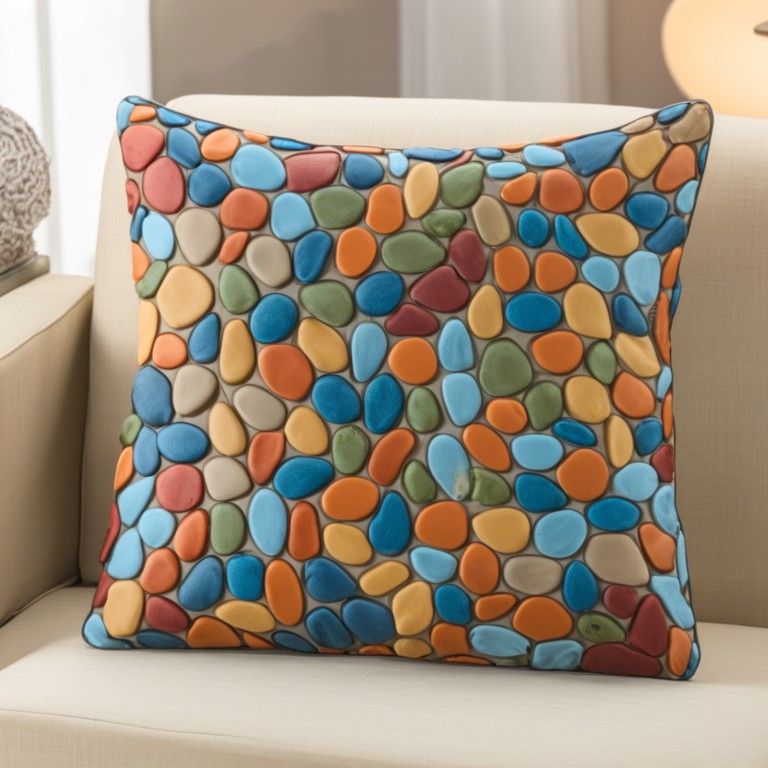} \\
                    \includegraphics[width=2.2cm,height=2.2cm]{images/qualitative_results/2_way/patterns/object.jpg} & \includegraphics[width=2.2cm,height=2.2cm]{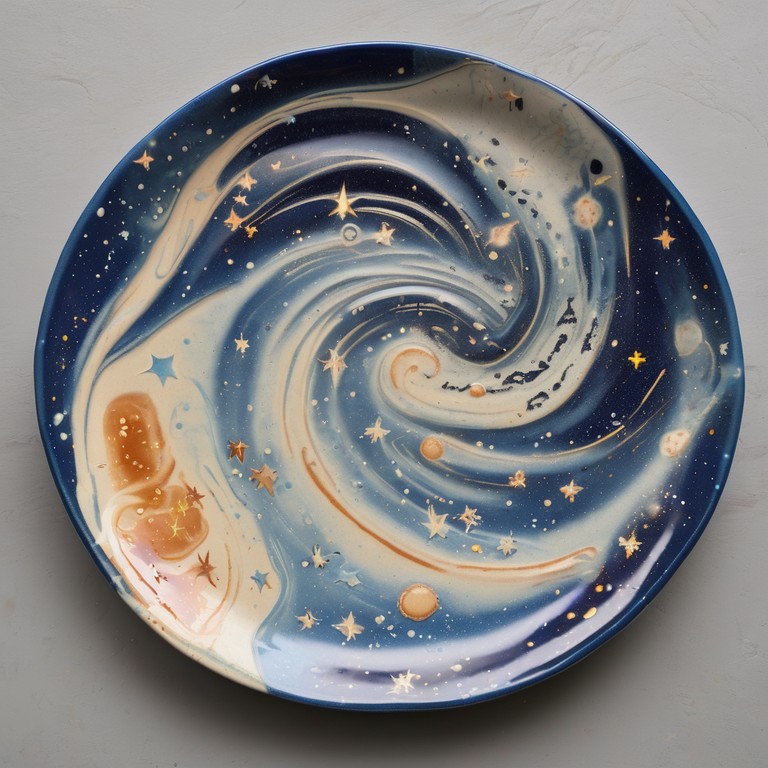} & \includegraphics[width=2.2cm,height=2.2cm]{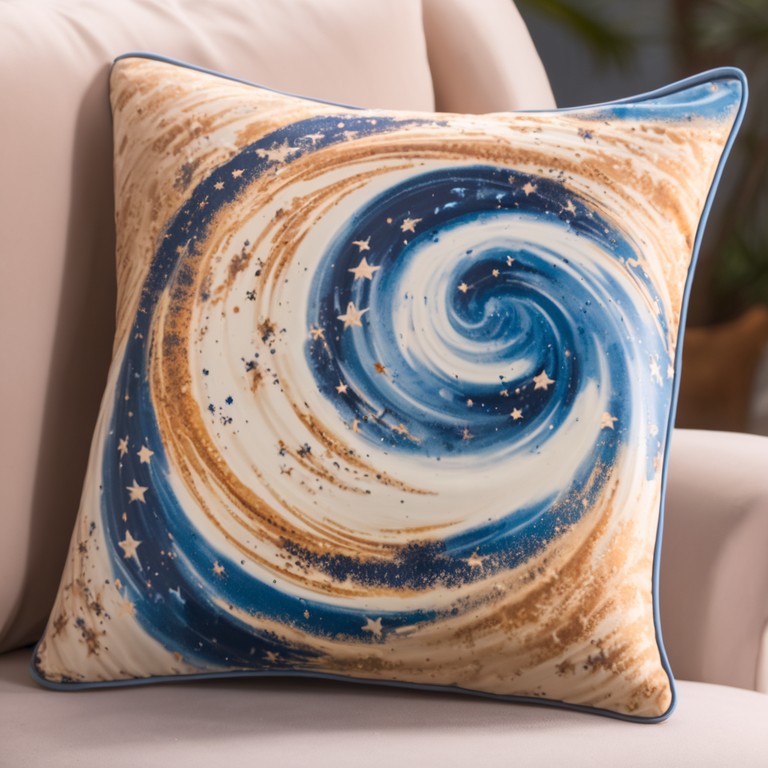} \\
                };

                \node[below=-0.05cm of m3-2-1] {\small ``Object''};
                \node[below=-0.05cm of m3-2-2] {\small ``Pattern''};
                \node[below=-0.05cm of m3-2-3] {\small Result};

                \path (m3-1-1) -- (m3-1-2) node[midway, yshift=5.0pt] {\Large $+$};
                \path (m3-1-2) -- (m3-1-3) node[midway, yshift=5.0pt] {\Large $=$};
                \path (m3-2-1) -- (m3-2-2) node[midway, yshift=5.0pt] {\Large $+$};
                \path (m3-2-2) -- (m3-2-3) node[midway, yshift=5.0pt] {\Large $=$};
                
                \begin{pgfonlayer}{background}
                    \node[draw, rounded corners, inner sep=0.3cm, fit=(m2), line width=1pt] {};
                    \node[draw, rounded corners, inner sep=0.3cm, fit=(m3), line width=1pt] {};
                \end{pgfonlayer}
            \end{tikzpicture}
        };
        
        \begin{pgfonlayer}{background}
            \node[draw, rounded corners, inner sep=0.3cm, fit=(m1), line width=1pt] {};
        \end{pgfonlayer}
    \end{tikzpicture}
    \caption{Examples of visual concept compositions enabled by IP-Composer. Our method can seamlessly tackle texture-based tasks like colorization and pattern changes, but also convey layouts or modify object-level content. }
    \label{fig:our_results}
\end{figure*}

%% file: 4_experiments.tex
\input{figures/qualitative_results_with_text}
\section{Experiments}

\paragraph{\textbf{Qualitative Results}}
We begin by showcasing our method's ability to generate novel images, composed of attributes derived from a set of input images. As demonstrated in \Cref{fig:our_results,fig:teaser,fig:additional_1,fig:additional_2}, our approach can handle a wide range of attributes. These range from injecting a subject into an existing scene, to conditioning generation on times-of-day, transferring patterns or clothing, or even mimicking poses. For many of these, collecting data to train a supervised model would be challenging, but our text-based approach can seamlessly handle them. Notably, our method is not restricted to input pairs, but can generalize to multiple conditioning components. However, this is restricted by the limited dimensionality of the embedding space. With enough components, or when using components that require high subspace dimensions, we eventually observe leakage. 

While most of our experiments involve only image-based conditioning, we note that our approach can also involve text conditioning, in the same fashion as the baseline IP-Adapter model. See \Cref{fig:results_text} for examples.

\paragraph{\textbf{Qualitative Comparisons}}
Next, we compare our method against a set of baselines aimed at tackling the same composable generation task. Specifically, we consider three approaches: (1) pOps~\citep{richardson2024popsphotoinspireddiffusionoperators}, which fine-tunes a CLIP-conditioned model to accept multiple image inputs and combine them into a single embedding for a given task (e.g., texturing or object insertion). Importantly, this approach requires a per-task dataset of roughly $50,000$ samples. We use the official pre-trained ``scene'' (subject insertion), ``texturing'' and ``union'' operators, where the last one is used as a default for scenarios that do not match the first two. (2) ProSpect~\citep{zhang2023prospectpromptspectrumattributeaware}, which inverts an image into a set of word embeddings~\citep{gal2022textual}, each containing a different aspect of the image (layout, materials etc.). New prompts can then combine embeddings from different images in order to create compositions. This approach requires lengthy per-image optimization, and is limited only to concepts which the diffusion model naturally decomposes when learning multiple embeddings. For pattern-object compositions, we follow ProSpect's content-material separation scheme, while for other concept pairs where both inputs represent content (e.g. dog-background combination), we alternate between prompts from each concept during generation. (3) Describe \& Compose, where we first ask a VLM~\citep{liu2023improvedllava} to describe the desired concept in each image, and then create a new image conditioned on both text descriptions using Composable-Diffusion~\cite{liu2023compositionalvisualgenerationcomposable}.

Since ProSpect requires lengthy optimization per image, we conduct comparisons on a small set spanning $4$ scenarios, each of which contains $2$ images for each of $2$ concepts (for a total of $16$ combinations). Importantly, our set spans both scenarios where an object should be swapped (e.g., outfits) but also ones where an object is added. Our scenarios also span components that are typically represented at different steps of the diffusion process, from layout-affecting compositions (object insertion) to appearance changes (pattern transfer). Finally, our set contains both generated and real images, ensuring that the evaluated methods are not restricted to strictly in-domain inputs. For a fair comparison, we do not tune our subspace rank for each task, but set $r=30$ for low-variation tasks like outfit replacement and $r=120$ for high-variation tasks like pattern changes.

Sample qualitative results are shown in \cref{fig:qualitative_comp}. When compared to the training-based pOps, our method achieves comparable results on subject insertion without requiring any specialized datasets or model tuning. For tasks where no dedicated pOps encoder is available, we utilized their 'union' operator, as it is the most relevant for achieving the desired results. On these tasks, our approach significantly outperformed the results achieved by pOps, highlighting the generalization capabilities of our text-based approach. Compared to ProSpect, our approach can better handle layout changes and can tackle concepts which are not naturally separated when tuning multiple word embeddings. Finally, the description-based method struggles to convey the specifics of each concept and has significant leakage (cherry petals in the dog image, hat in the emotion transfer). As also shown in previous work~\citep{gal2022textual}, we further observe that the use of long descriptions makes the model more likely to discard parts of the prompt in the generated results.

\input{figures/quantitative_comparison}

\input{figures/user_study}
\input{figures/qualitative_comparisons}
\input{figures/qualitative_ablation}
\paragraph{\textbf{Quantitative comparisons.}}
To better evaluate the performance of our method, we  conduct a quantitative evaluation. Here, we ask GPT-4 to create a description of the target concept in each image, then manually verify the description and modify it to ensure that it does not contain leakage or misses important attributes. We then compute the CLIP-space distance between the text describing each concept in an input pair and the generated image that aims to combine them. To further quantify the leakage of unwanted properties, we employ the same method to generate descriptions of all image properties \textit{not related} to the concept that we aim to extract. We similarly measure the CLIP-space similarity between each generated image and these descriptions. However, here the goal is to achieve a lower score, indicating that the non-concept properties did not leak into the output.

The results are shown in ~\cref{fig:quantitative_comparison}. These results mirror those of the qualitative evaluation, demonstrating that our approach can achieve high similarity to both desired concepts, while minimizing undesired leakage. Here, we additionally report results when tuning our method's rank parameter for each individual task (``IP-Composer (Tuned)''). Doing so further enhances our performance, demonstrating that while even default parameters can achieve state-of-the-art results, a dedicated user has room to improve them further.

Finally, we verify our results using a user study. Here, we use a 2-alternative forced choice setup. We show each user a pair of input images with a caption denoting which concept should be extracted from each. Then, we show them an image generated by our method and an image generated by one of the baselines, and ask them to select the one that better combines the visual concepts from the two images. We collected a total of $560$ responses from $35$ different users. The results are reported in ~\cref{fig:user_study}, confirming that our approach is significantly preferred over existing baselines.

\input{figures/quantitative_ablation}
\input{figures/multi_step}

\paragraph{\textbf{Ablation}}
Next, we conduct an ablation study where we explore the use of different ways to combine IP-Adapter encodings for compositional generation. IP-Adapter itself first extracts a CLIP-embedding from each image, then converts these embeddings into a set of tokens which are used to condition the diffusion model through new cross-attention layers. Hence, we examine a scenario where we interpolate between the CLIP-embeddings of the two input images, as well as one where we concatenate their IP-Adapter tokens before feeding them into the diffusion model. Finally, we also examine a scenario where we span the CLIP-subspace of a concept using images rather than text. Here, we use the same LLM-generated descriptions of variations of a concept in order to generate images depicting these variations. Then, we encode them using CLIP, and use their CLIP-embeddings to span the concept space.

We evaluate the approaches using the same metrics of the quantitative comparisons. However, since none of the approaches require per-image training, we expand our evaluation set to $150$ images. Results are shown in \cref{fig:quantitative_ablation,fig:ablation}. Compared to concatenations and interpolations, our approach offers better control, with the ability to designate specific concepts to be extracted from an image and avoid significant leakage. The image-based approach suffers from increased leakage because the generated images tend to fill-in content unrelated to the prompt, such as the creation of an appropriate background. Since this content varies between the different images, it is represented in the dominant directions in the SVD.

%% file: figures/qualitative_results_with_text.tex
\begin{table}[htbp]
    \centering
    \setlength{\tabcolsep}{3pt}
    \small
    \begin{tabular}{cccc}
        Reference & Concept & Text & Result  \\
        Image & Image &  Prompt &  \\[3pt]
        \raisebox{-.5\height}{\includegraphics[width=0.225\linewidth,height=0.225\linewidth]{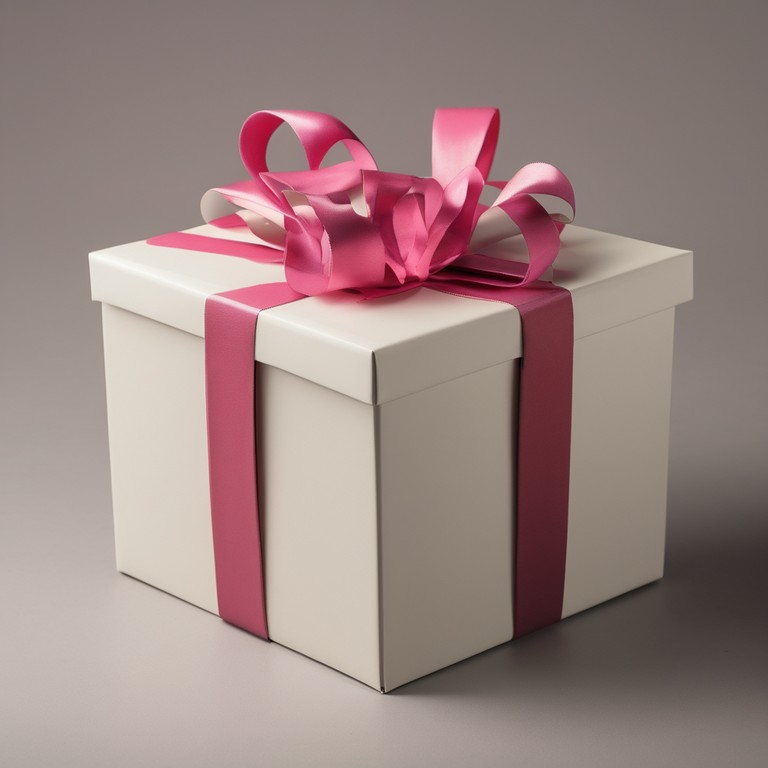}} &
        \raisebox{-.5\height}{\includegraphics[width=0.225\linewidth,height=0.225\linewidth]{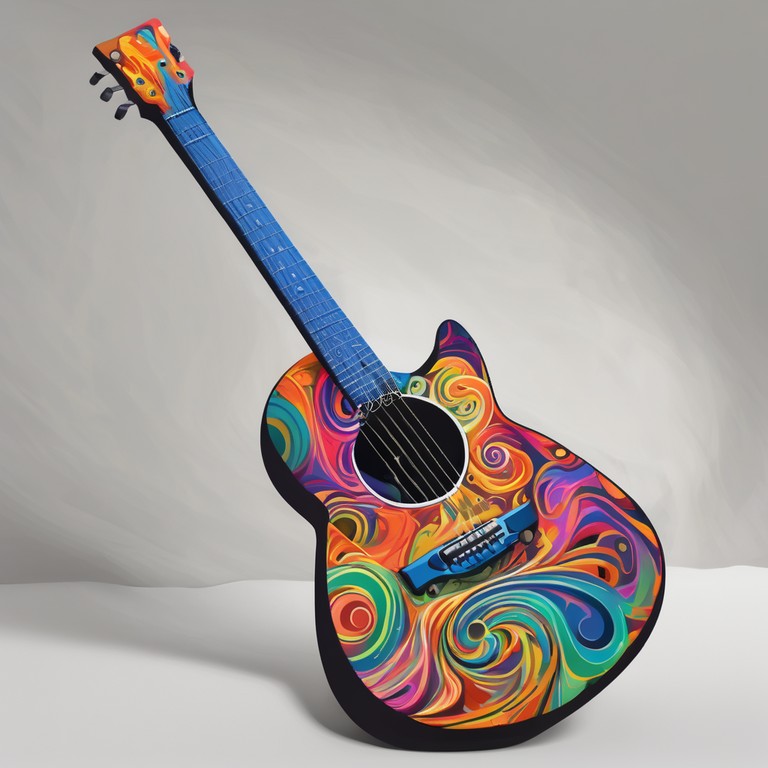}} &
        \raisebox{-.5\height}{\parbox[c]{0.225\linewidth}{\centering\large ``... Round''}} &
        \raisebox{-.5\height}{\includegraphics[width=0.225\linewidth,height=0.225\linewidth]{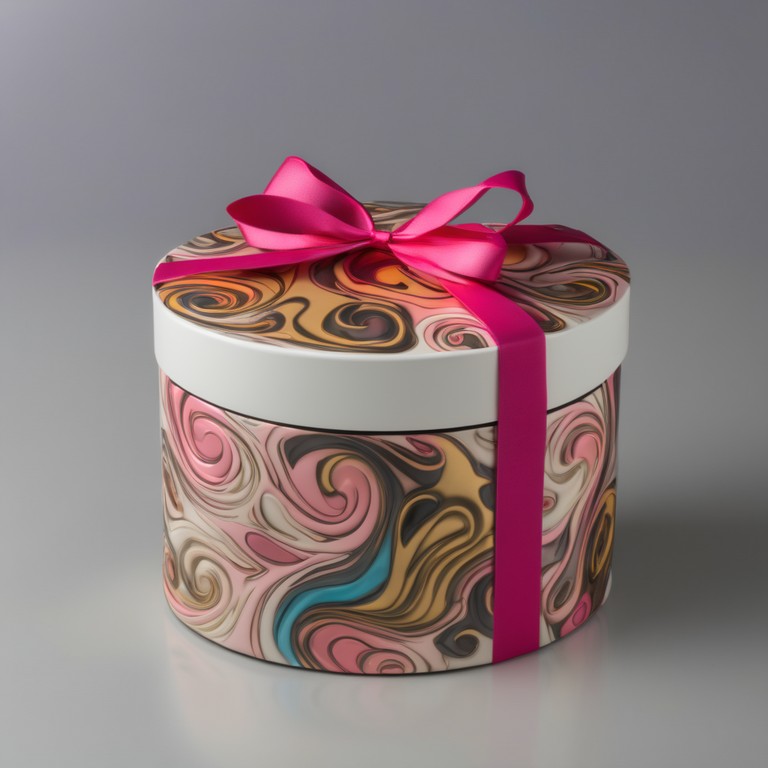}} \\[2ex]
        ``Object'' & ``Pattern'' &  & \\[3pt]
        \raisebox{-.5\height}{\includegraphics[width=0.225\linewidth,height=0.225\linewidth]{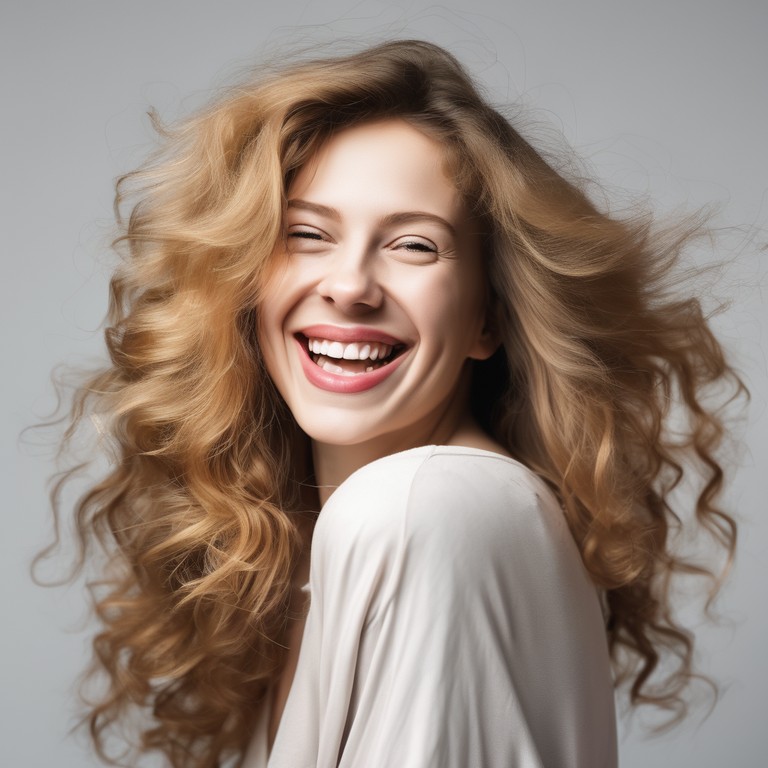}} &
        \raisebox{-.5\height}{\includegraphics[width=0.225\linewidth,height=0.225\linewidth]{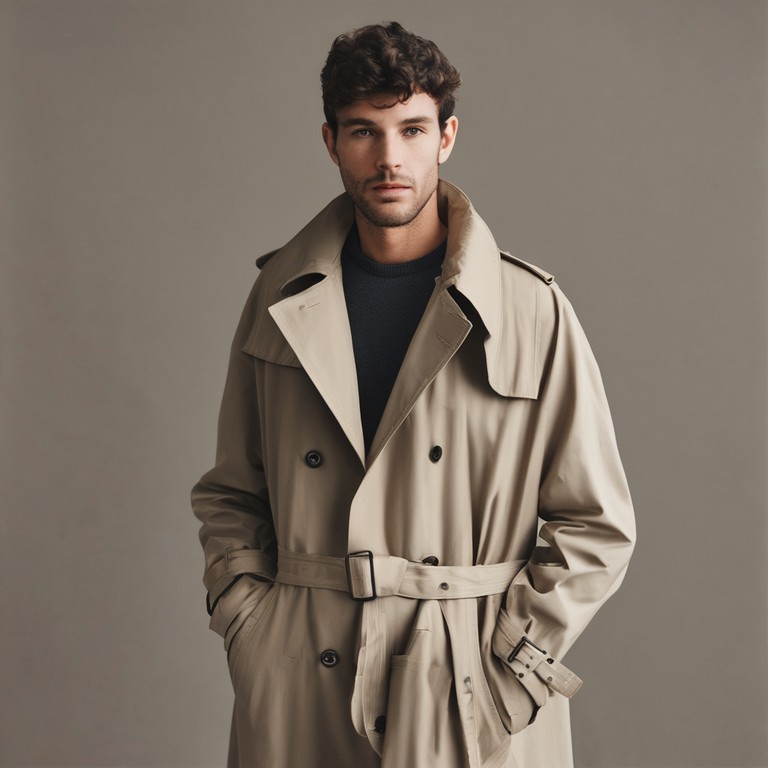}} &
        \raisebox{-.5\height}{\parbox[c]{0.225\linewidth}{\centering\large ``... Wearing sunglasses''}} &
        \raisebox{-.5\height}{\includegraphics[width=0.225\linewidth,height=0.225\linewidth]{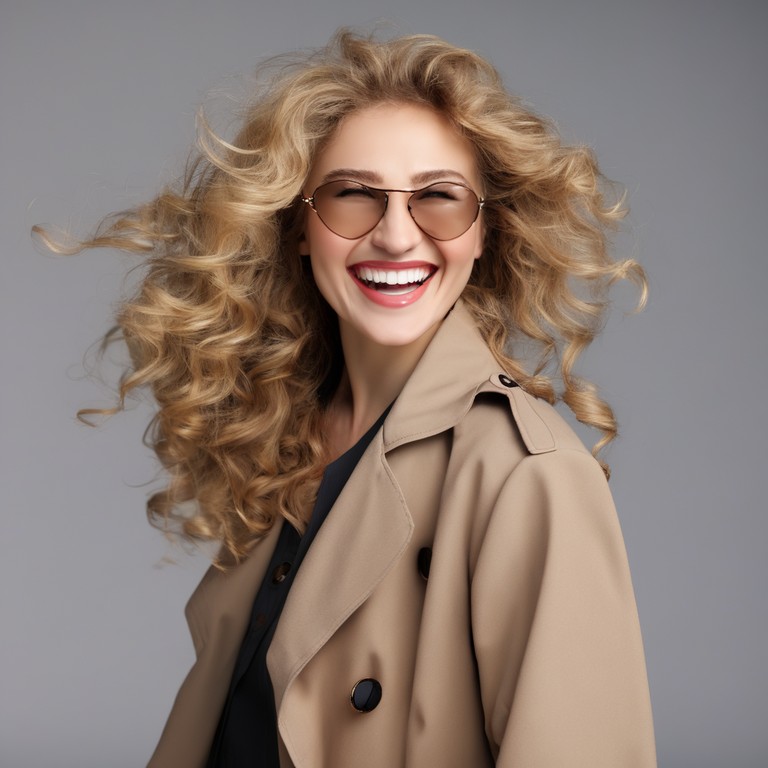}} \\[2ex]
        ``Person'' & ``Outfit'' &  & 
    \end{tabular}
    \captionof{figure}{Results demonstrating our method’s ability to integrate text prompts alongside image embeddings, leveraging IP-Adapter’s built-in support for text conditioning.}
    \label{fig:results_text}
\end{table}

%% file: figures/quantitative_comparison.tex
\begin{figure}
    \begin{tabular}{cc}
        \includegraphics[width=0.45\linewidth,height=0.45\linewidth]{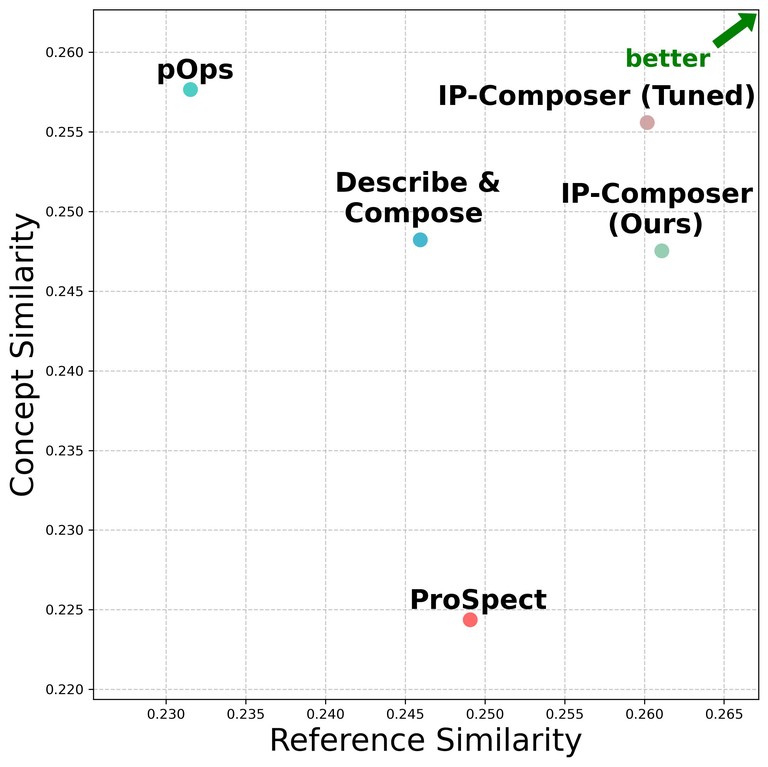} &
        \includegraphics[width=0.45\linewidth,height=0.45\linewidth]{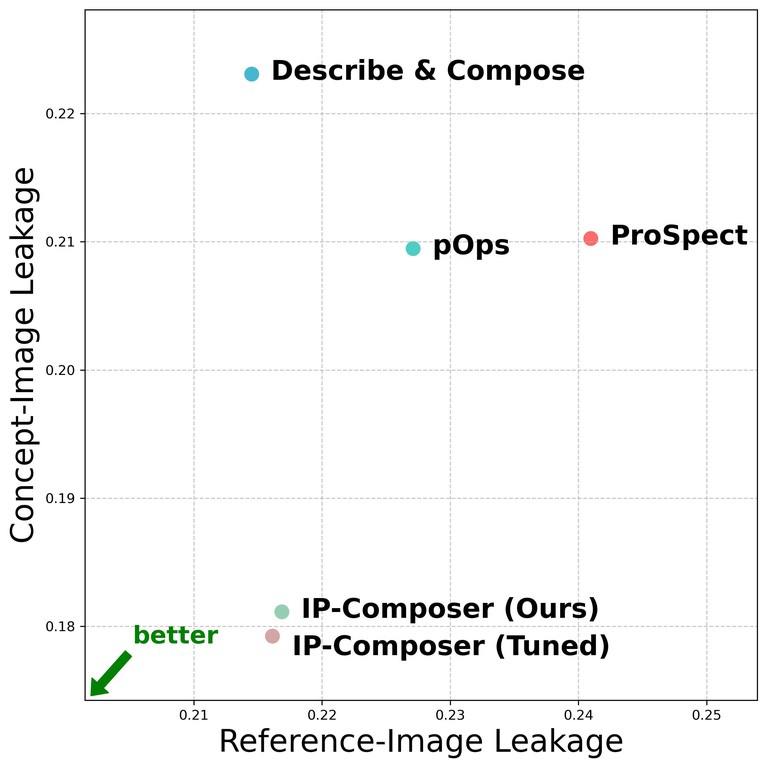} 
    \end{tabular}
    \caption{Quantitative results mimic our qualitative observations, showing that IP-Composer can successfully compete with and even outperform existing training-based methods. }\label{fig:quantitative_comparison}
\end{figure}

%% file: figures/user_study.tex
\begin{figure}
    \centering
    \includegraphics[width=0.95\linewidth]{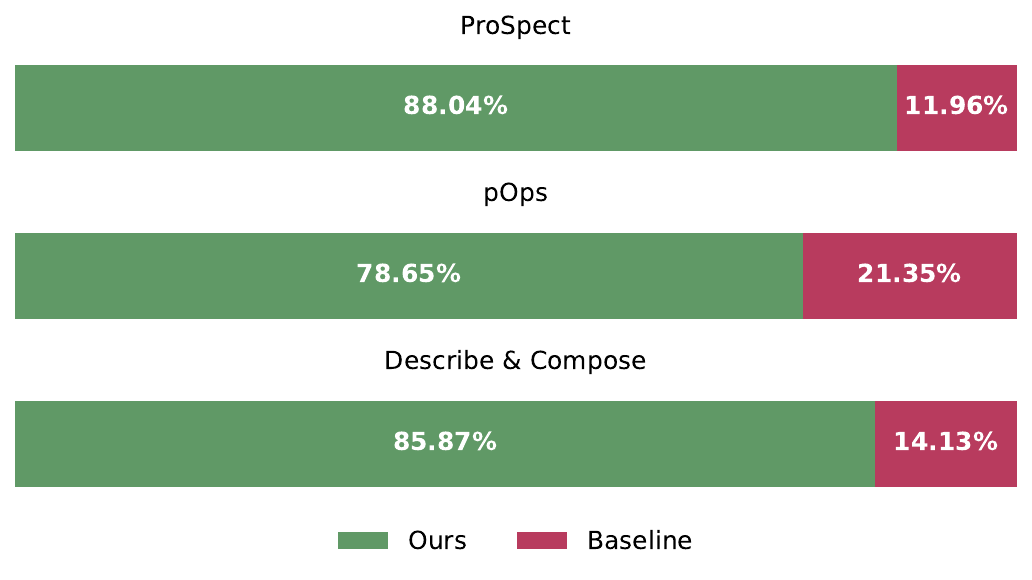}
    \caption{User study results. Our approach is commonly preferred by users, even when compared with training-based methods.}\label{fig:user_study}
\end{figure}

%% file: figures/qualitative_comparisons.tex
\begin{table*}[htbp]
    \centering

    \setlength{\belowcaptionskip}{-5pt}
    \setlength{\abovecaptionskip}{4pt}
    
    \begin{tabular}{cc:cccc}
        
        Reference & Concept & IP-Composer & \multirow{2}{*}{pOps} & \multirow{2}{*}{ProSpect} & Describe \& \\

        Image & Image & (ours) &  &  & Compose \\
        \includegraphics[width=0.132\textwidth,height=0.132\textwidth]{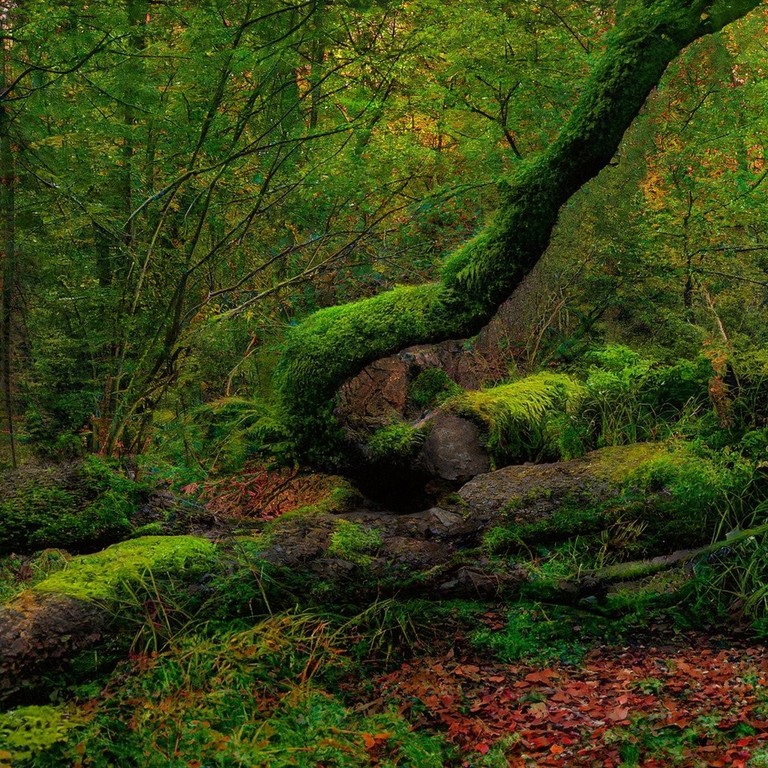} &
        \includegraphics[width=0.132\textwidth,height=0.132\textwidth]{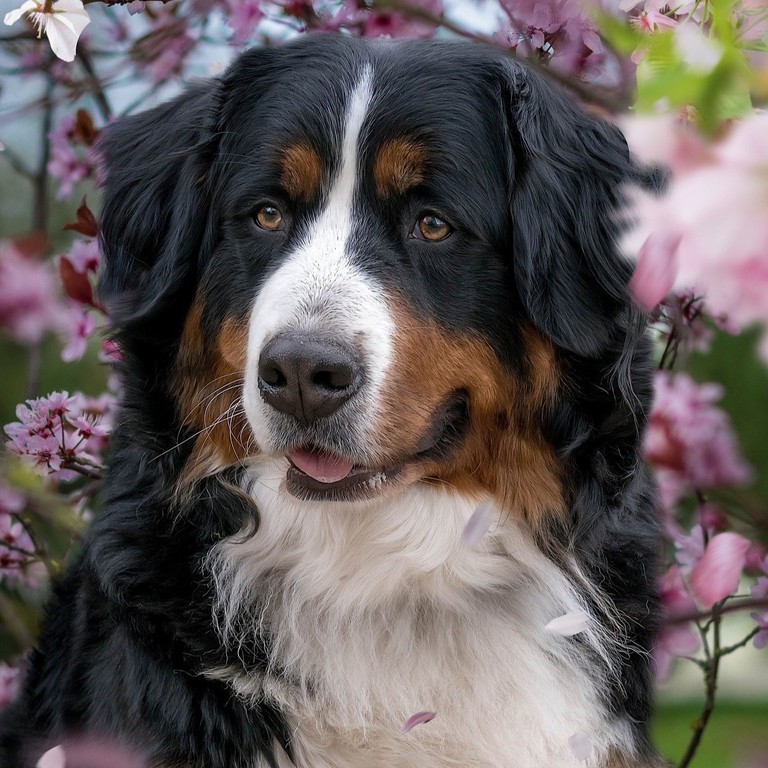} &
        \includegraphics[width=0.132\textwidth,height=0.132\textwidth]{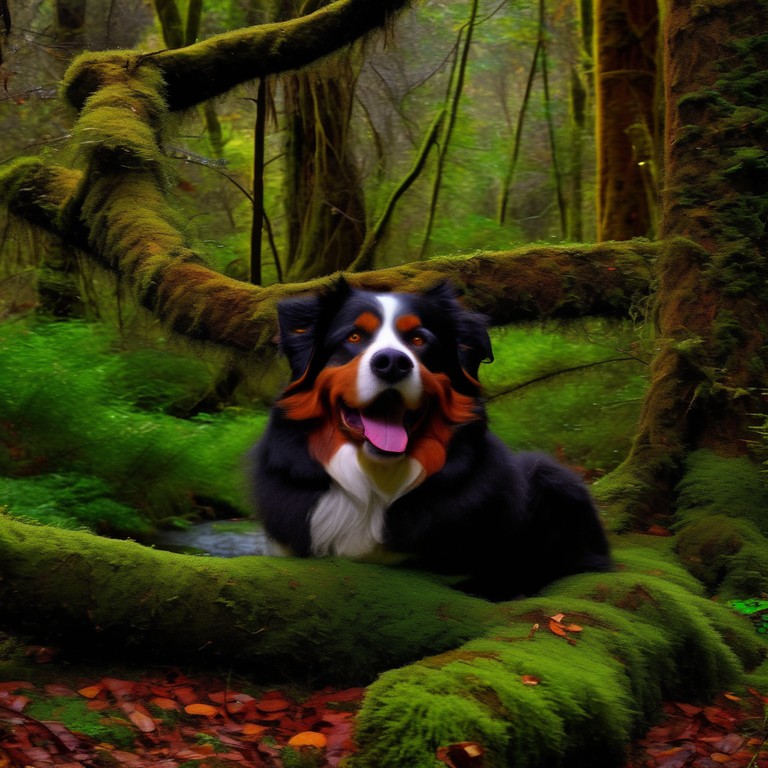} &
        \includegraphics[width=0.132\textwidth,height=0.132\textwidth]{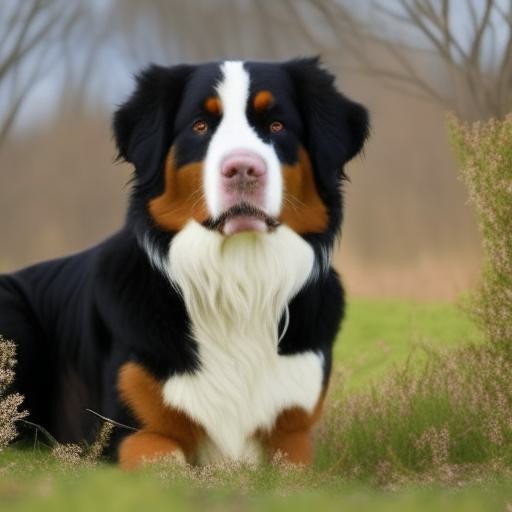} &
        \includegraphics[width=0.132\textwidth,height=0.132\textwidth]{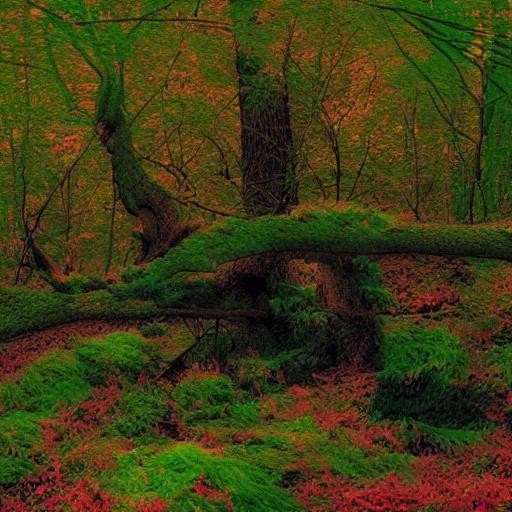} &
        \includegraphics[width=0.132\textwidth,height=0.132\textwidth]{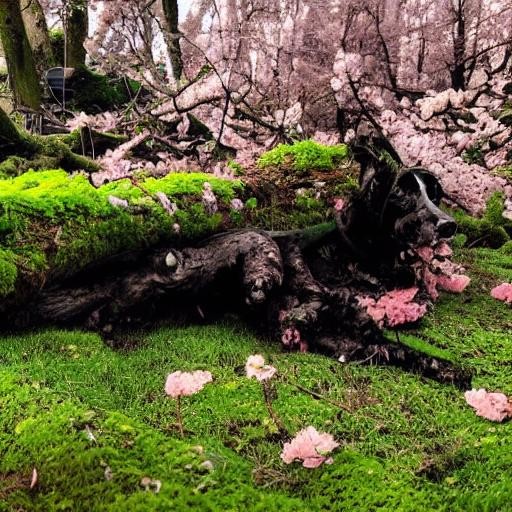} \\
        ``Background'' & ``Dog'' &  &  &  &  \\
        
        \includegraphics[width=0.132\textwidth,height=0.132\textwidth]{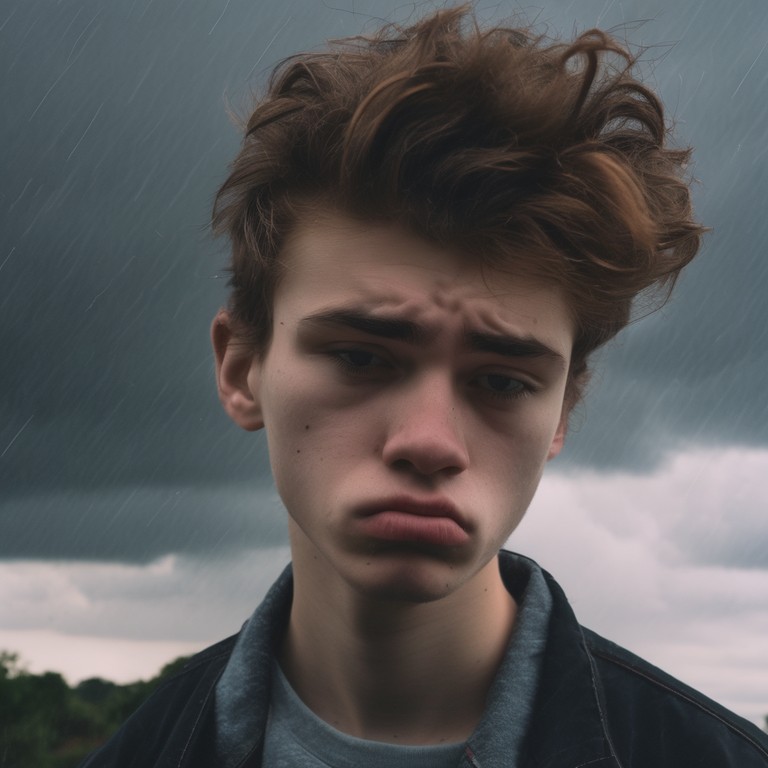} &
        \includegraphics[width=0.132\textwidth,height=0.132\textwidth]{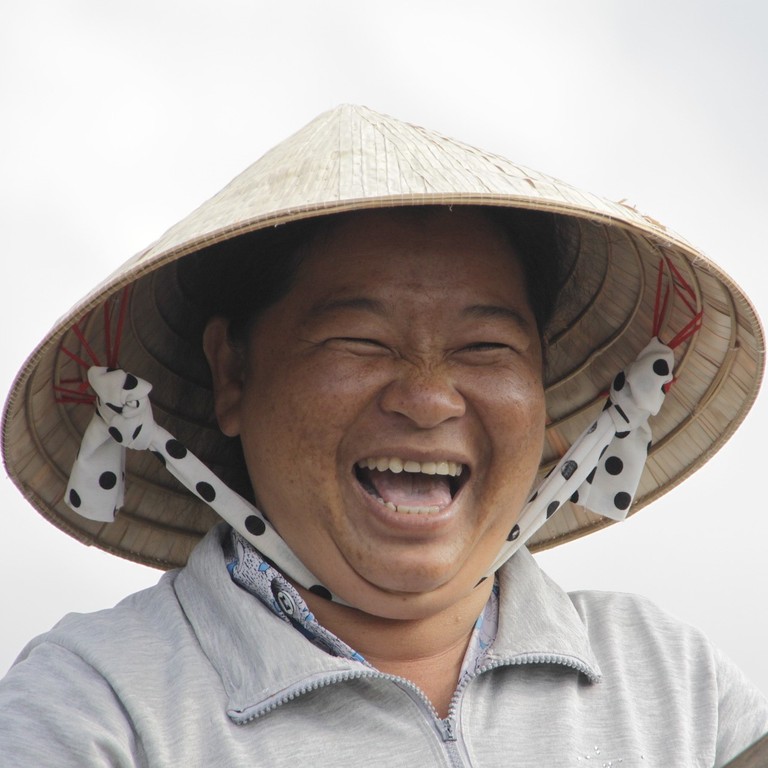} &
        \includegraphics[width=0.132\textwidth,height=0.132\textwidth]{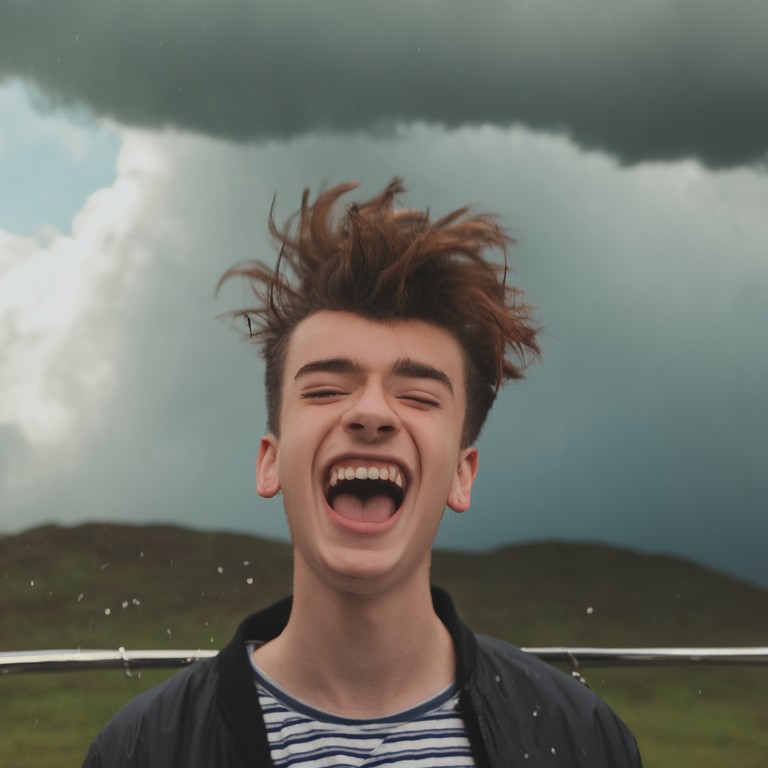} &
        \includegraphics[width=0.132\textwidth,height=0.132\textwidth]{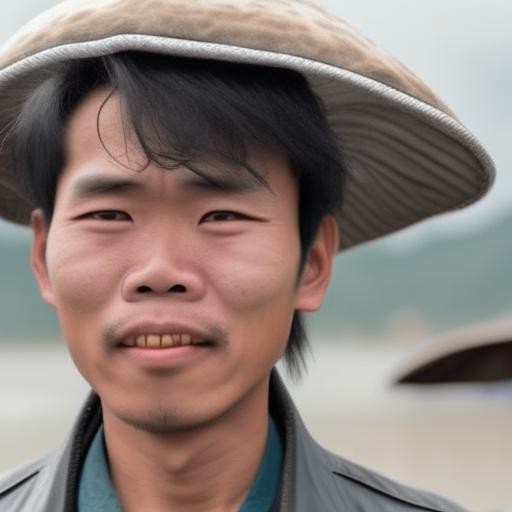} &
        \includegraphics[width=0.132\textwidth,height=0.132\textwidth]{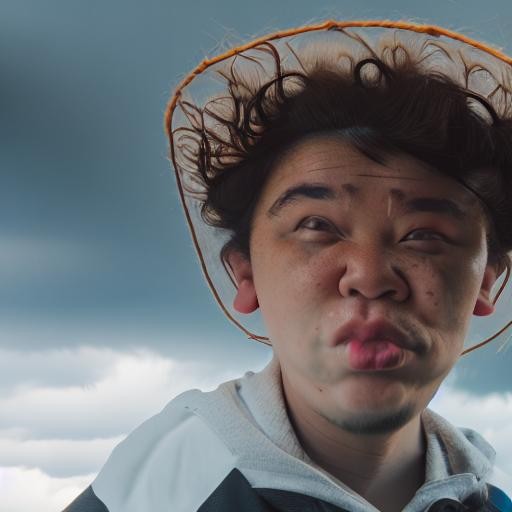} &
        \includegraphics[width=0.132\textwidth,height=0.132\textwidth]{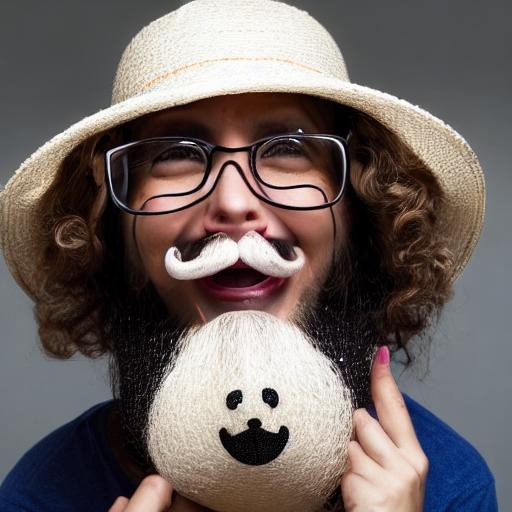} \\
        ``Person'' & ``Emotion'' &  &  &  &  \\
        
        \includegraphics[width=0.132\textwidth,height=0.132\textwidth]{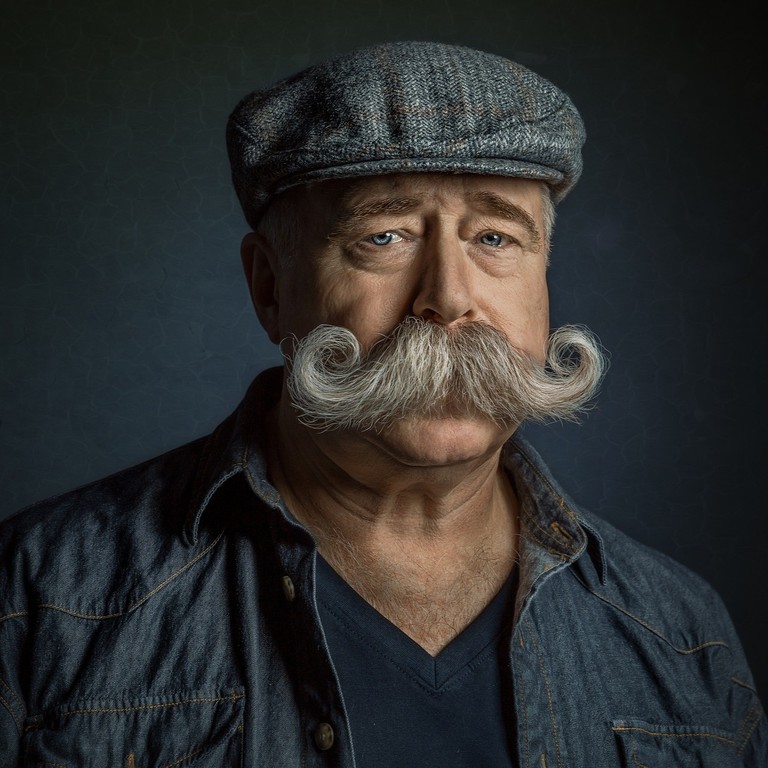} &
        \includegraphics[width=0.132\textwidth,height=0.132\textwidth]{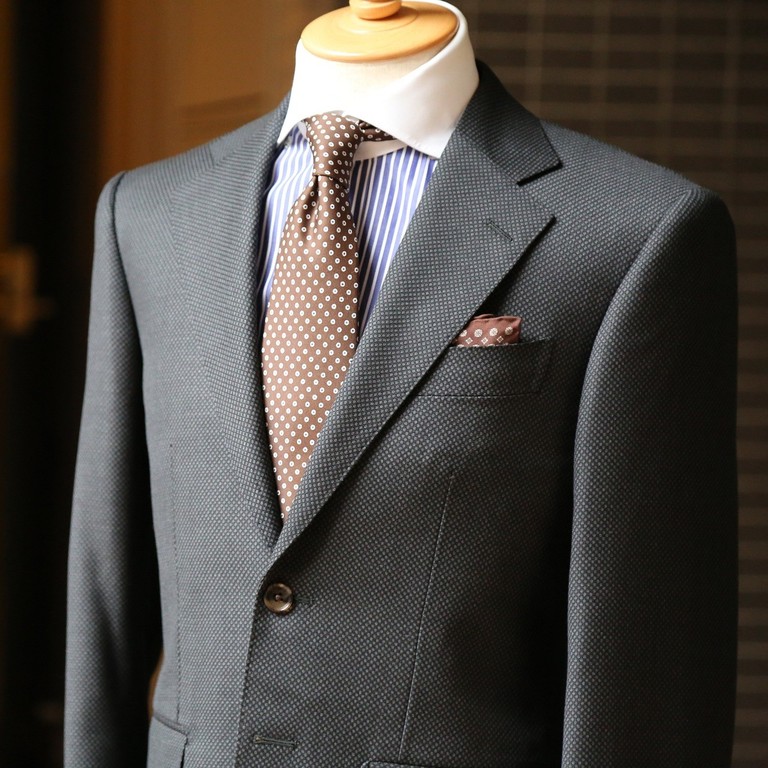} &
        \includegraphics[width=0.132\textwidth,height=0.132\textwidth]{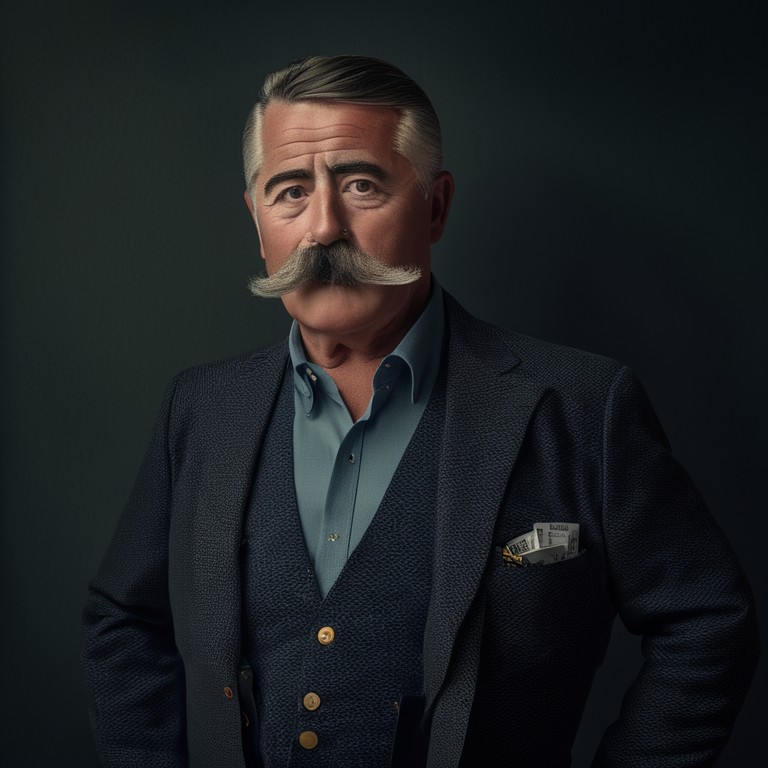} &
        \includegraphics[width=0.132\textwidth,height=0.132\textwidth]{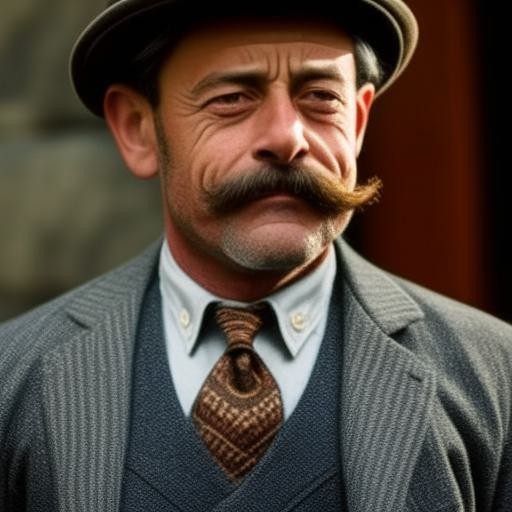} &
        \includegraphics[width=0.132\textwidth,height=0.132\textwidth]{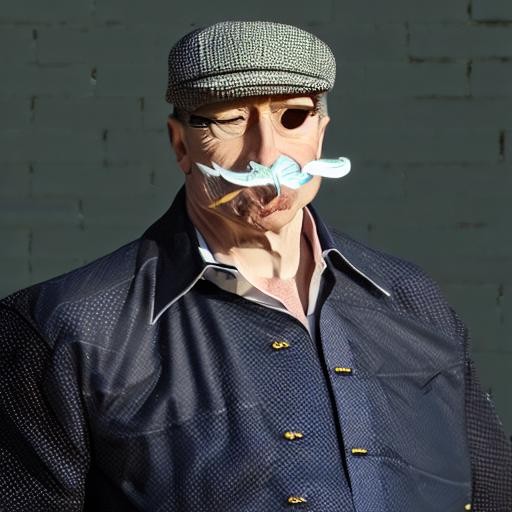} &
        \includegraphics[width=0.132\textwidth,height=0.132\textwidth]{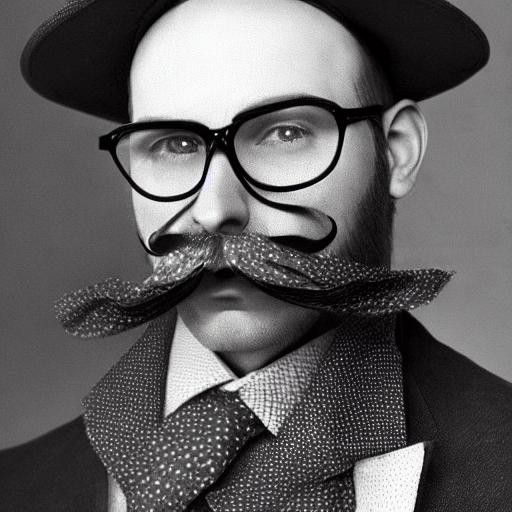} \\
        ``Person'' & ``Outfit'' &  &  &  &  \\
        
        \includegraphics[width=0.132\textwidth,height=0.132\textwidth]{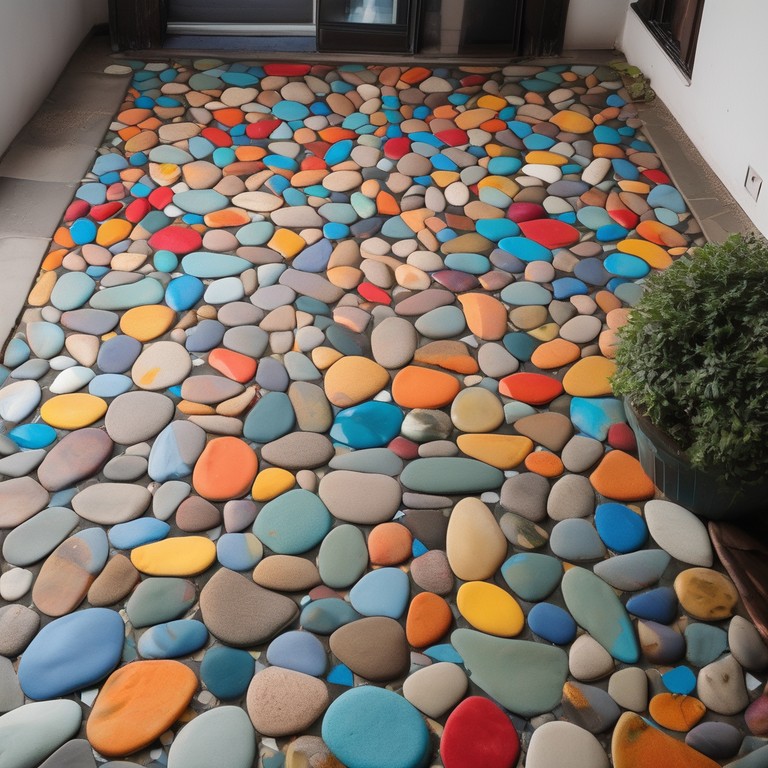} &
        \includegraphics[width=0.132\textwidth,height=0.132\textwidth]{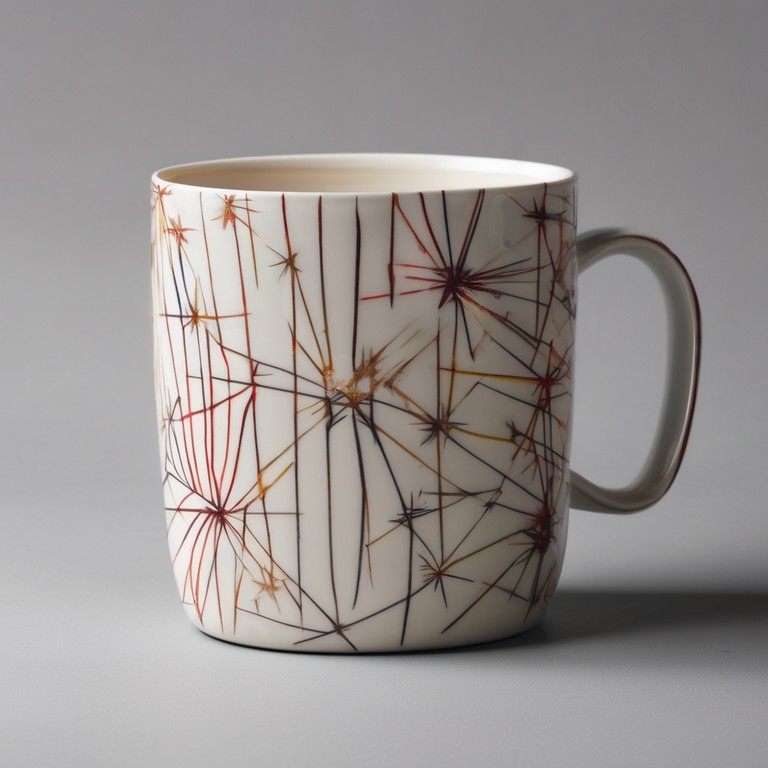} &
        \includegraphics[width=0.132\textwidth,height=0.132\textwidth]{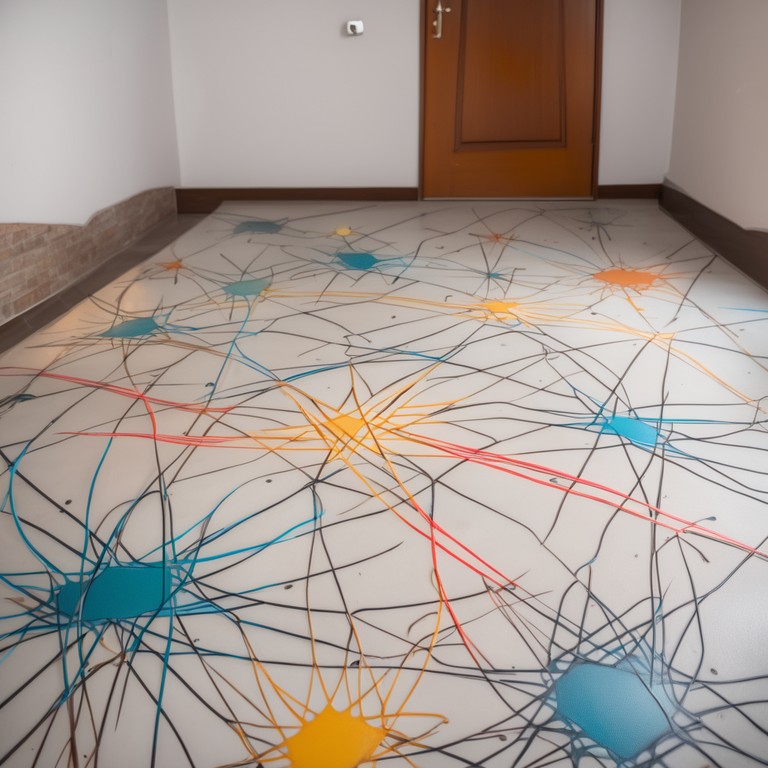} &
        \includegraphics[width=0.132\textwidth,height=0.132\textwidth]{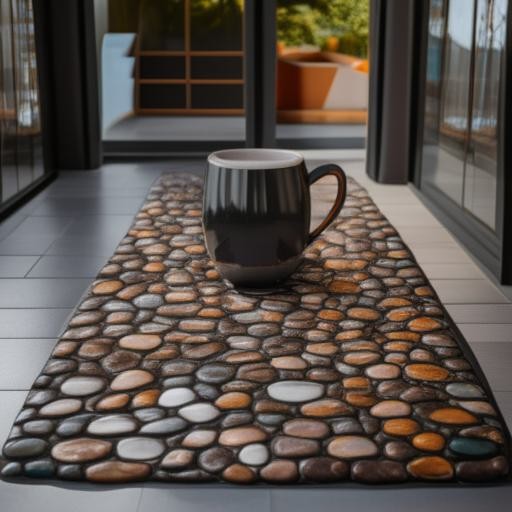} &
        \includegraphics[width=0.132\textwidth,height=0.132\textwidth]{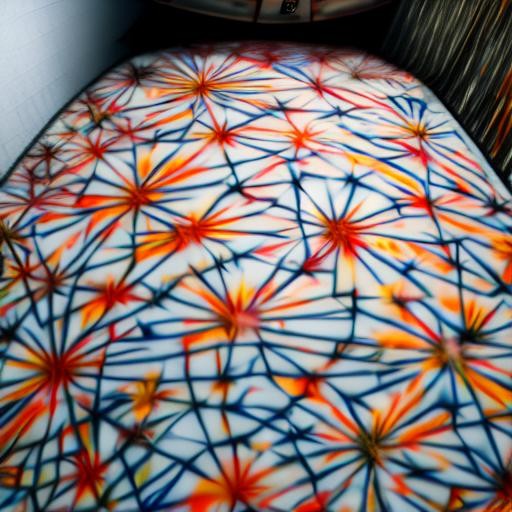} &
        \includegraphics[width=0.132\textwidth,height=0.132\textwidth]{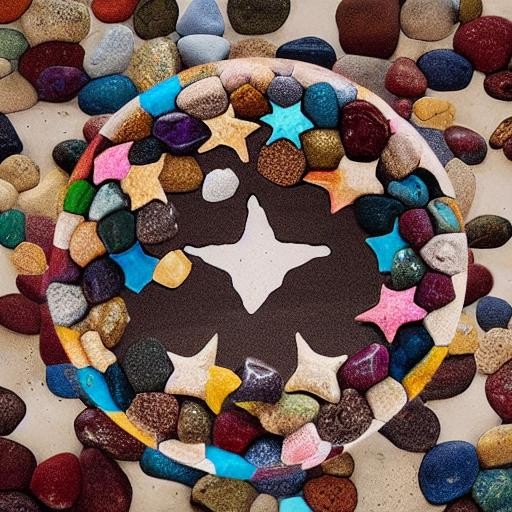} \\
        ``Object'' & ``Pattern'' &  &  &  & 
    \end{tabular}
    \captionof{figure}{Baseline comparisons. IP-Composer achieves comparable or better results than the baselines, including those trained on the task using task-specific data.}
    \label{fig:qualitative_comp}
\end{table*}

%% file: figures/qualitative_ablation.tex
\begin{table*}[htbp]
    \centering
    \setlength{\belowcaptionskip}{-5pt}
    \setlength{\abovecaptionskip}{4pt}
    
    \begin{tabular}{cc:cccc}
        
        Reference & Concept & IP-Composer & IPA & IPA & Image-based \\

        Image & Image & (ours) & w/ Concat & w/ Interpolation & Subspace \\
        \includegraphics[width=0.132\textwidth,height=0.132\textwidth]{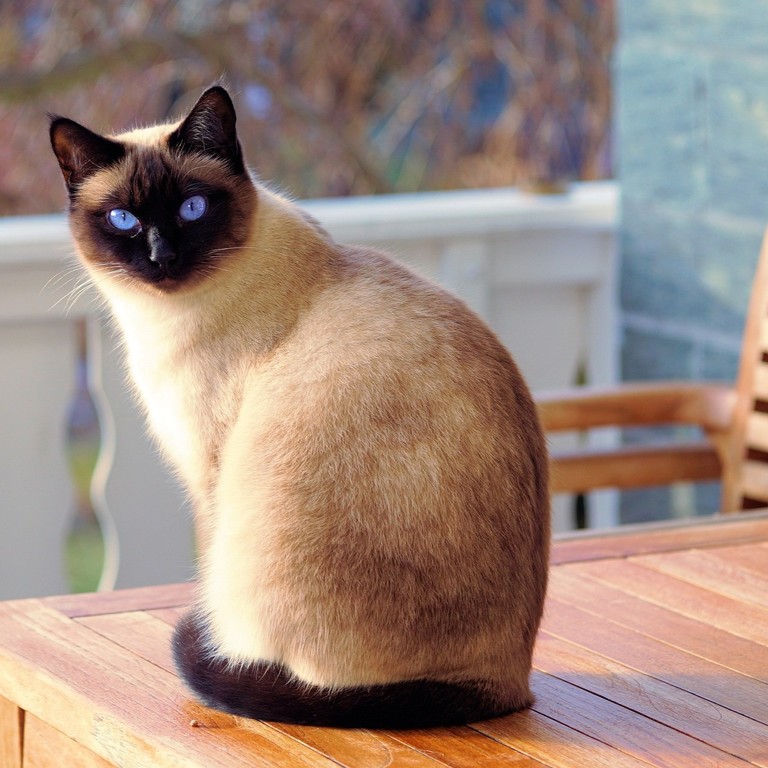} &
        \includegraphics[width=0.132\textwidth,height=0.132\textwidth]{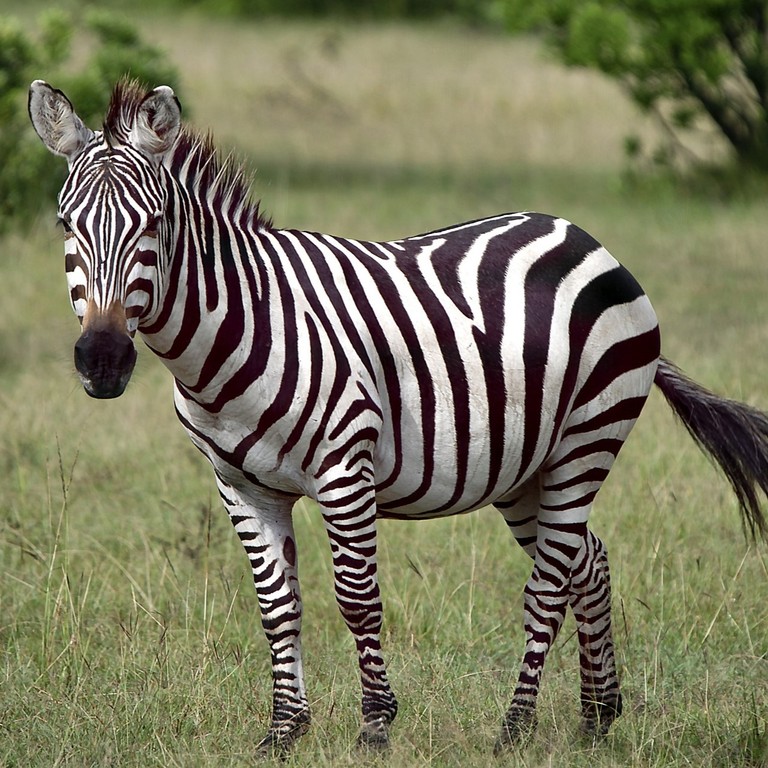} &
        \includegraphics[width=0.132\textwidth,height=0.132\textwidth]{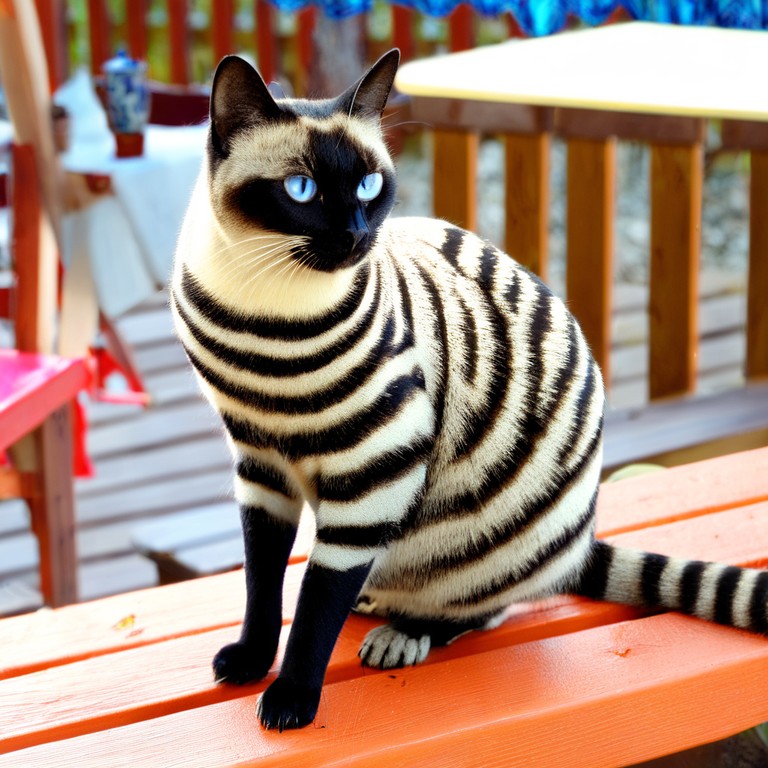} &
        \includegraphics[width=0.132\textwidth,height=0.132\textwidth]{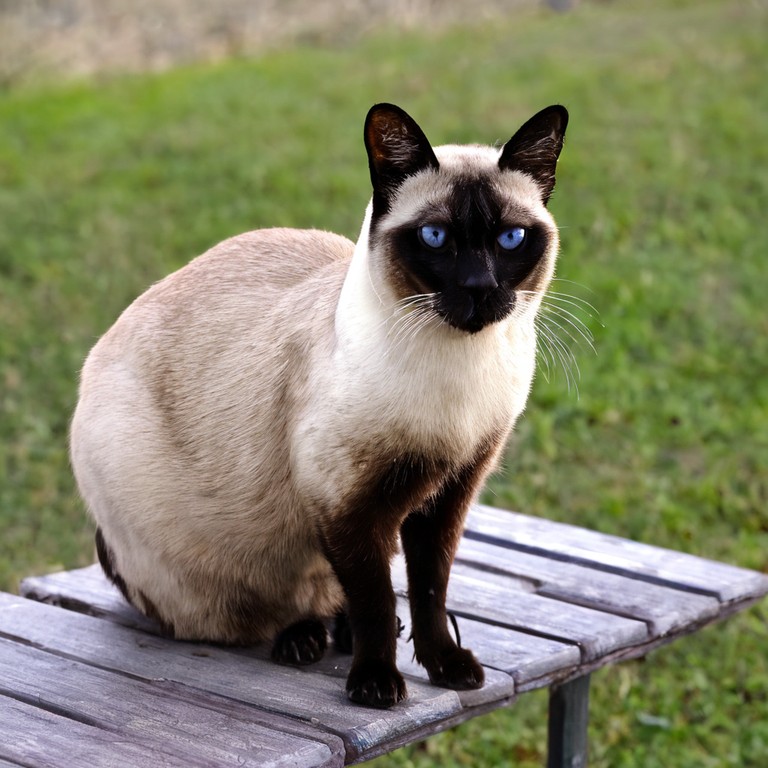} &
        \includegraphics[width=0.132\textwidth,height=0.132\textwidth]{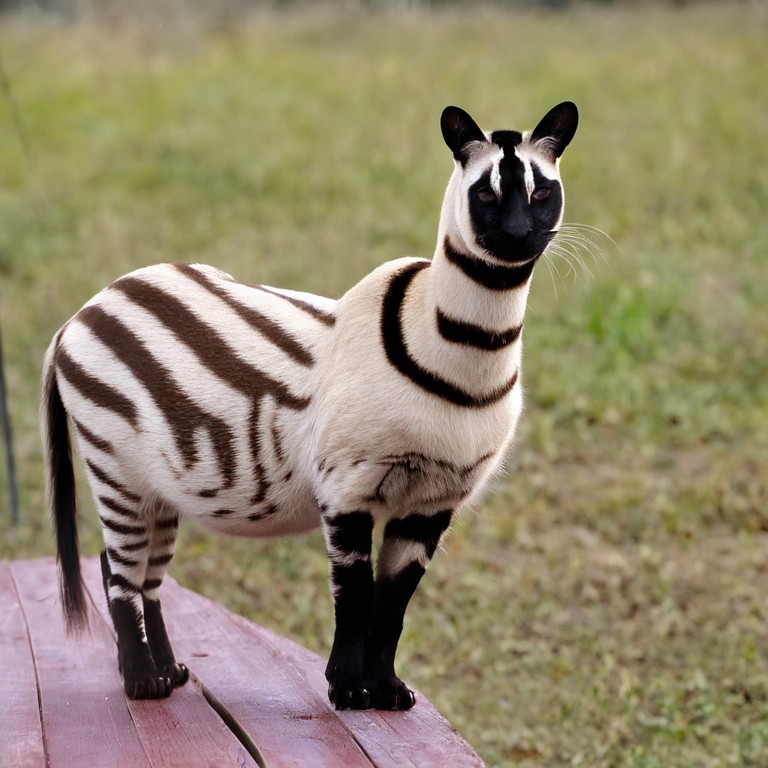} &
        \includegraphics[width=0.132\textwidth,height=0.132\textwidth]{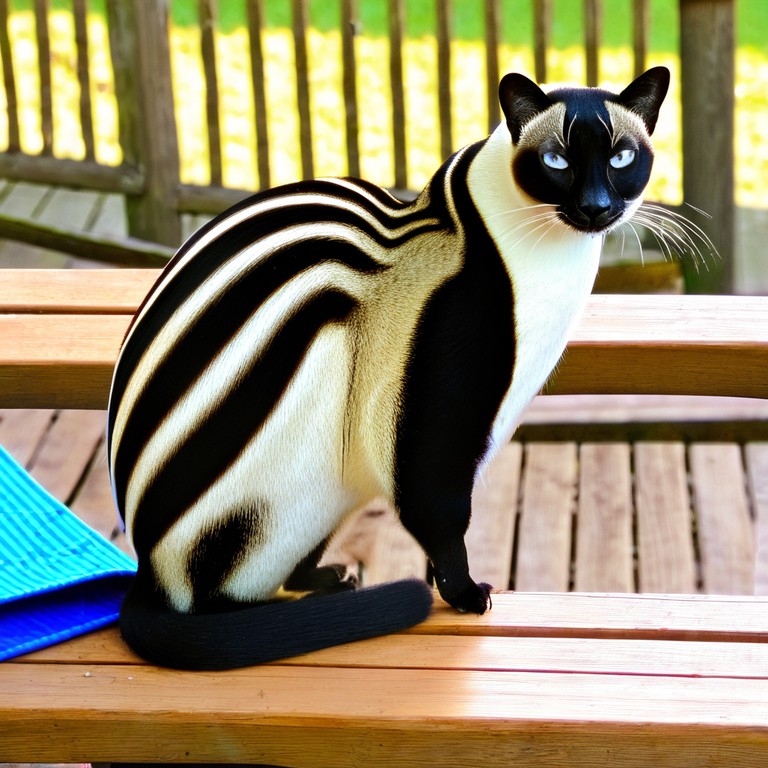} \\
        ``Animal'' & ``Fur'' &  &  &  &  \\
        
        \includegraphics[width=0.132\textwidth,height=0.132\textwidth]{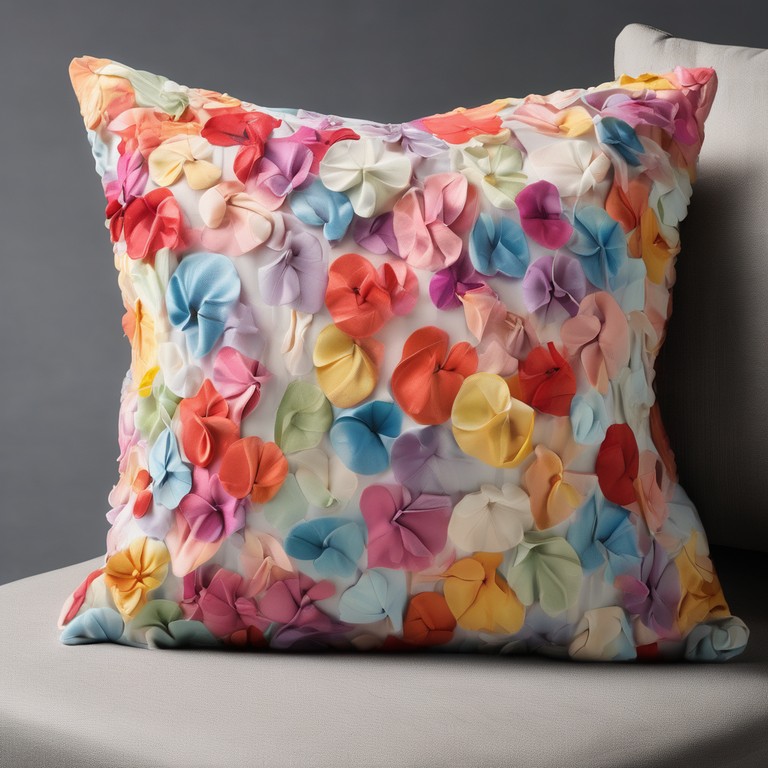} &
        \includegraphics[width=0.132\textwidth,height=0.132\textwidth]{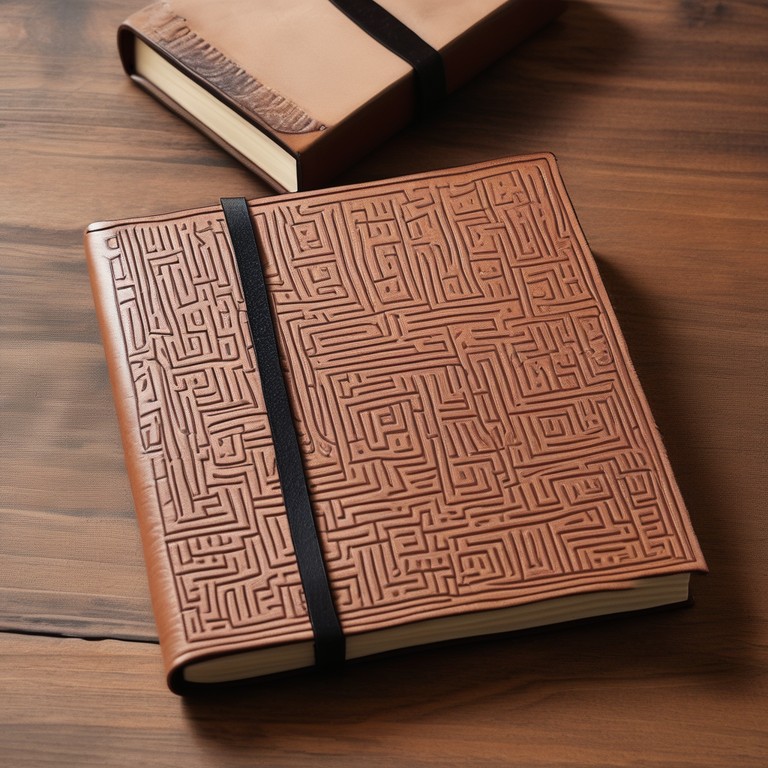} &
        \includegraphics[width=0.132\textwidth,height=0.132\textwidth]{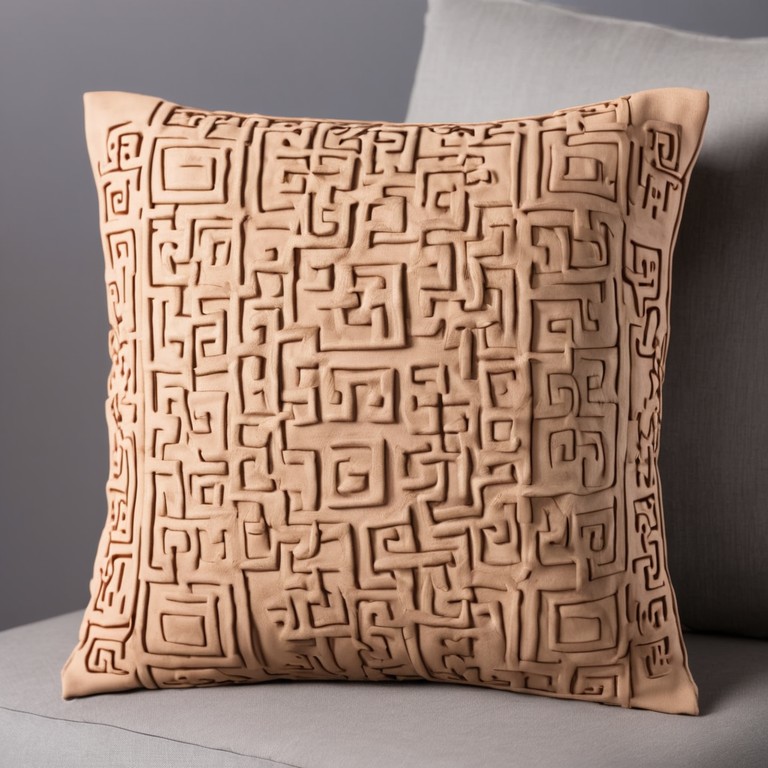} &
        \includegraphics[width=0.132\textwidth,height=0.132\textwidth]{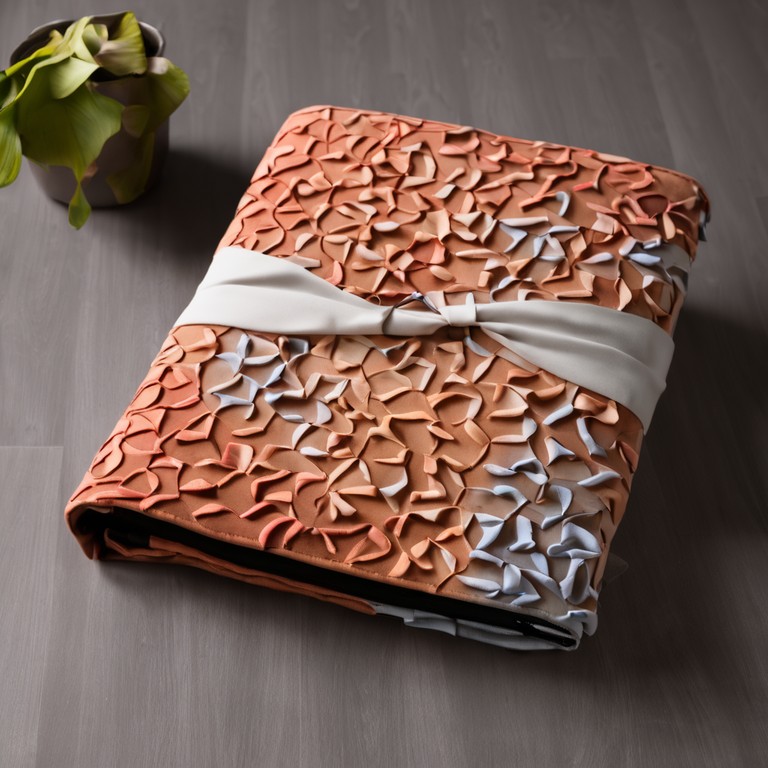} &
        \includegraphics[width=0.132\textwidth,height=0.132\textwidth]{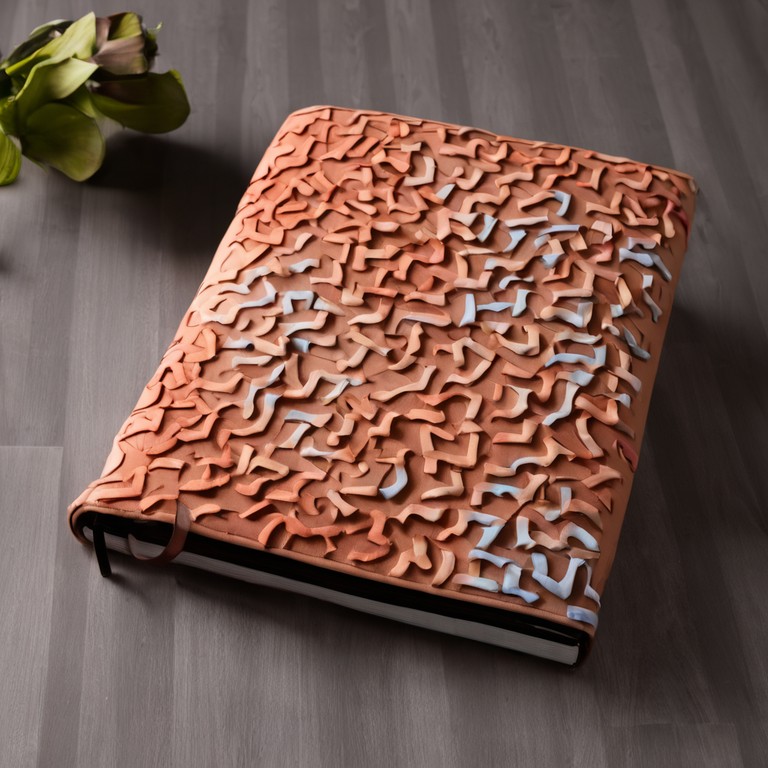} &
        \includegraphics[width=0.132\textwidth,height=0.132\textwidth]{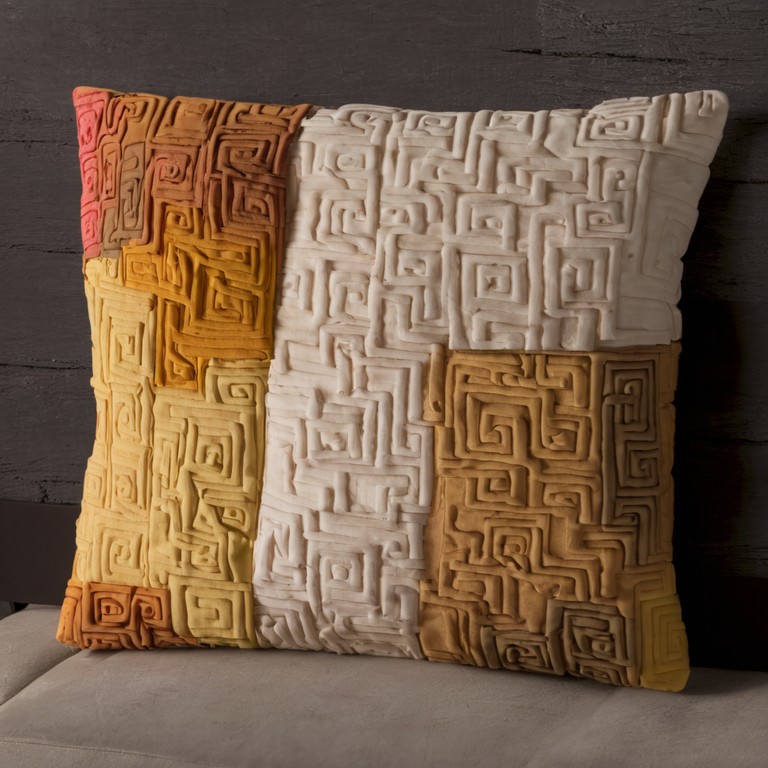} \\
        ``Object'' & ``Pattern'' &  &  &  &  \\
        
    \end{tabular}
    \captionof{figure}{Ablation results. IP-Composer outperforms alternative approaches for combining visual cues using an IP-Adapter backbone, allowing for more accurate concept specification and for reduced leakage.}
    \label{fig:ablation}
\end{table*}

%% file: figures/quantitative_ablation.tex
\begin{figure}
    \begin{tabular}{cc}
        \includegraphics[width=0.45\linewidth,height=0.45\linewidth]{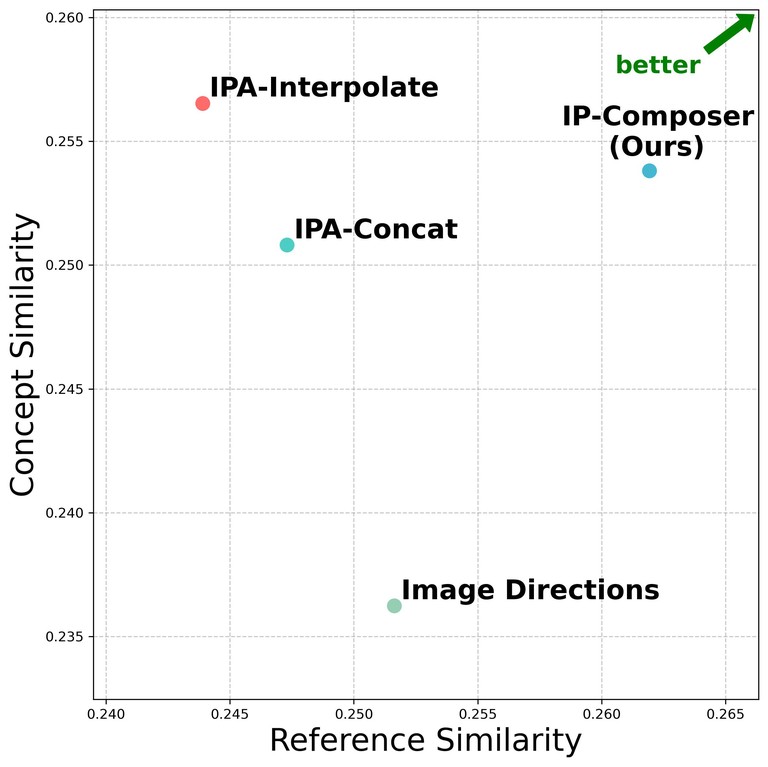} &
        \includegraphics[width=0.45\linewidth,height=0.45\linewidth]{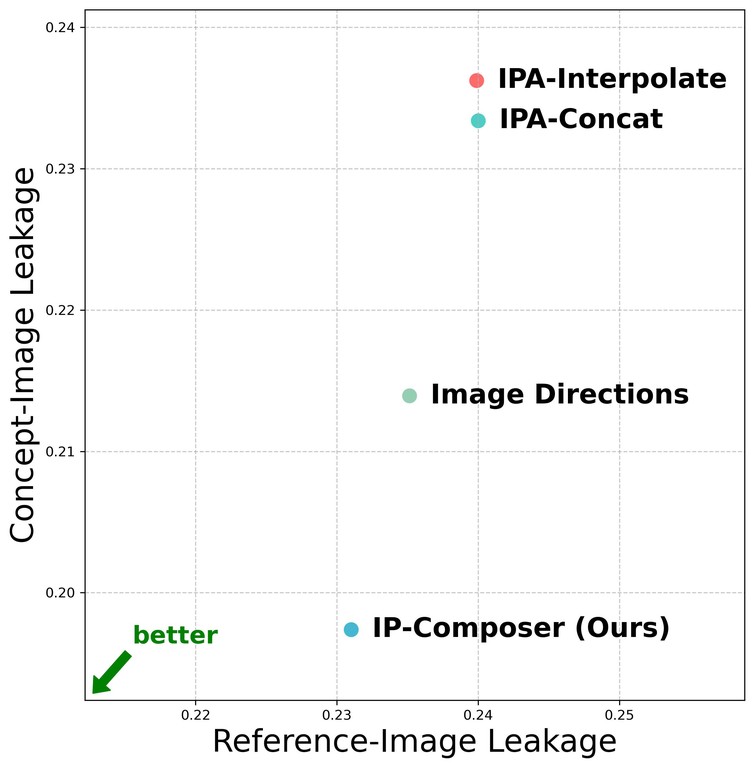} 
    \end{tabular}
    \caption{Quantiative ablation study. Our approach achieves comparable concept similarity to the most performant alternatives, but preserves higher similarity to the reference and has greatly reduced leakage.}\label{fig:quantitative_ablation}
\end{figure}

%% file: figures/multi_step.tex
\begin{table}[htbp]
    \centering
    \setlength{\tabcolsep}{3pt}
    \small
    \begin{tabular}{ccc:cc}
        
        Reference & Concept 1 & Concept 2 & One & Multiple  \\
        Image & Image & Image &  Step & Steps \\

        \includegraphics[width=0.185\linewidth,height=0.185\linewidth]{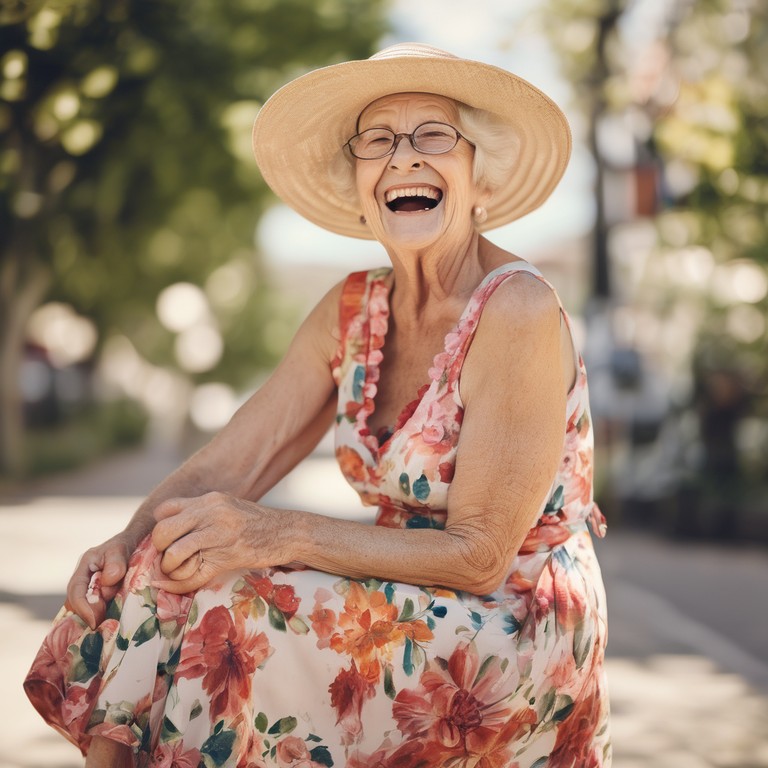} &
        \includegraphics[width=0.185\linewidth,height=0.185\linewidth]{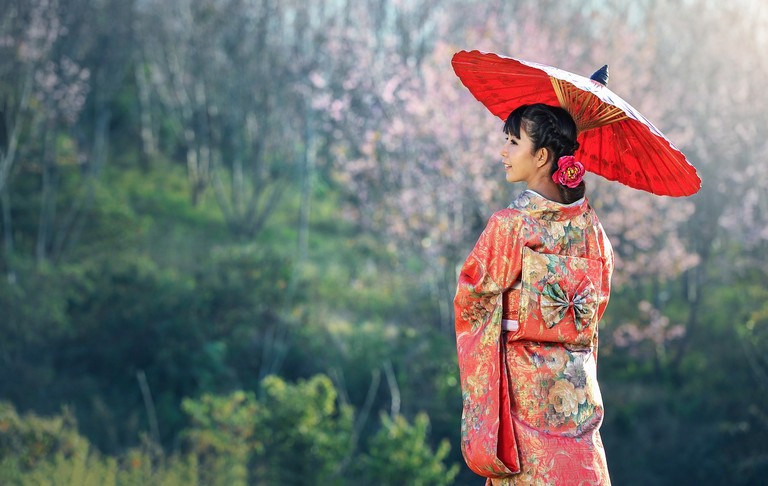} &
        \includegraphics[width=0.185\linewidth,height=0.185\linewidth]{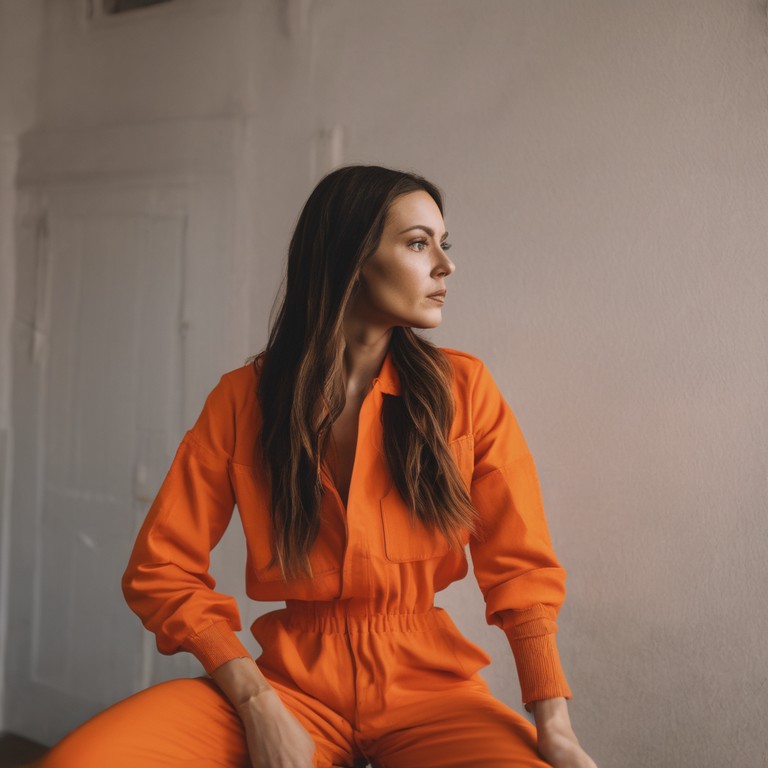} &
        \includegraphics[width=0.185\linewidth,height=0.185\linewidth]{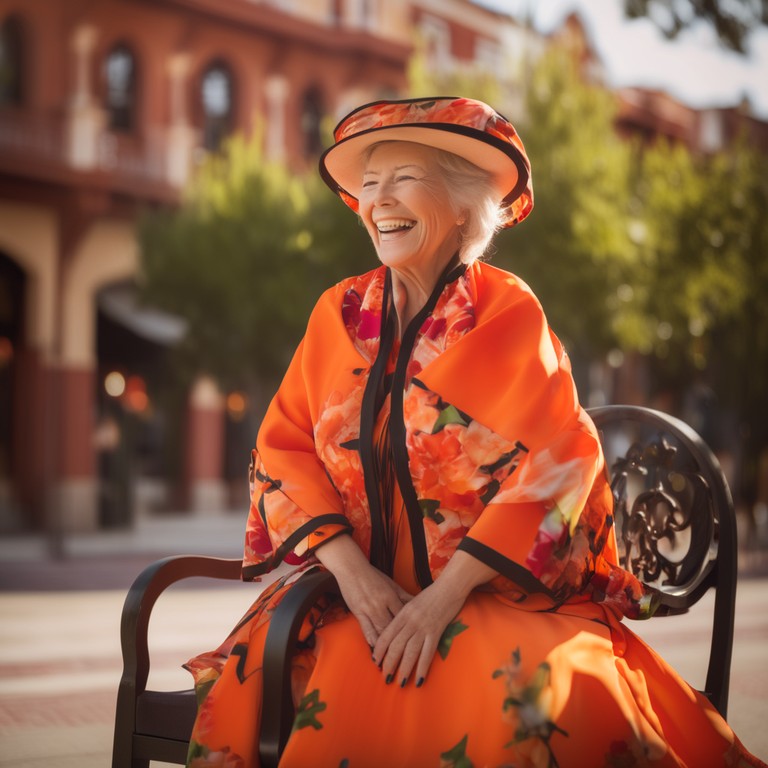} &
        \includegraphics[width=0.185\linewidth,height=0.185\linewidth]{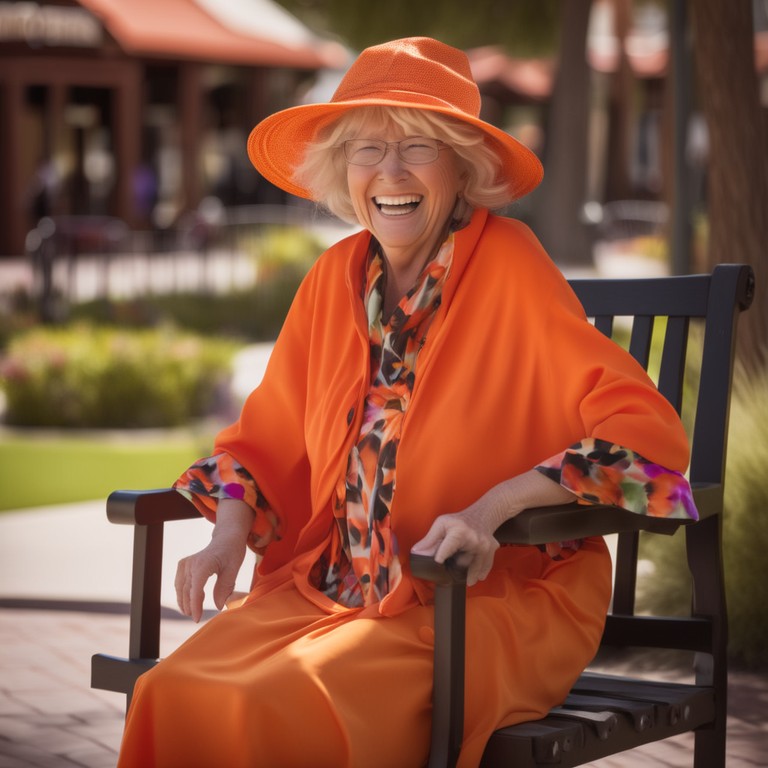} \\
        ``Person'' & ``Outfit'' & ``Color'' &  & \\

        \includegraphics[width=0.185\linewidth,height=0.185\linewidth]{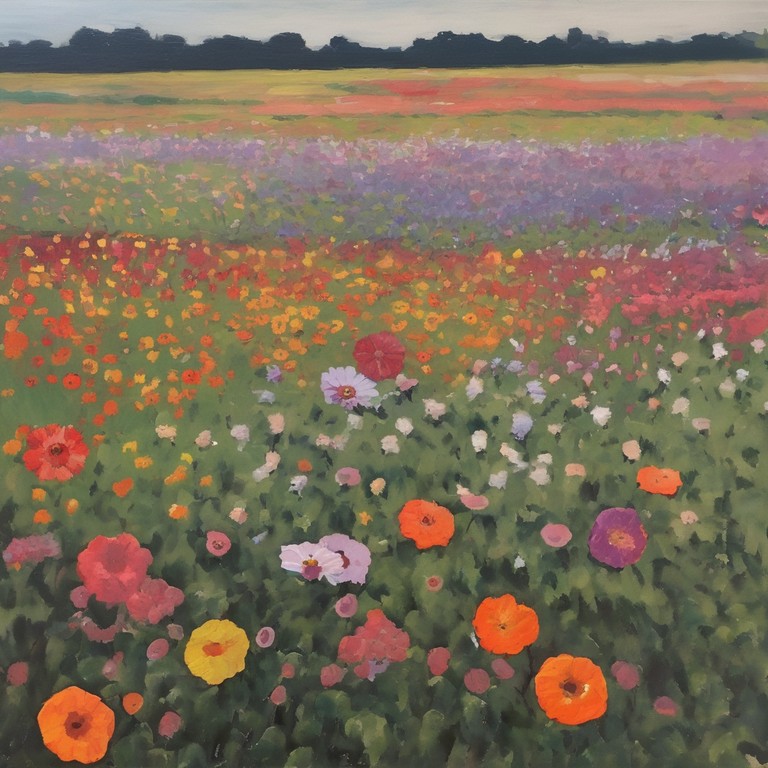} &
        \includegraphics[width=0.185\linewidth,height=0.185\linewidth]{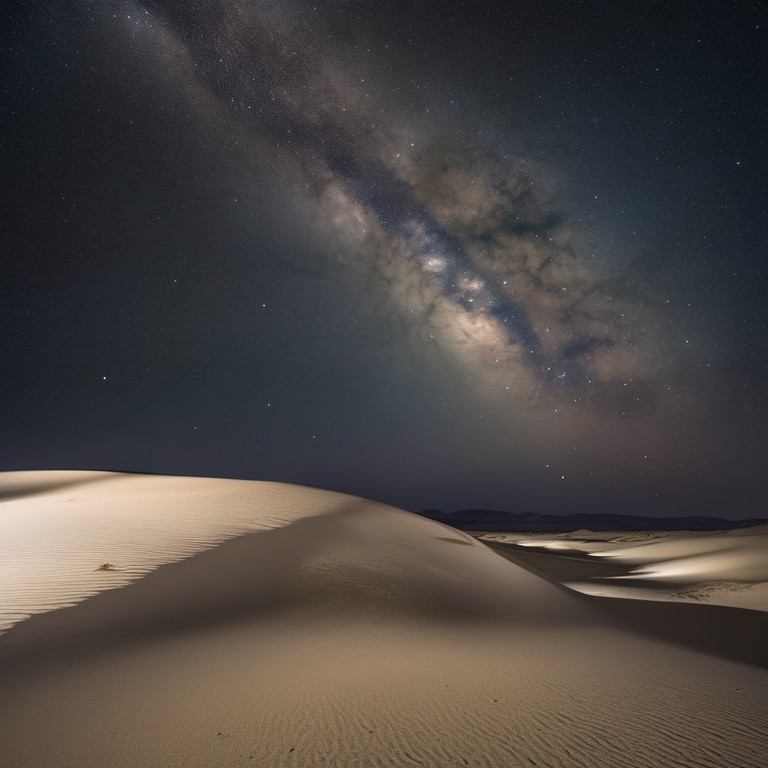} &
        \includegraphics[width=0.185\linewidth,height=0.185\linewidth]{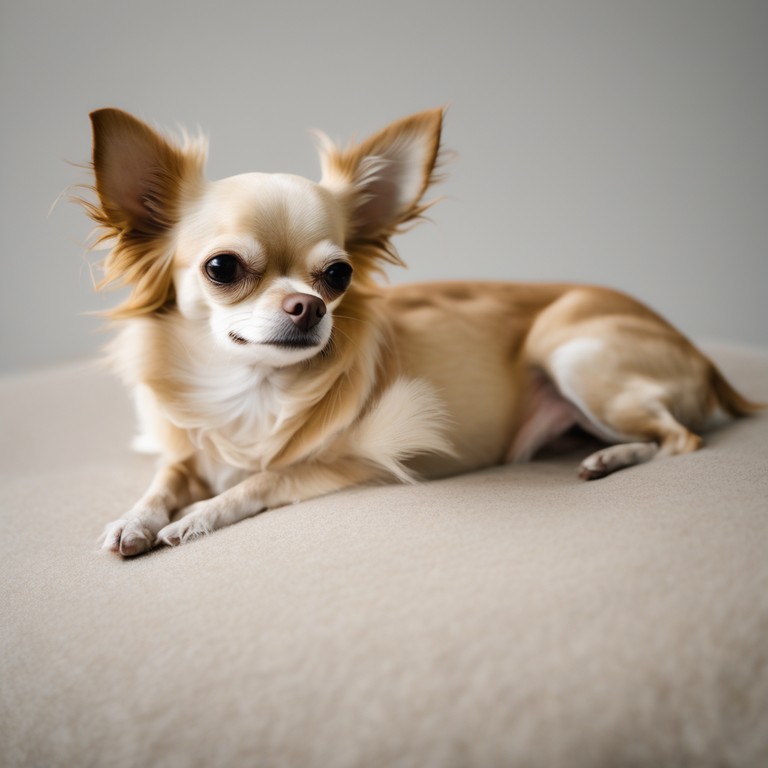} &
        \includegraphics[width=0.185\linewidth,height=0.185\linewidth]{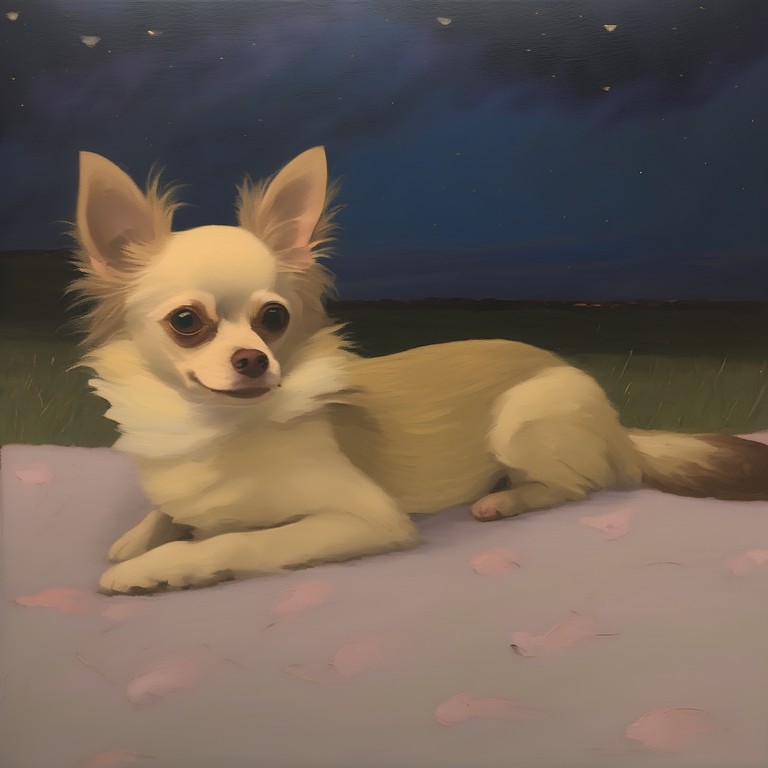} &
        \includegraphics[width=0.185\linewidth,height=0.185\linewidth]{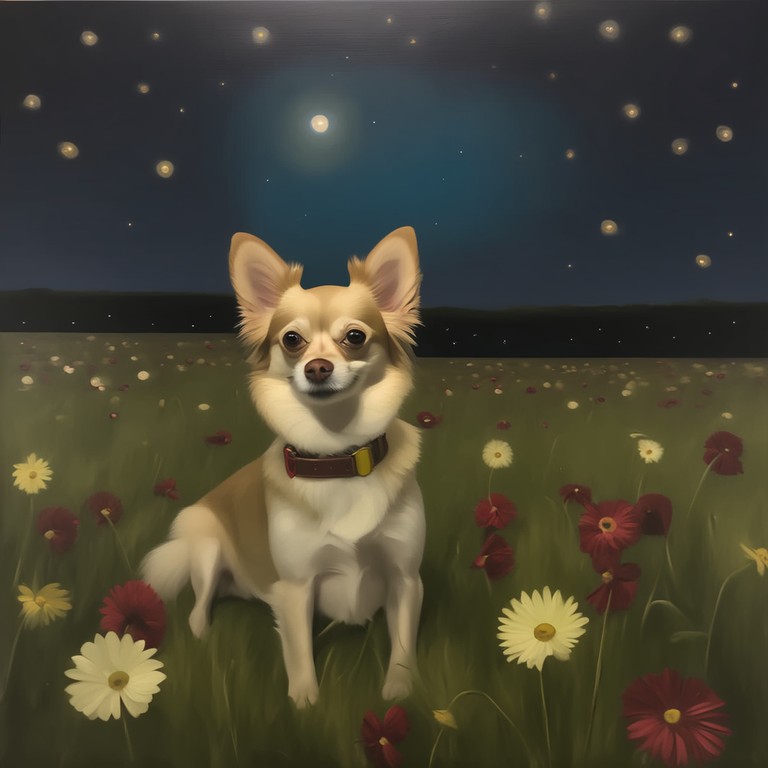} \\
        ``Background'' & ``Lighting'' &  ``Dog'' &  &

    \end{tabular}
    \captionof{figure}{When combining more than two concepts, we can either join them two at-a-time and generate an image to use as the reference for the next step, or combine many concepts at once. In some cases, the multi-step approach can reduce leakage of undesired features. However, it also increases the possibility of losing details from the input images, harming the end result.}
    \label{fig:multi_step}
\end{table}

%% file: 5_conclusions.tex
\input{figures/limitations}

\section{Limitations}

While our approach is typically more general than current training-based approaches, it still has limitations. One limitation arises from surprising entanglements in the CLIP and diffusion feature spaces. For example, when attempting to combine a zebra's body with a leopard fur pattern (\cref{fig:limitations} (top)), the diffusion model tends to produce animals with the head of a giraffe, even though no giraffe appears in either input image. We hypothesize that this may be related to the tendency of diffusion models to represent some concepts as a composition of more basic visual components~\citep{chefer2023hidden}, but leave further investigation to future work.

On the other hand, some concepts may be \textit{more} disentangled in CLIP-space than intuitively expected. For example, outfit types and colors are disentangled in CLIP-space, hence, an ``outfit'' subspace spanned with descriptions of different types of outfits (``dress'', ``tuxedo''...) will not preserve outfit colors (\cref{fig:limitations} (bottom)). However, this can be easily amended by also specifying colors in the spanning texts (``\textit{red} dress'', ``\textit{blue} tuxedo''...).

Finally, we note that IP-Adapter itself is limited in the level of detail captured from the input image. Hence, our approach will not be sufficient for capturing delicate details such as exact identities. Stronger encoders may achieve higher fidelity, but it is not clear that our embedding-space projections would generalize to more complex feature spaces.

\section{Conclusions}

We presented IP-Composer, a training-free method that allows a user to compose novel images from visual concepts derived through a set of input images. To do so, our approach uses a CLIP-based IP-Adapter, leveraging their joint disentangled subspace structure. Through this approach, we achieve comparable or better performance compared with existing training-based methods, and can more easily generalize to novel concepts derived solely from textual descriptions. 

We hope that our work can serve as an additional component of the creative toolbox, and open the way to additional composable-concept discovery methods. 

\section{Acknowledgment}
We would like to thank Ron Mokady and Yoad Tewel for providing feedback and helpful suggestions.

%% file: figures/limitations.tex
\begin{table}[htbp]
    \centering
    \begin{tabular}{ccc}
        \includegraphics[width=0.13\textwidth,height=0.13\textwidth]{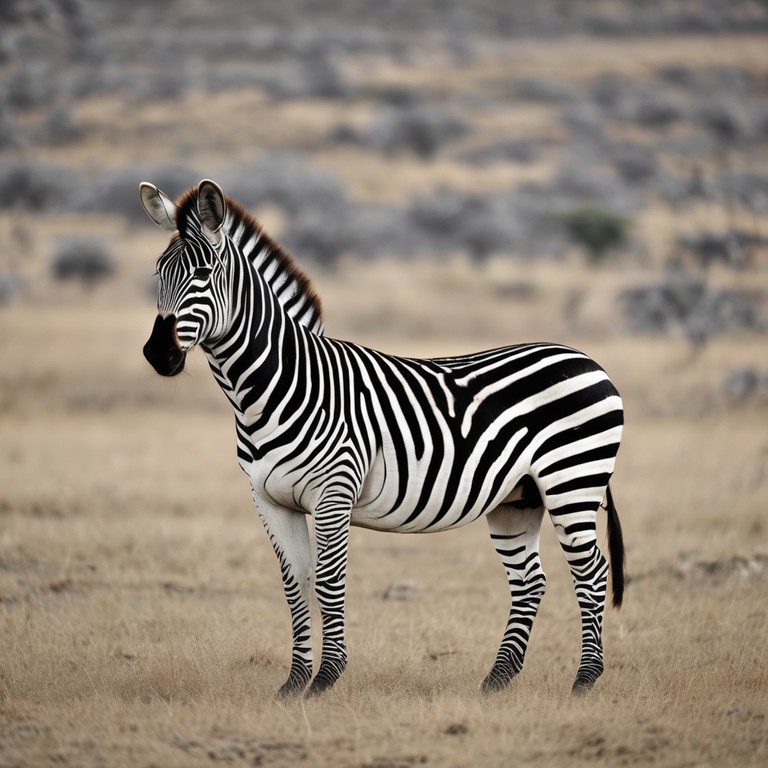} &
        \includegraphics[width=0.13\textwidth,height=0.13\textwidth]{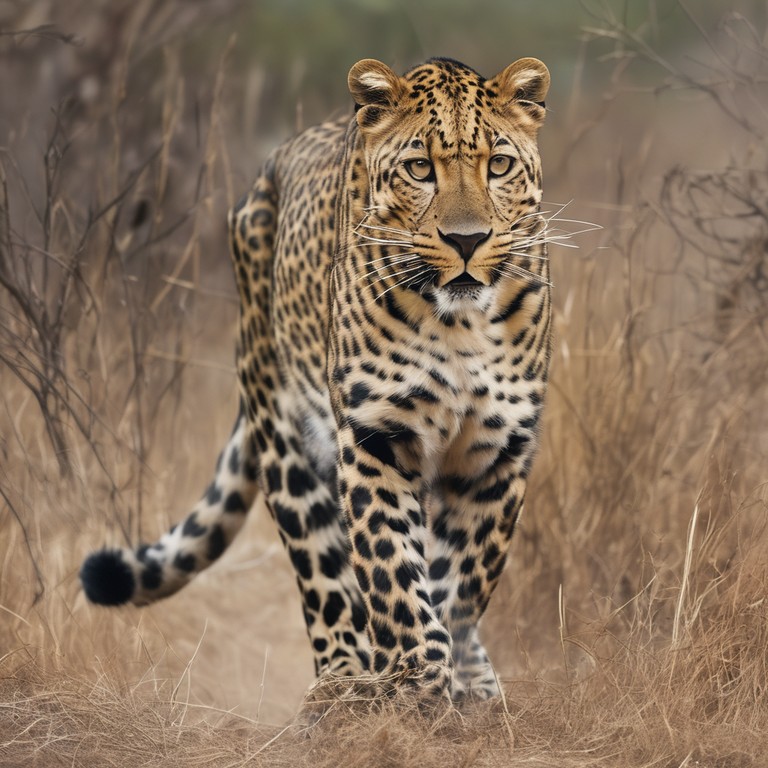} &
        \includegraphics[width=0.13\textwidth,height=0.13\textwidth]{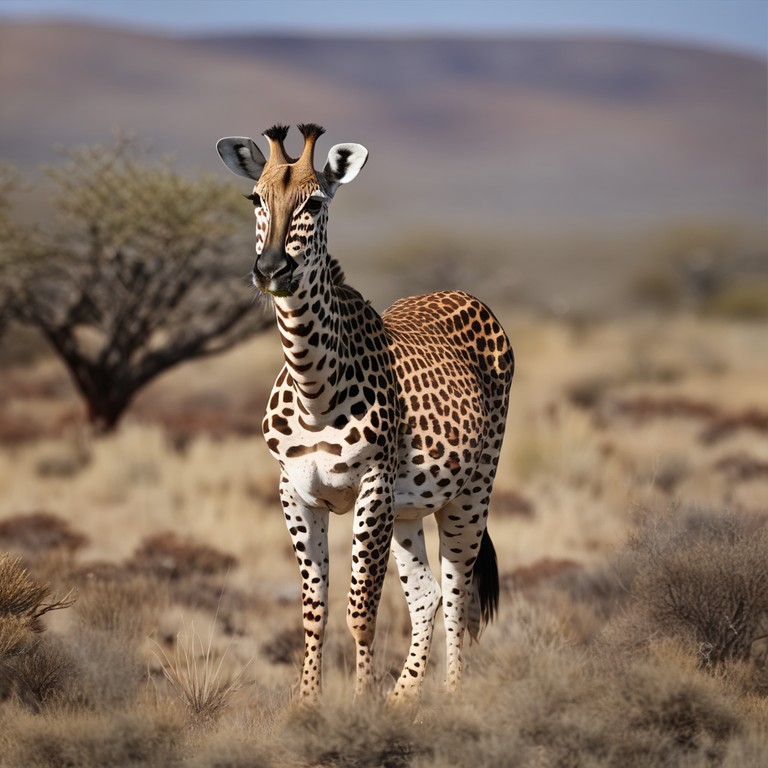} \\
         ``Animal'' & ``Fur'' &  Result \\
        \includegraphics[width=0.13\textwidth,height=0.13\textwidth]{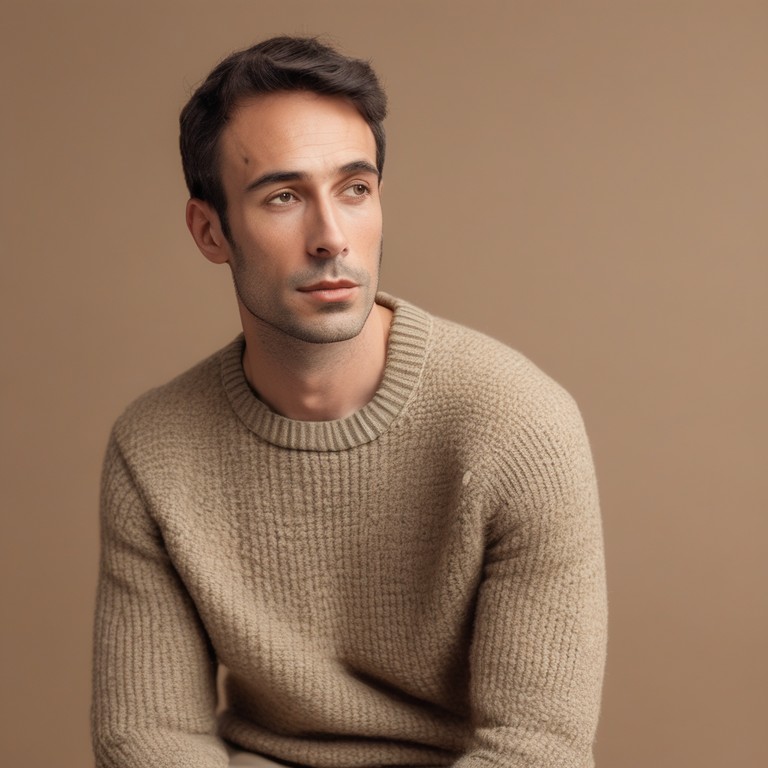} &
        \includegraphics[width=0.13\textwidth,height=0.13\textwidth]{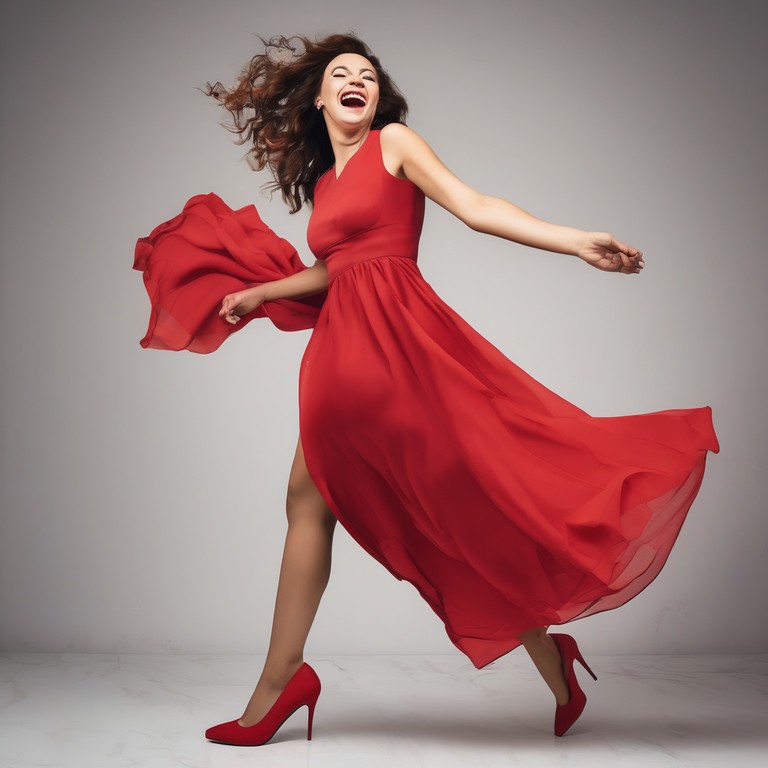} &
        \includegraphics[width=0.13\textwidth,height=0.13\textwidth]{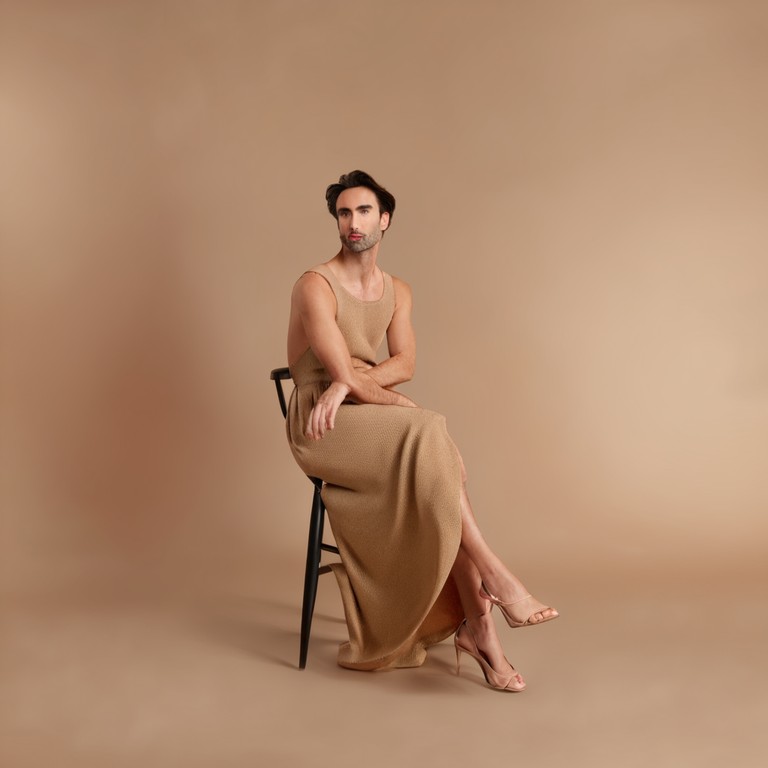} \\
        ``Person'' & ``Outfit'' &  Result
        
    \end{tabular}
    \captionof{figure}{Limitations. Demonstrating the affect of concept entanglement/disentanglement on our method. (Top) When attempting to compose leopard pattern with a zebra's body, the combination may produce giraffe-like features. (Bottom) When using descriptions that only specify outfit style, our method doesn't transfer the outfit color, demonstrating the gap between CLIP-disentanglement and the common intuition. This can be resolved by using more specific concept prompts.}
    \label{fig:limitations}
\end{table}

%% file: figures/additional_qualitative_results_1.tex
\begin{figure*}[t]
    \centering
    \setlength{\belowcaptionskip}{-5pt}
    \setlength{\abovecaptionskip}{4pt}
    
    \begin{tikzpicture}
        \matrix (m1) [matrix of nodes,
            nodes={draw, minimum width=2cm, minimum height=1cm, inner sep=0pt, line width=1.5pt},
            row sep=0.4cm,
            column sep=0.443cm
        ] {
            \includegraphics[width=2cm,height=2cm]{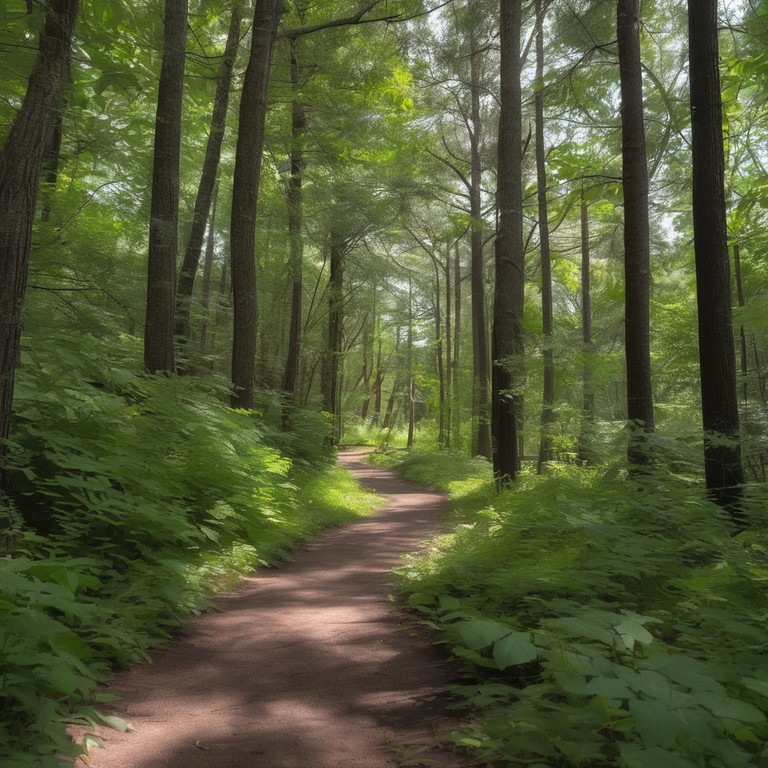} & \includegraphics[width=2cm,height=2cm]{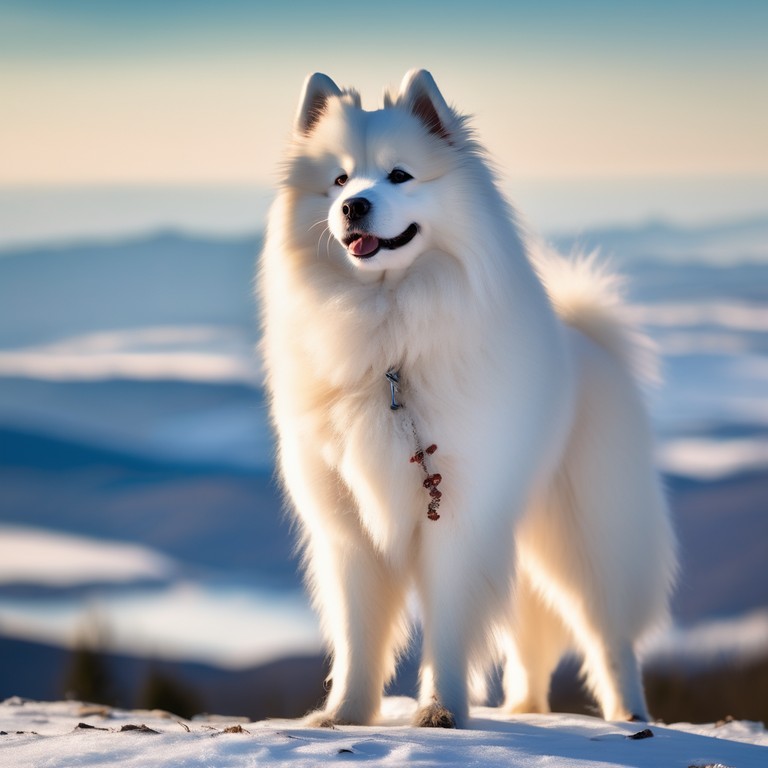} & \includegraphics[width=2cm,height=2cm]{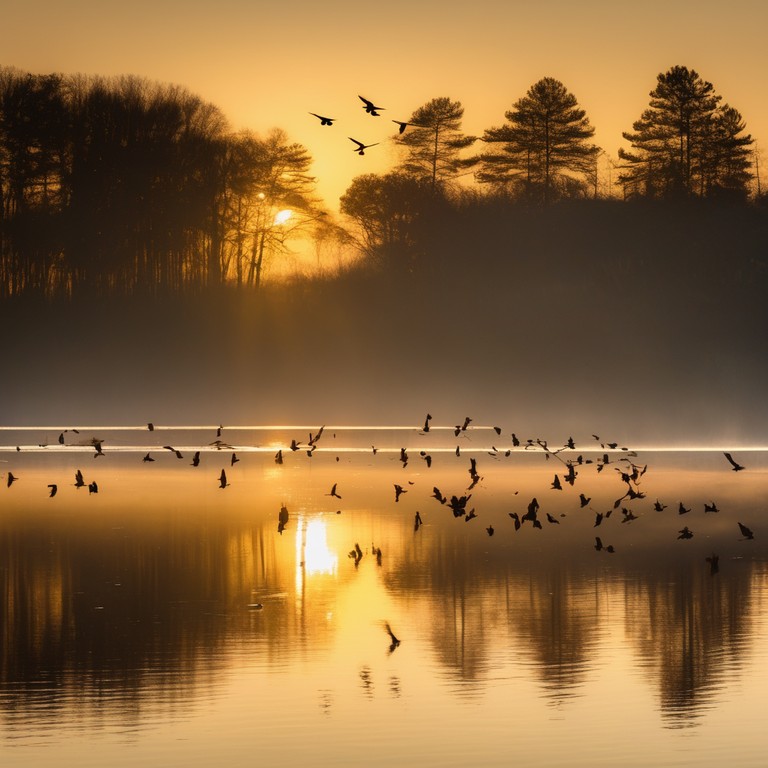} & \includegraphics[width=2cm,height=2cm]{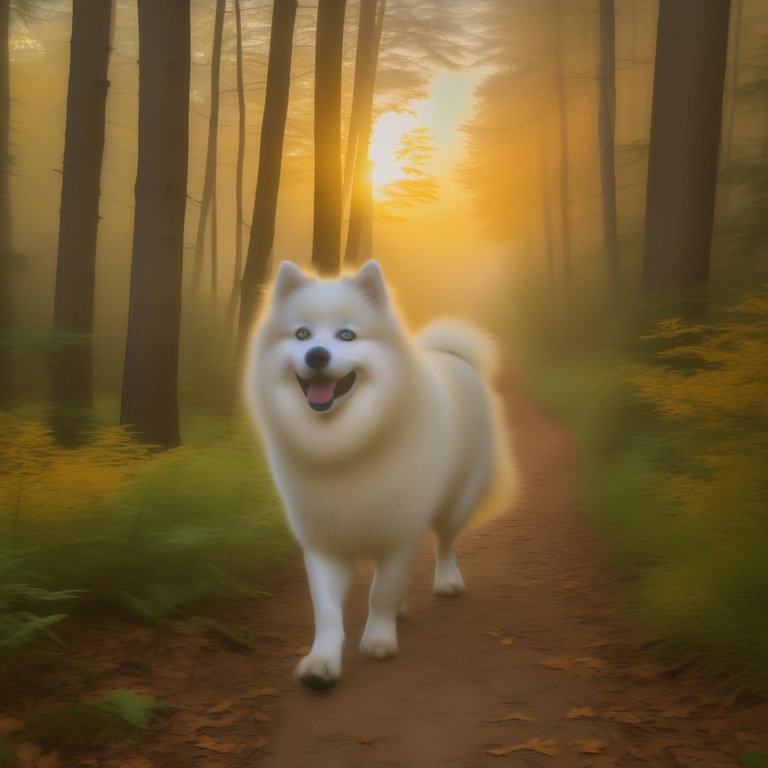} & \includegraphics[width=2cm,height=2cm]{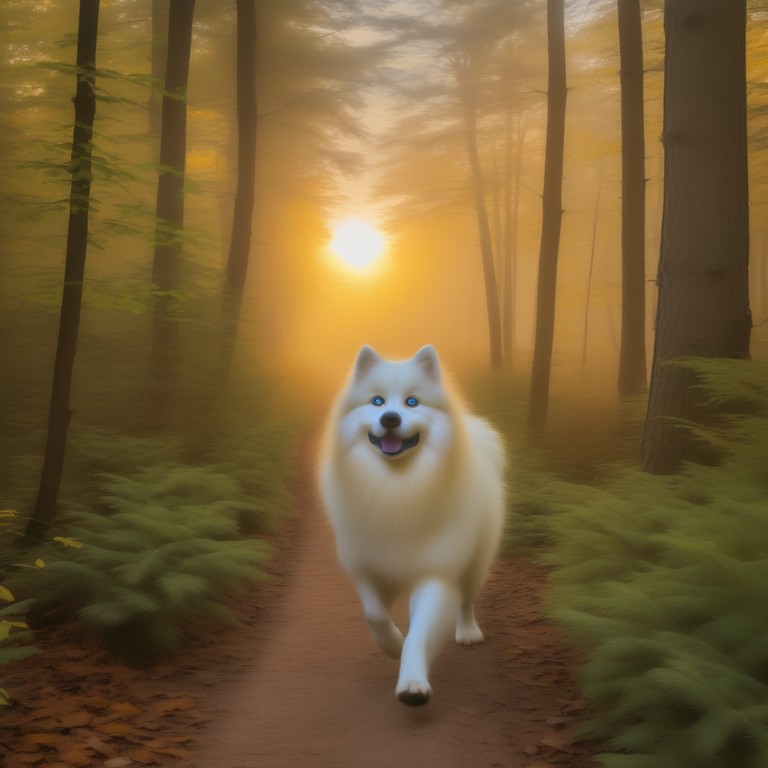} & \includegraphics[width=2cm,height=2cm]{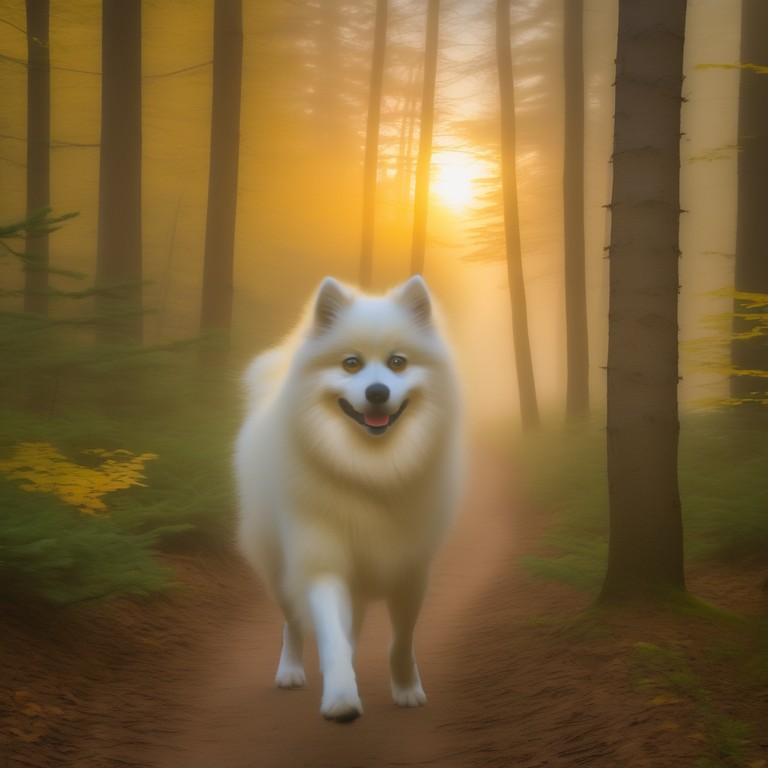} \\
            \includegraphics[width=2cm,height=2cm]{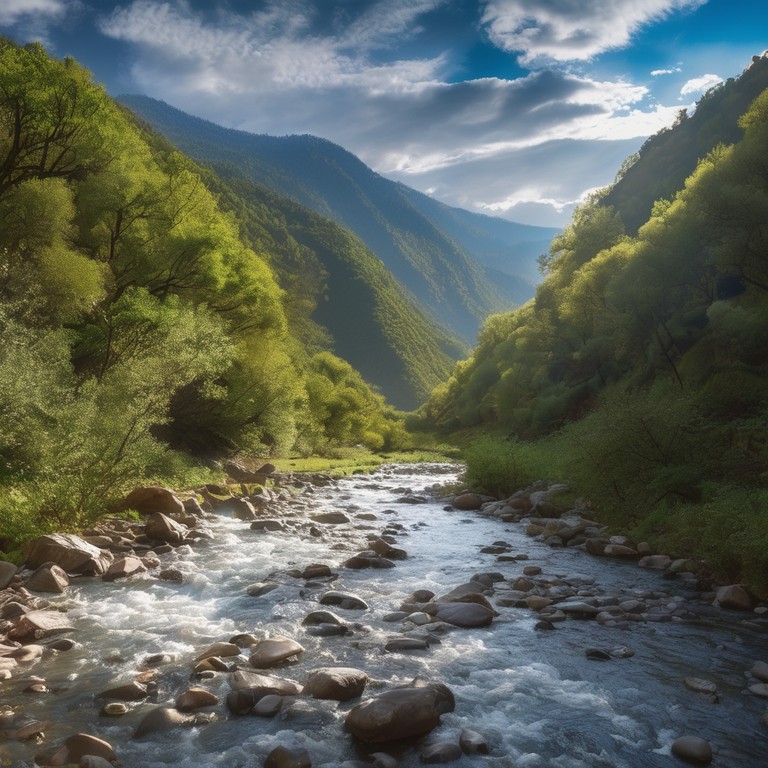} & \includegraphics[width=2cm,height=2cm]{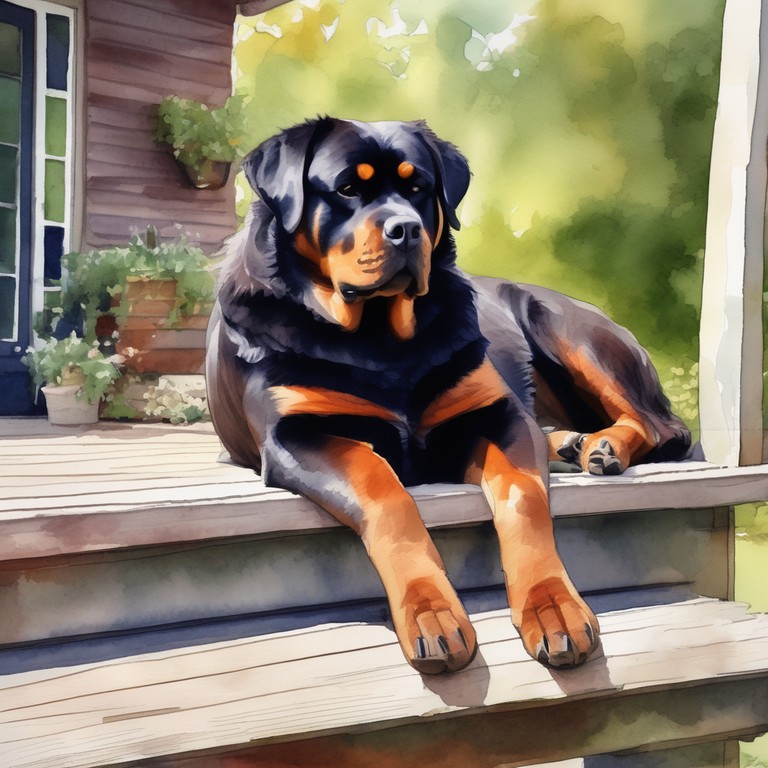} & \includegraphics[width=2cm,height=2cm]{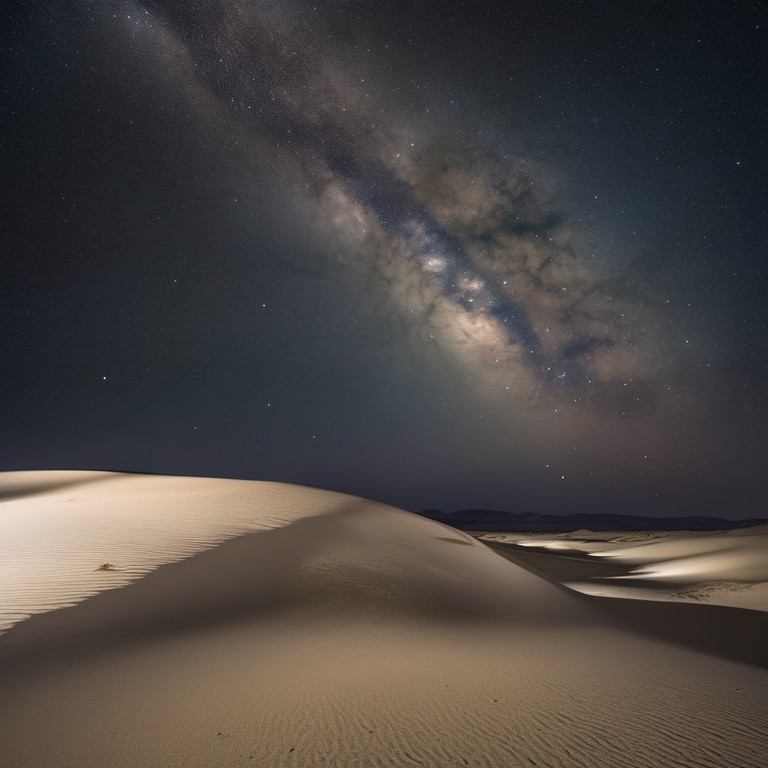} & \includegraphics[width=2cm,height=2cm]{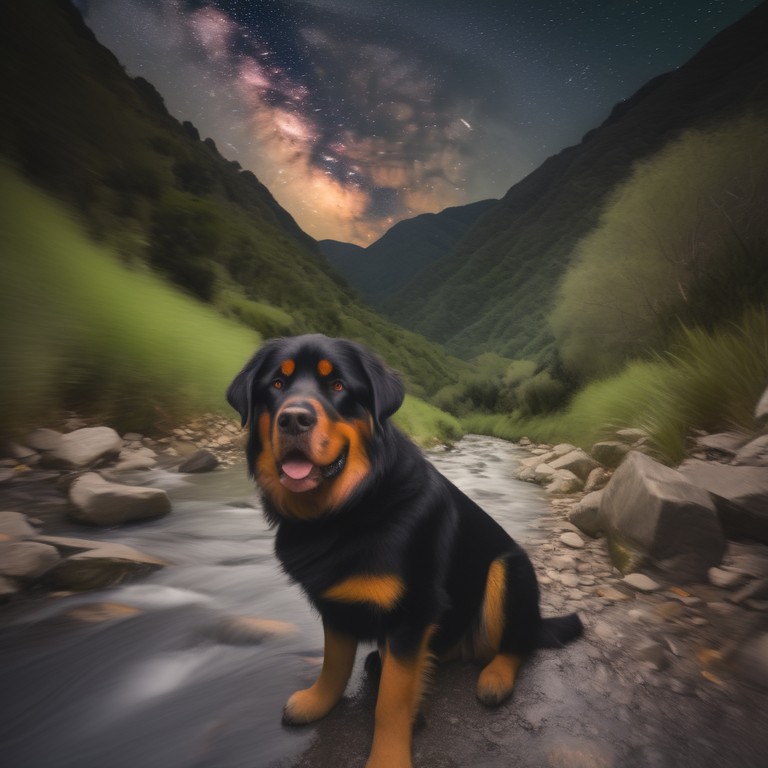} & \includegraphics[width=2cm,height=2cm]{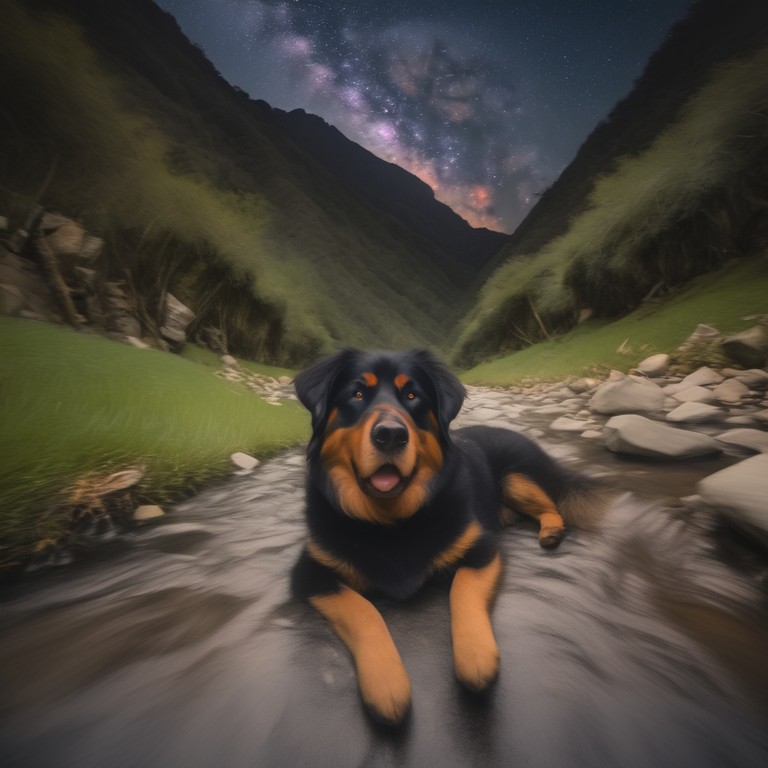} & \includegraphics[width=2cm,height=2cm]{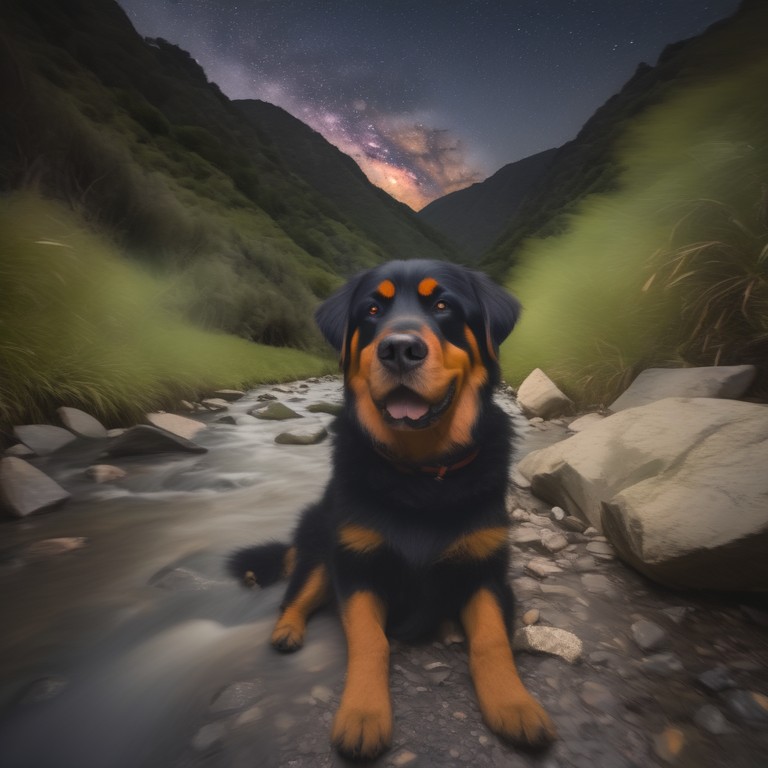} \\
            \includegraphics[width=2cm,height=2cm]{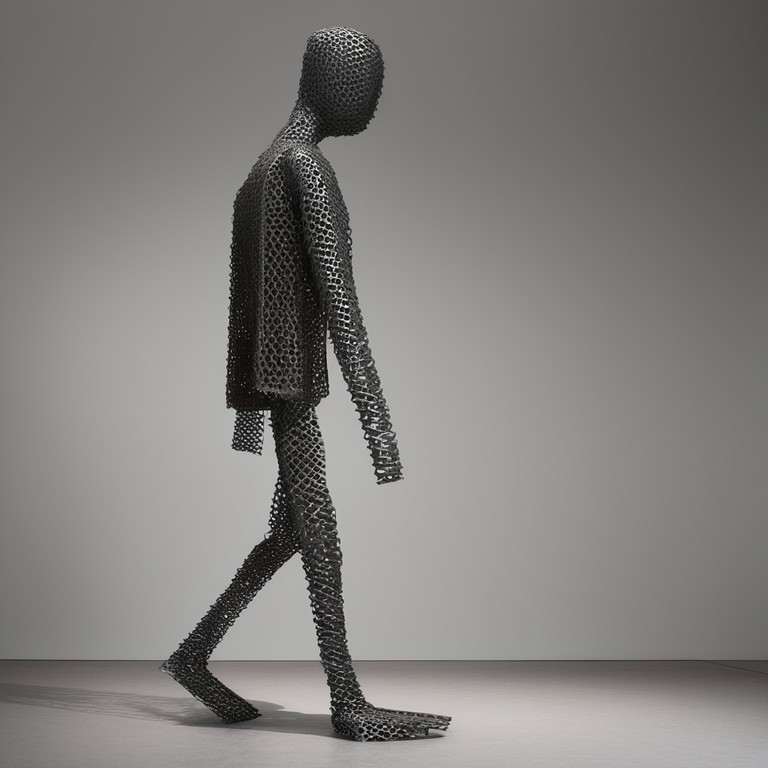} & \includegraphics[width=2cm,height=2cm]{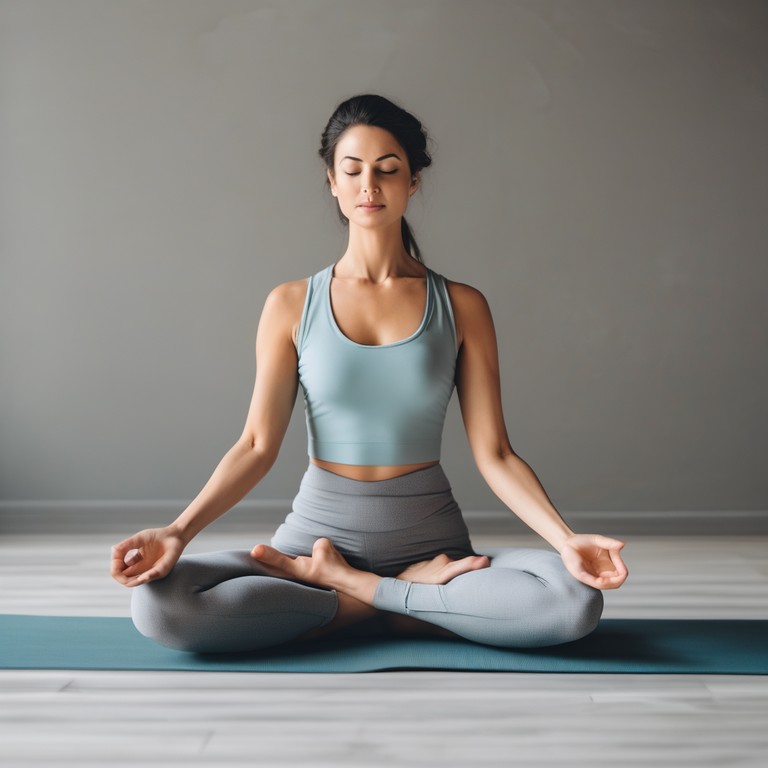} & \includegraphics[width=2cm,height=2cm]{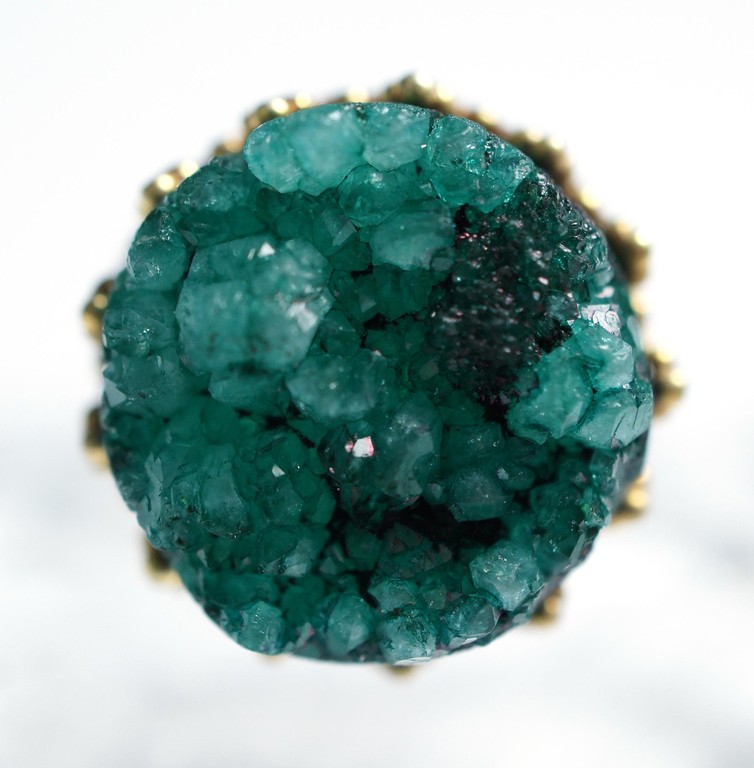} & \includegraphics[width=2cm,height=2cm]{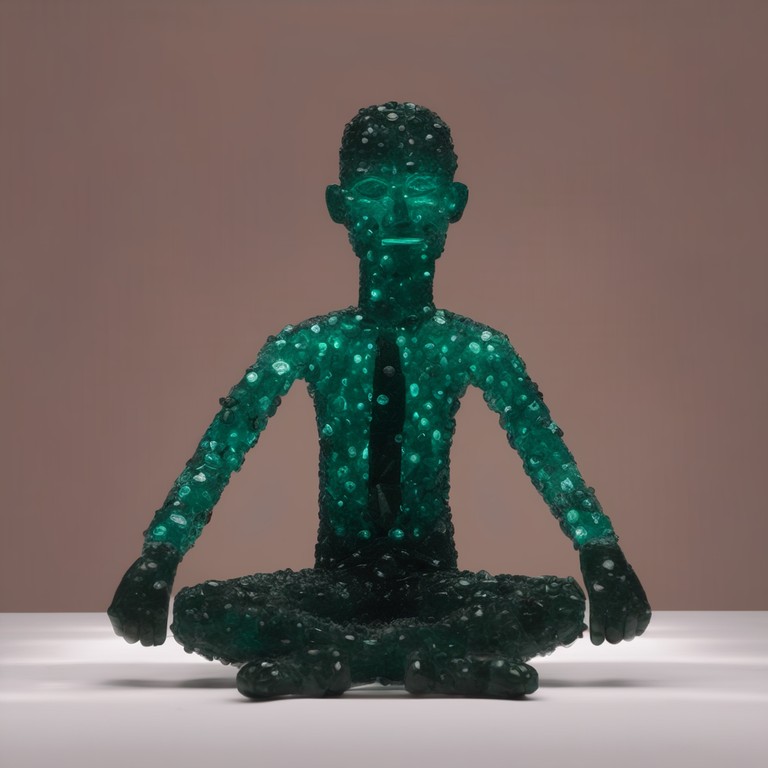} & \includegraphics[width=2cm,height=2cm]{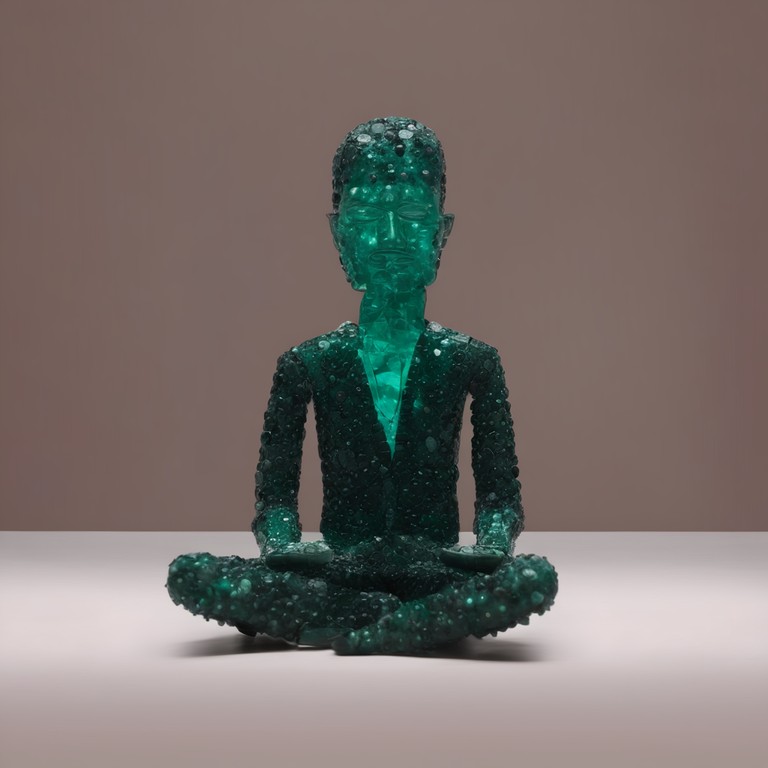} & \includegraphics[width=2cm,height=2cm]{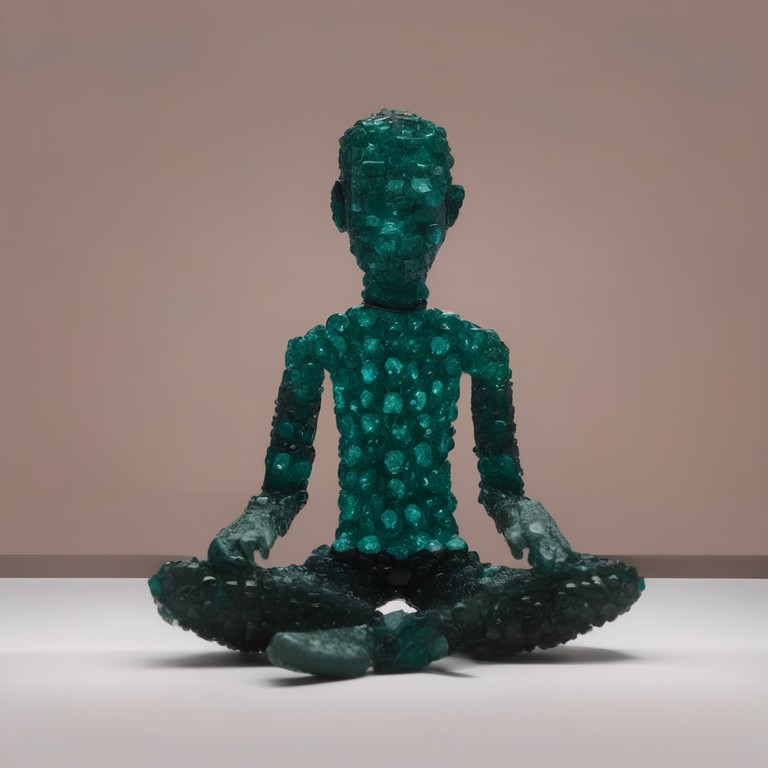} \\
            \includegraphics[width=2cm,height=2cm]{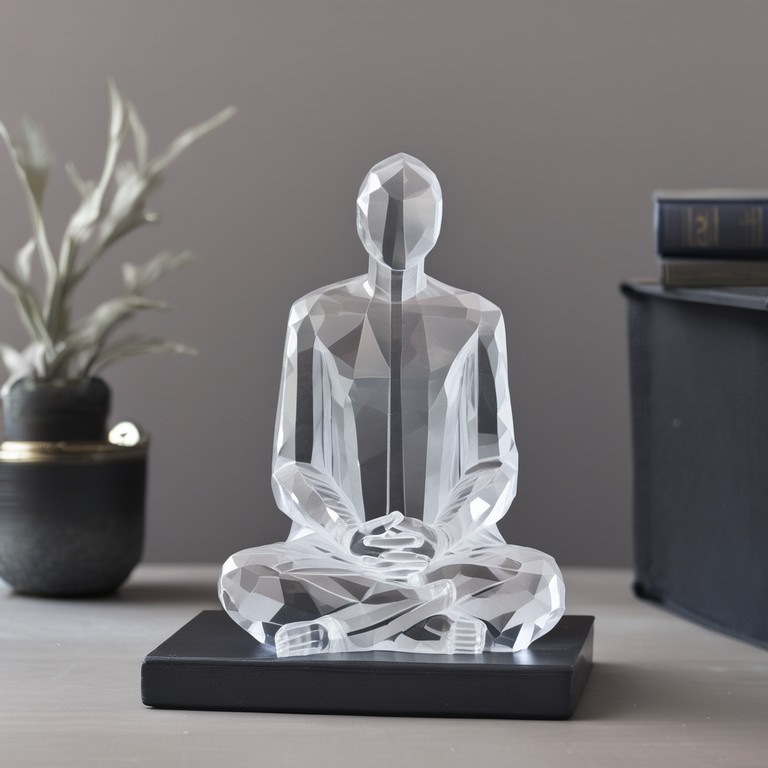} & \includegraphics[width=2cm,height=2cm]{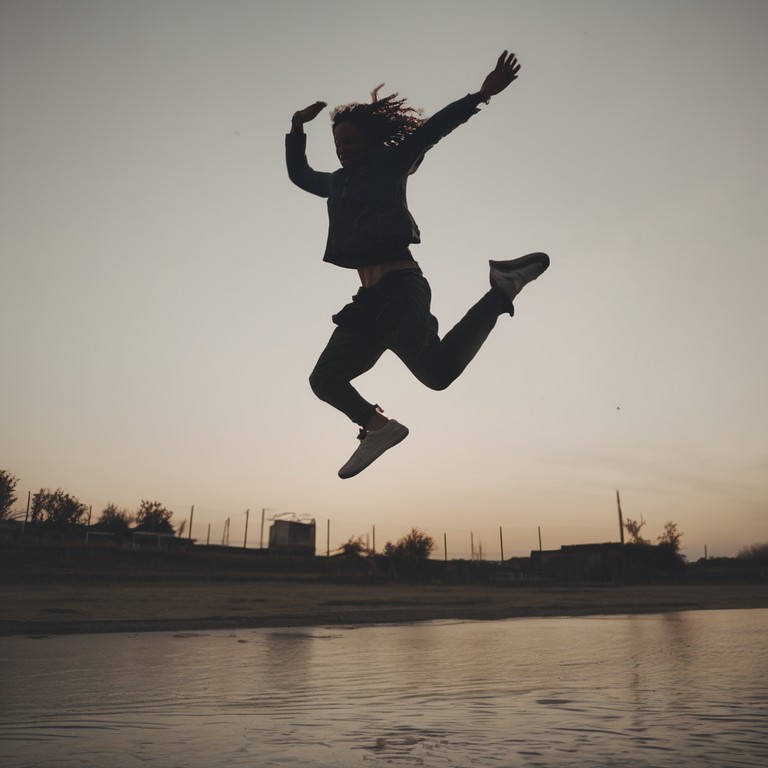} & \includegraphics[width=2cm,height=2cm]{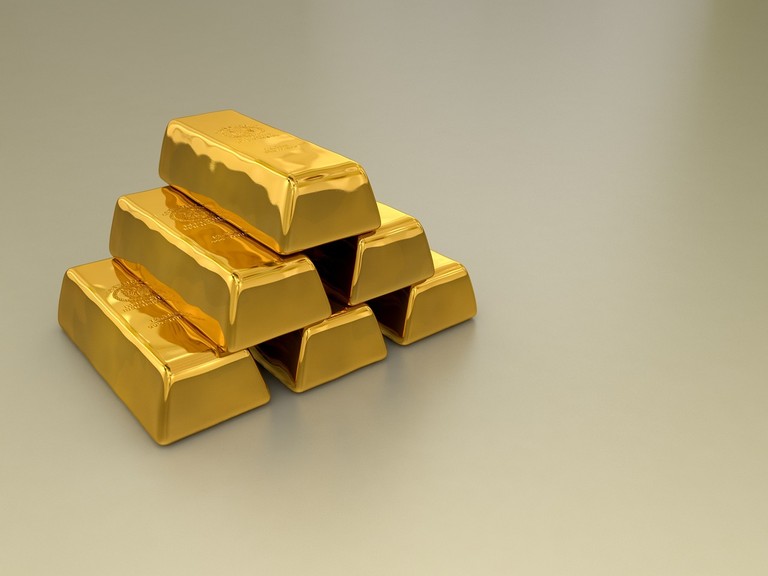} & \includegraphics[width=2cm,height=2cm]{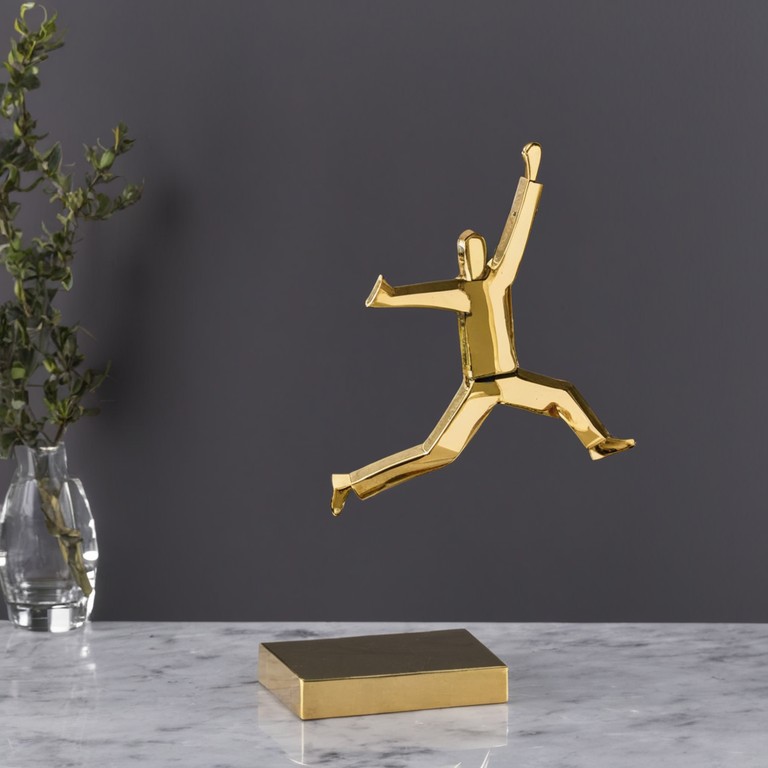} & \includegraphics[width=2cm,height=2cm]{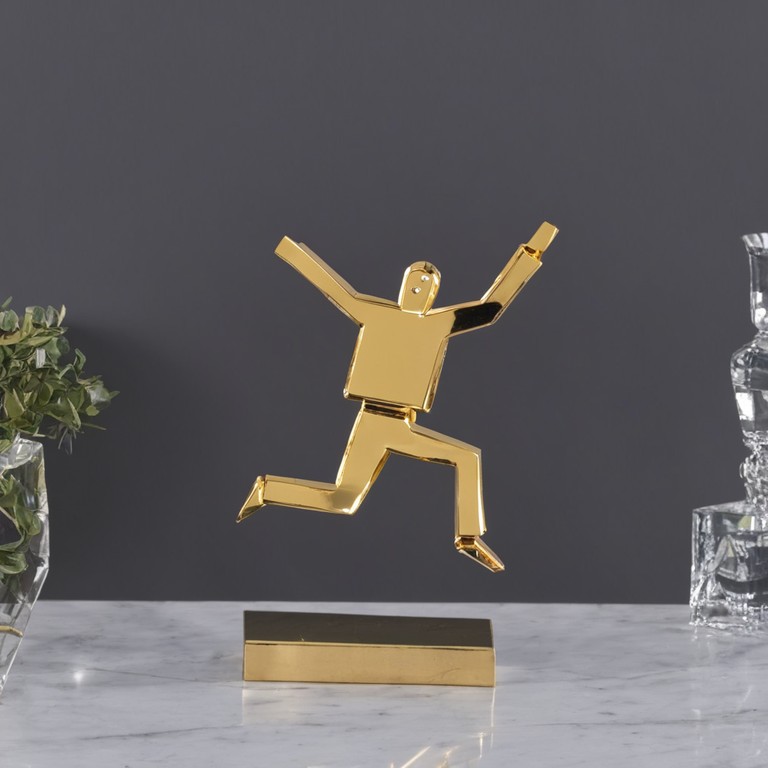} & \includegraphics[width=2cm,height=2cm]{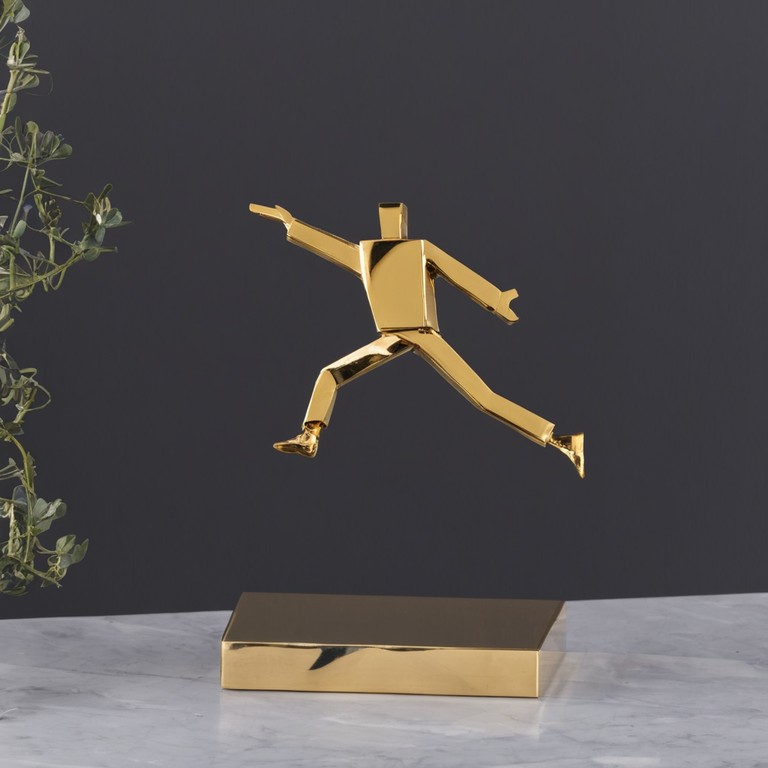} \\
        };

        \path (m1-1-1) -- (m1-1-2) node[midway] {\Large $+$};
        \path (m1-1-2) -- (m1-1-3) node[midway] {\Large $+$};
        \path (m1-1-3) -- (m1-1-4) node[midway] {\Large $=$};

        \path (m1-2-1) -- (m1-2-2) node[midway] {\Large $+$};
        \path (m1-2-2) -- (m1-2-3) node[midway] {\Large $+$};
        \path (m1-2-3) -- (m1-2-4) node[midway] {\Large $=$};

        \path (m1-3-1) -- (m1-3-2) node[midway] {\Large $+$};
        \path (m1-3-2) -- (m1-3-3) node[midway] {\Large $+$};
        \path (m1-3-3) -- (m1-3-4) node[midway] {\Large $=$};

        \path (m1-4-1) -- (m1-4-2) node[midway] {\Large $+$};
        \path (m1-4-2) -- (m1-4-3) node[midway] {\Large $+$};
        \path (m1-4-3) -- (m1-4-4) node[midway] {\Large $=$};

        \node[below=-0.05cm of m1-2-1] {\small ``Scene''};
        \node[below=-0.05cm of m1-2-2] {\small ``Dog''};
        \node[below=-0.05cm of m1-2-3] {\small ``Light''};
        \node[below=-0.05cm of m1-2-5] {\small Results};

        \node[below=-0.05cm of m1-4-1] {\small ``Object''};
        \node[below=-0.05cm of m1-4-2] {\small ``Pose''};
        \node[below=-0.05cm of m1-4-3] {\small ``Material''};
        \node[below=-0.05cm of m1-4-5] {\small Results};
        
        \node[below=0.5cm of m1] (bottom-container) {
            \begin{tikzpicture}
                \matrix (m2) [
                    matrix of nodes,
                    nodes={draw, minimum width=2cm, minimum height=1cm, inner sep=0pt, line width=1.5pt},
                    row sep=0.2cm,
                    column sep=0.3cm
                ] {
                    \includegraphics[width=2cm,height=2cm]{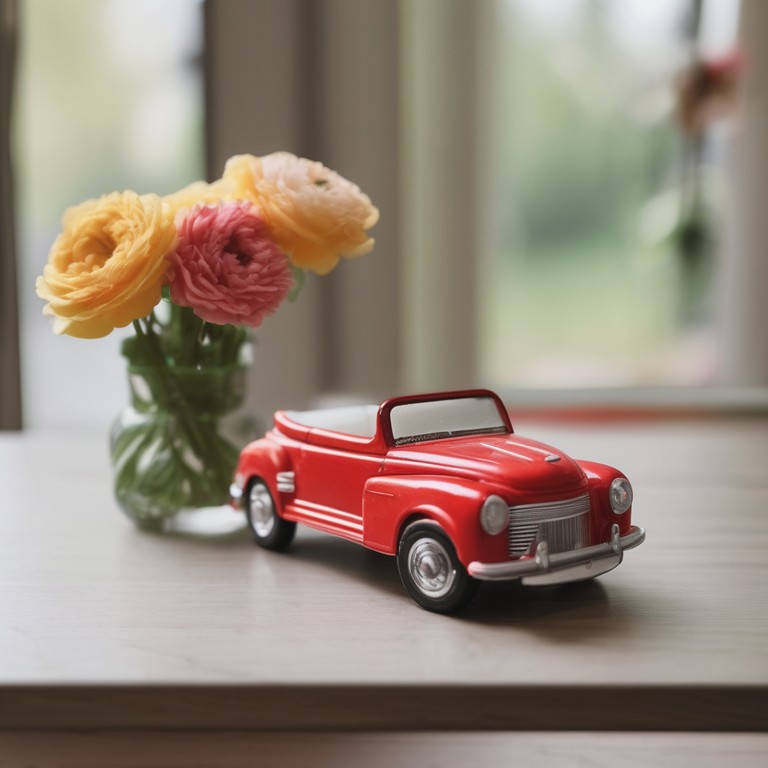} & \includegraphics[width=2cm,height=2cm]{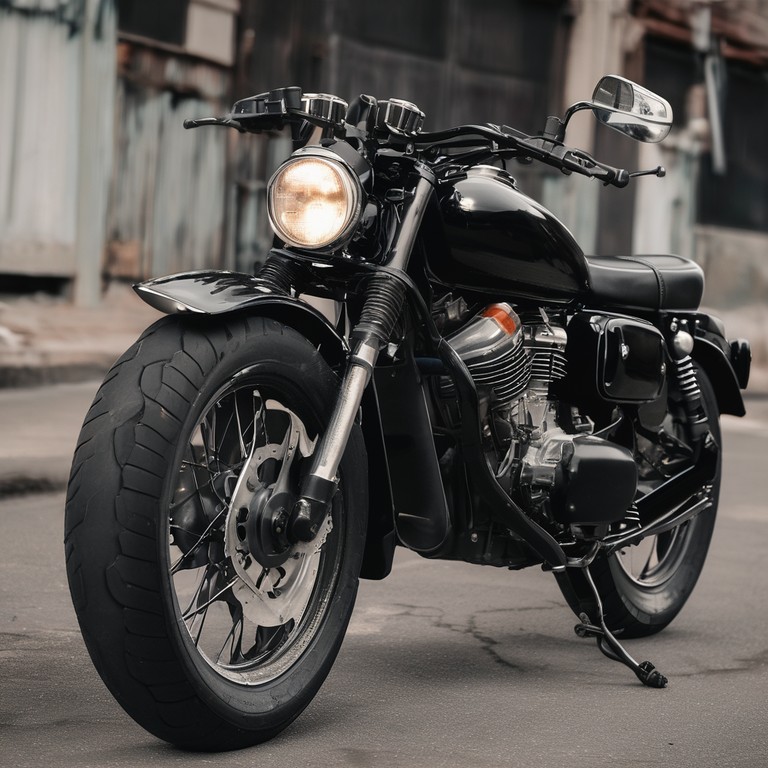} & \includegraphics[width=2cm,height=2cm]{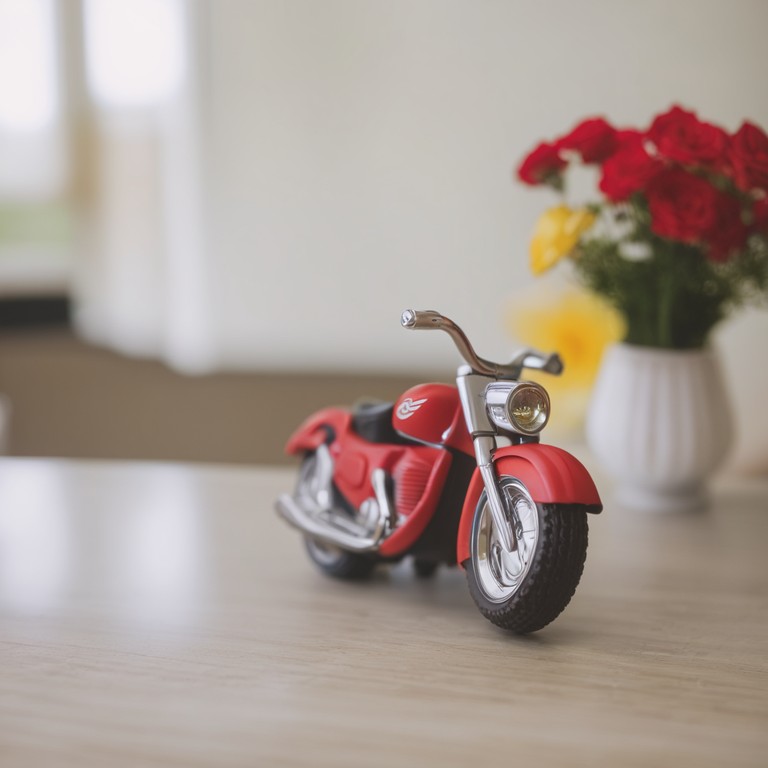} \\
                    \includegraphics[width=2cm,height=2cm]{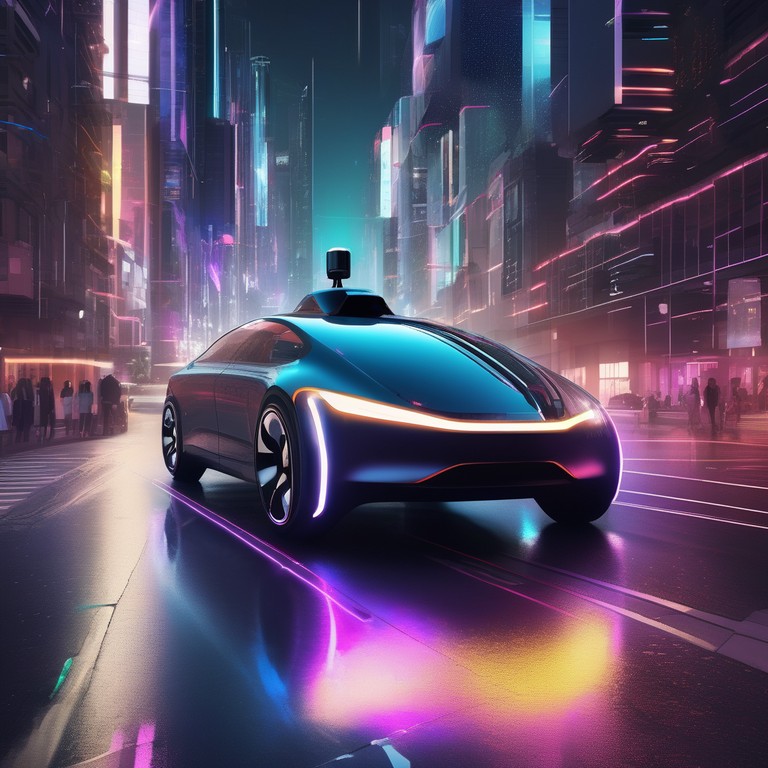} & \includegraphics[width=2cm,height=2cm]{images/additional_qualitative/2_way/vehicle/vehicle_1.jpg} & \includegraphics[width=2cm,height=2cm]{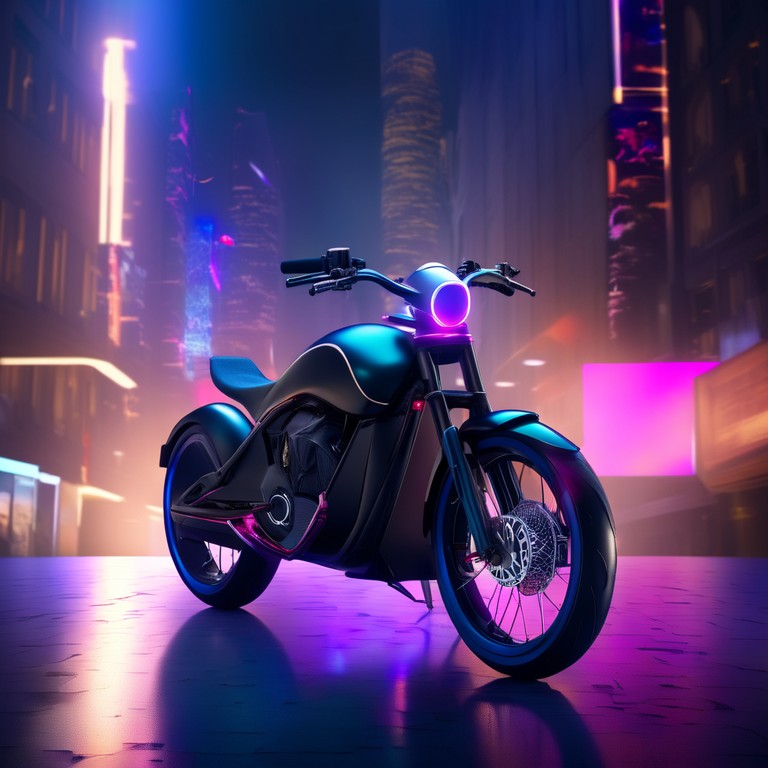} \\
                    \includegraphics[width=2cm,height=2cm]{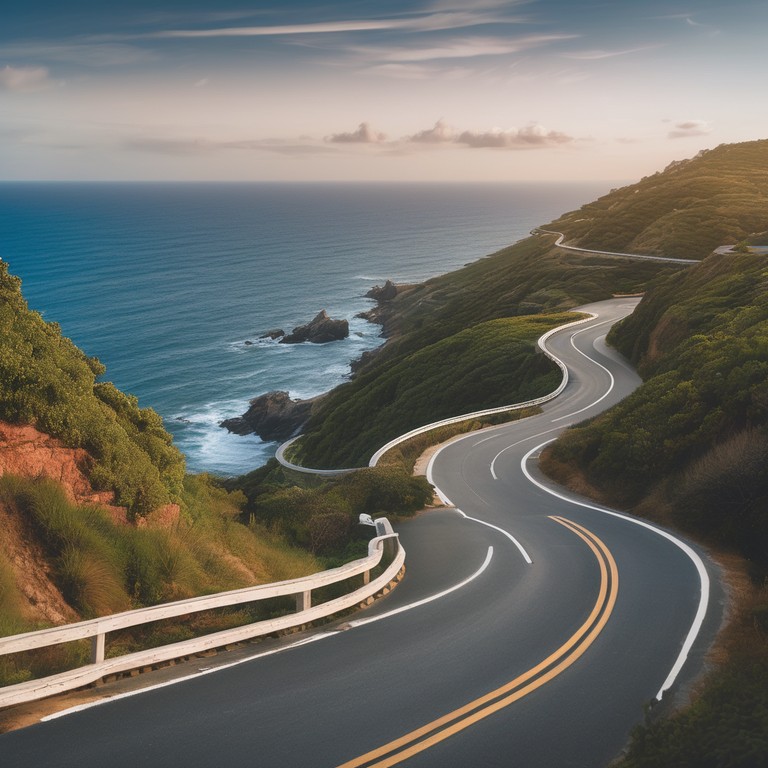} & \includegraphics[width=2cm,height=2cm]{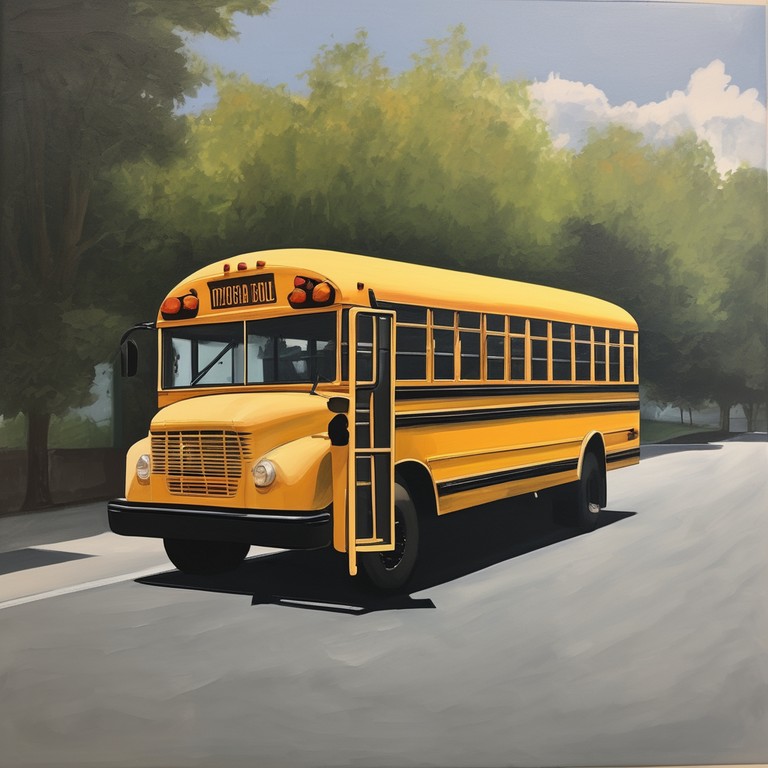} & \includegraphics[width=2cm,height=2cm]{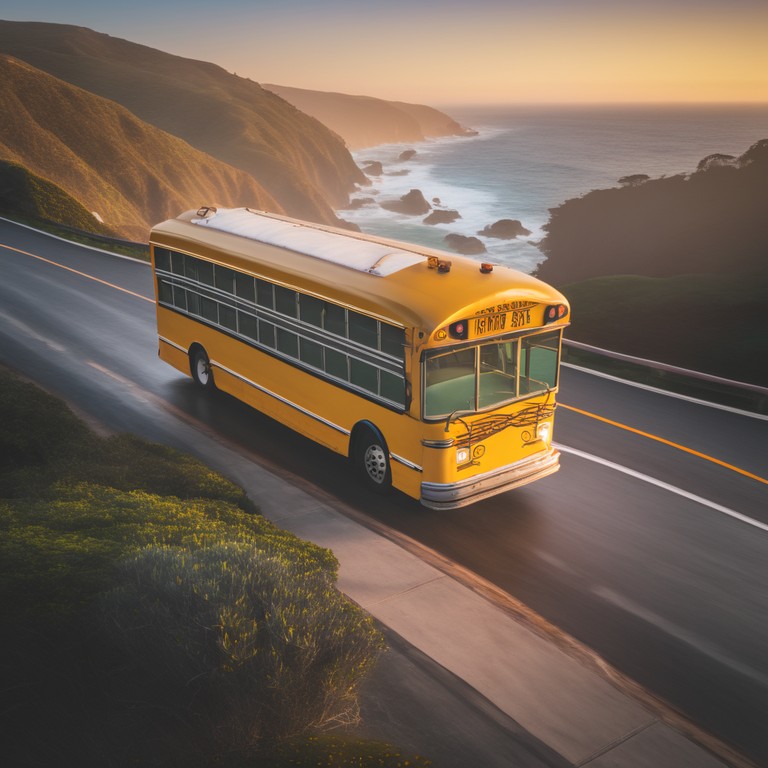} \\
                    \includegraphics[width=2cm,height=2cm]{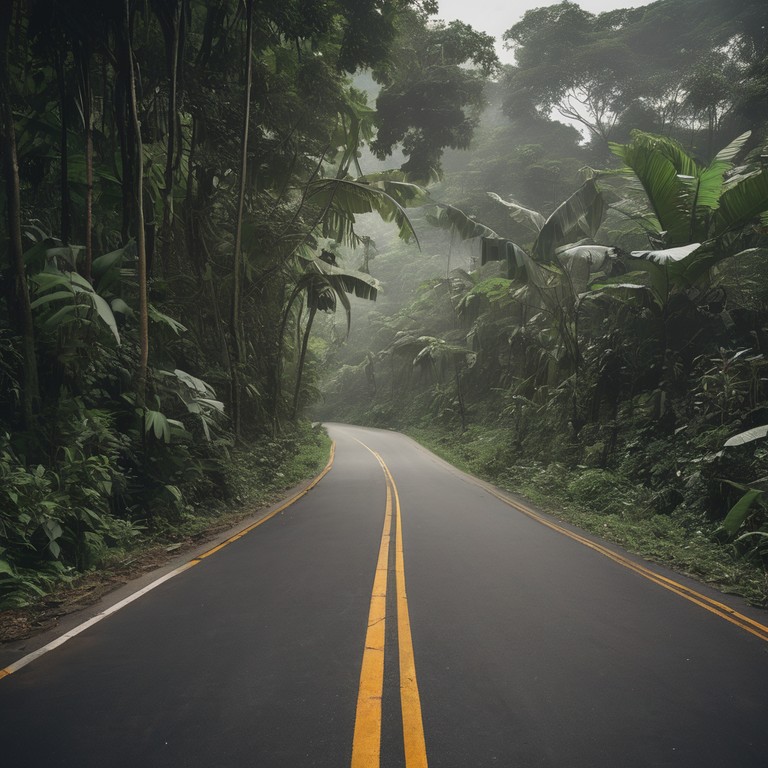} & \includegraphics[width=2cm,height=2cm]{images/additional_qualitative/2_way/vehicle/vehicle_2.jpg} & \includegraphics[width=2cm,height=2cm]{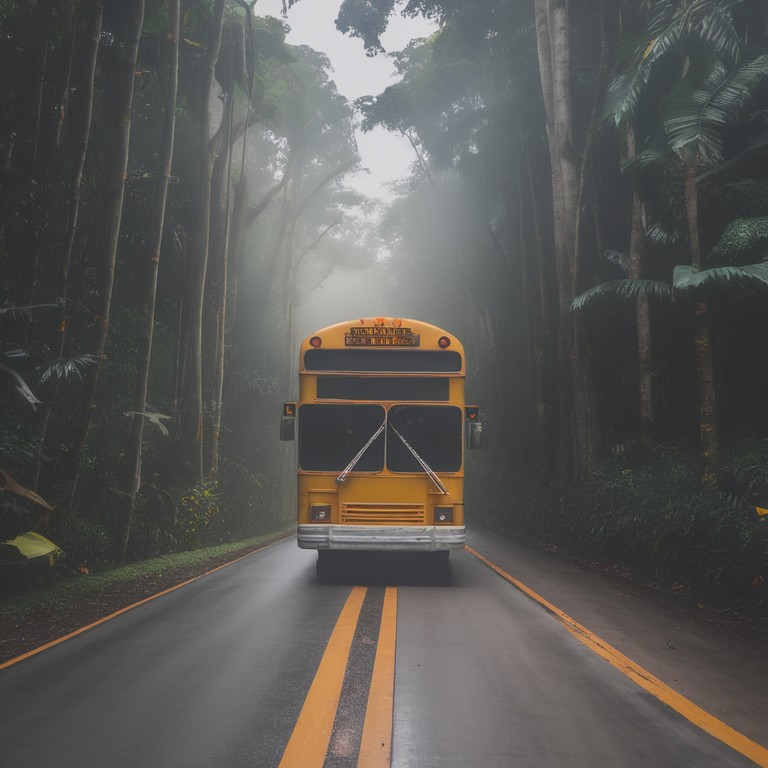} \\
                };

                \node[below=-0.05cm of m2-4-1] {\small ``Scene"};
                \node[below=-0.05cm of m2-4-2] {\small ``Vehicle"};
                \node[below=-0.05cm of m2-4-3] {\small Result};

                \path (m2-1-1) -- (m2-1-2) node[midway, yshift=5.0pt] {\Large $+$};
                \path (m2-1-2) -- (m2-1-3) node[midway, yshift=5.0pt] {\Large $=$};
                \path (m2-2-1) -- (m2-2-2) node[midway, yshift=5.0pt] {\Large $+$};
                \path (m2-2-2) -- (m2-2-3) node[midway, yshift=5.0pt] {\Large $=$};

                \path (m2-3-1) -- (m2-3-2) node[midway, yshift=5.0pt] {\Large $+$};
                \path (m2-3-2) -- (m2-3-3) node[midway, yshift=5.0pt] {\Large $=$};
                \path (m2-4-1) -- (m2-4-2) node[midway, yshift=5.0pt] {\Large $+$};
                \path (m2-4-2) -- (m2-4-3) node[midway, yshift=5.0pt] {\Large $=$};
                
                \matrix (m3) [
                    matrix of nodes,
                    nodes={draw, minimum width=2cm, minimum height=1cm, inner sep=0pt, line width=1.5pt},
                    row sep=0.2cm,
                    column sep=0.3cm,
                    right=0.8cm of m2
                ] {
                    \includegraphics[width=2cm,height=2cm]{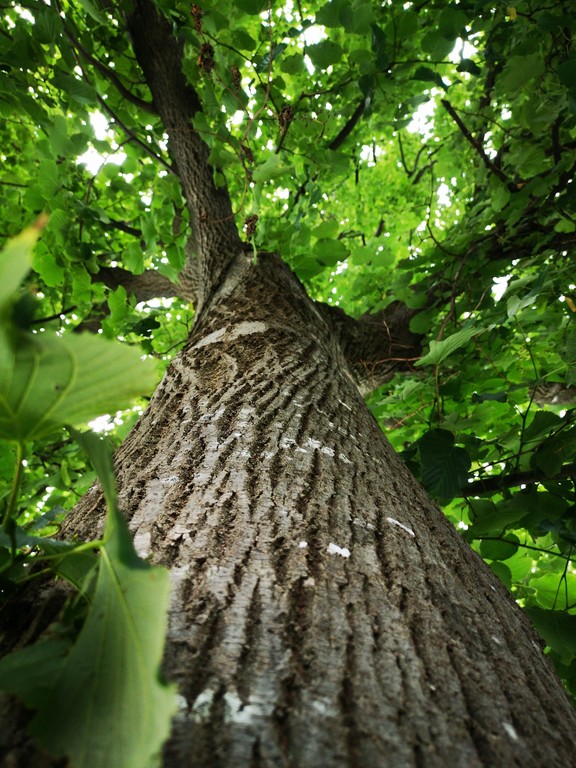} & \includegraphics[width=2cm,height=2cm]{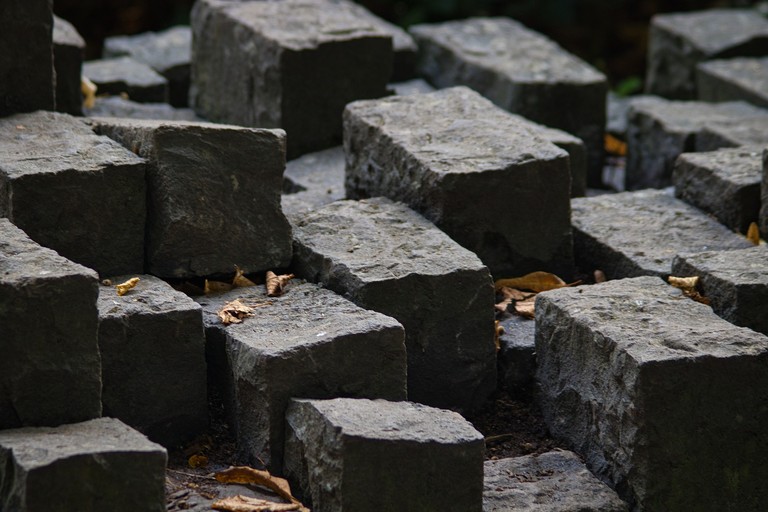} & \includegraphics[width=2cm,height=2cm]{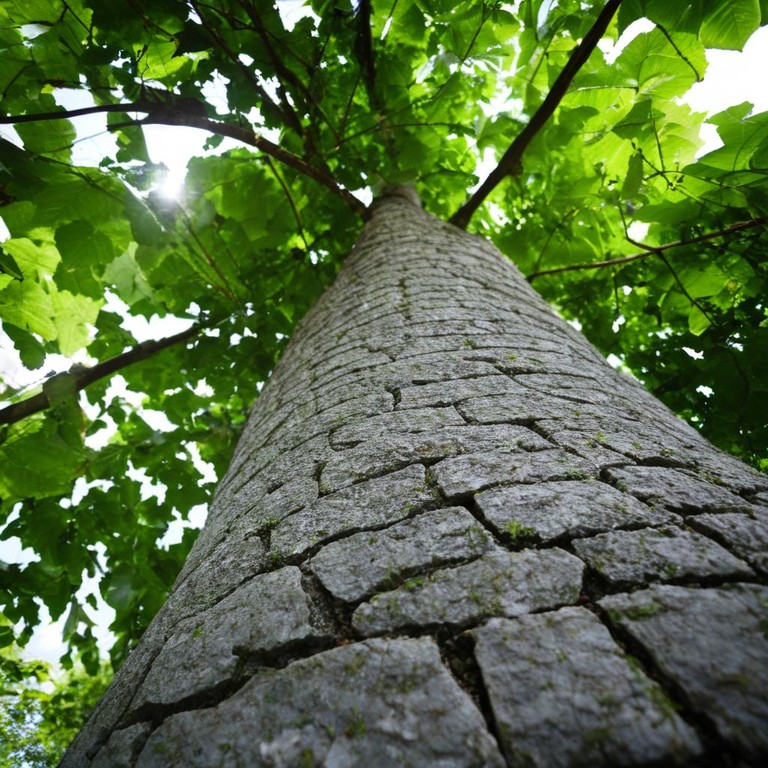} \\
                    \includegraphics[width=2cm,height=2cm]{images/additional_qualitative/2_way/material/input_1.jpg} & \includegraphics[width=2cm,height=2cm]{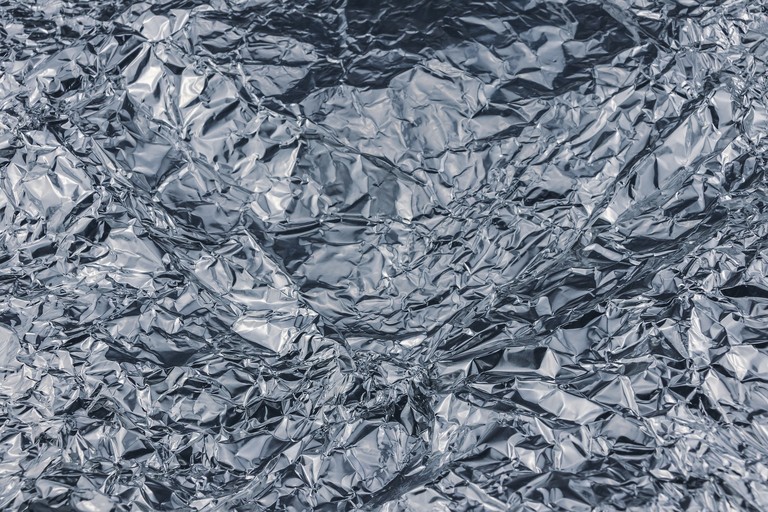} & \includegraphics[width=2cm,height=2cm]{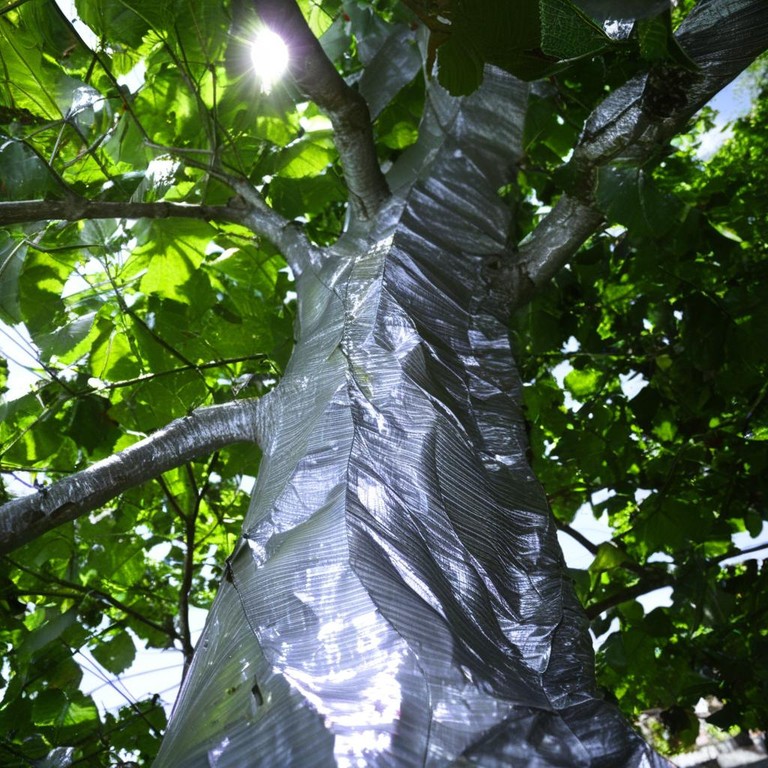} \\
                    \includegraphics[width=2cm,height=2cm]{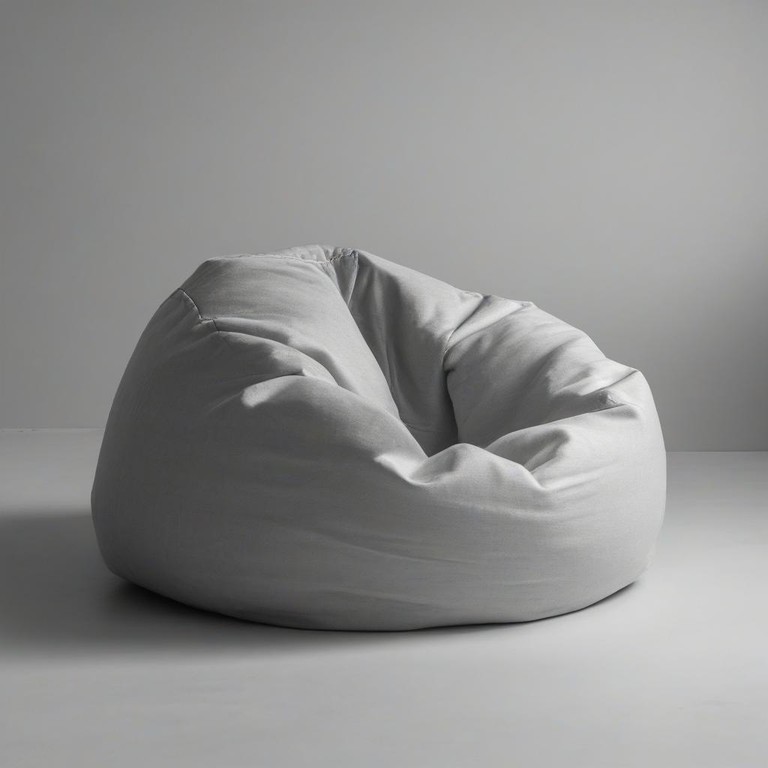} & \includegraphics[width=2cm,height=2cm]{images/additional_qualitative/2_way/material/material_1.jpg} & \includegraphics[width=2cm,height=2cm]{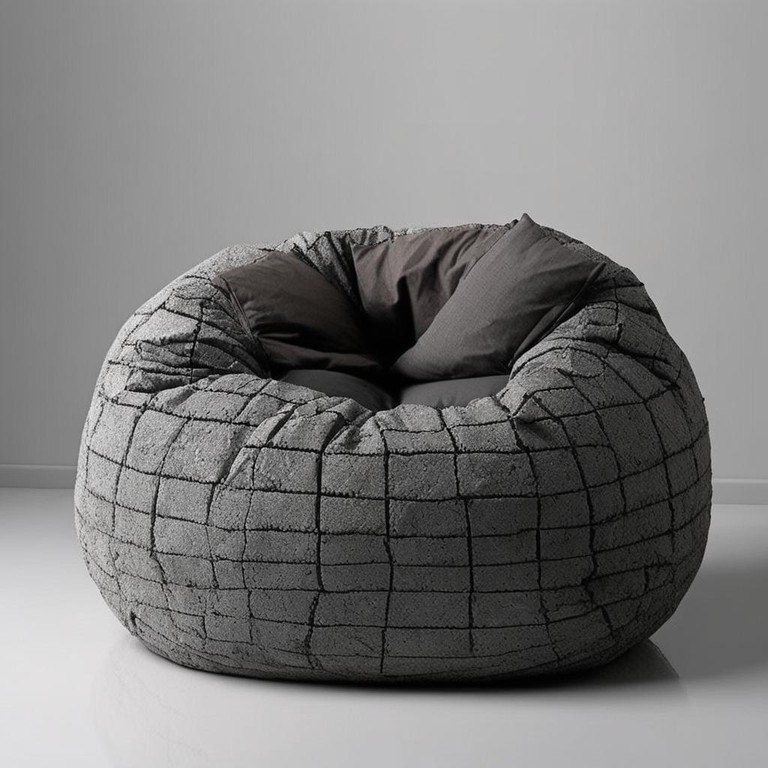} \\
                    \includegraphics[width=2cm,height=2cm]{images/additional_qualitative/2_way/material/input_2.jpg} & \includegraphics[width=2cm,height=2cm]{images/additional_qualitative/2_way/material/material_2.jpg} & \includegraphics[width=2cm,height=2cm]{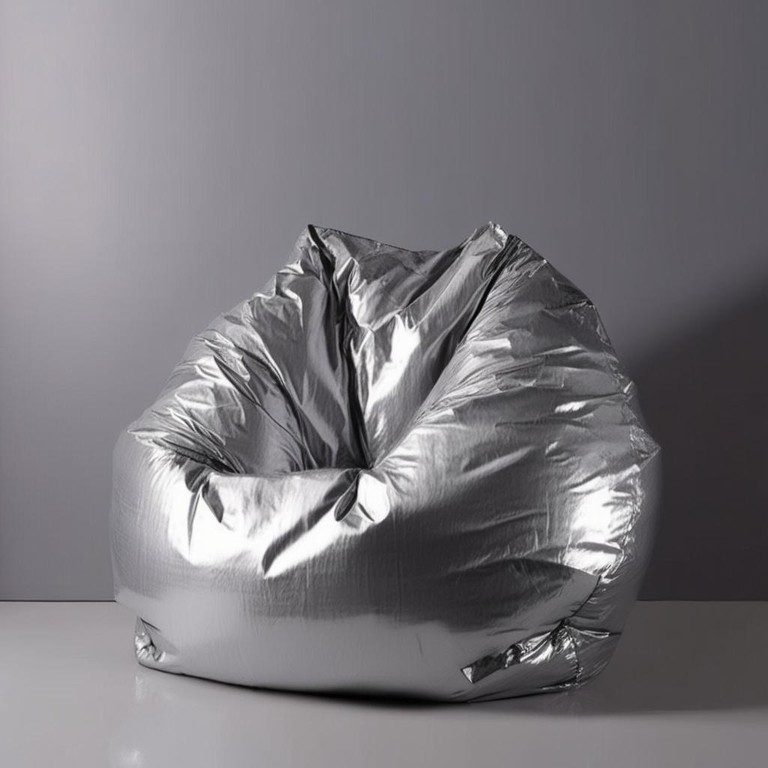} \\
                };

                \node[below=-0.05cm of m3-4-1] {\small ``Object"};
                \node[below=-0.05cm of m3-4-2] {\small ``Material"};
                \node[below=-0.05cm of m3-4-3] {\small Result};

                \path (m3-1-1) -- (m3-1-2) node[midway, yshift=5.0pt] {\Large $+$};
                \path (m3-1-2) -- (m3-1-3) node[midway, yshift=5.0pt] {\Large $=$};
                \path (m3-2-1) -- (m3-2-2) node[midway, yshift=5.0pt] {\Large $+$};
                \path (m3-2-2) -- (m3-2-3) node[midway, yshift=5.0pt] {\Large $=$};
                \path (m3-3-1) -- (m3-3-2) node[midway, yshift=5.0pt] {\Large $+$};
                \path (m3-3-2) -- (m3-3-3) node[midway, yshift=5.0pt] {\Large $=$};
                \path (m3-4-1) -- (m3-4-2) node[midway, yshift=5.0pt] {\Large $+$};
                \path (m3-4-2) -- (m3-4-3) node[midway, yshift=5.0pt] {\Large $=$};
                
                \begin{pgfonlayer}{background}
                    \node[draw, rounded corners, inner sep=0.3cm, fit=(m2), line width=1pt] {};
                    \node[draw, rounded corners, inner sep=0.3cm, fit=(m3), line width=1pt] {};
                \end{pgfonlayer}
            \end{tikzpicture}
        };
        
        \begin{pgfonlayer}{background}
            \node[draw, rounded corners, inner sep=0.3cm, fit=(m1), line width=1pt] {};
        \end{pgfonlayer}
    \end{tikzpicture}
    \caption{Additional qualitative results generated using IP-Composer.}
    \label{fig:additional_1}
\end{figure*}

%% file: figures/additional_qualitative_results_2.tex
\begin{figure*}[t]
    \centering
    \setlength{\belowcaptionskip}{-5pt}
    \setlength{\abovecaptionskip}{4pt}
    
    \begin{tikzpicture}
        \matrix (m1) [matrix of nodes,
            nodes={draw, minimum width=2cm, minimum height=1cm, inner sep=0pt, line width=1.5pt},
            row sep=0.4cm,
            column sep=0.443cm
        ] {
            \includegraphics[width=2cm,height=2cm]{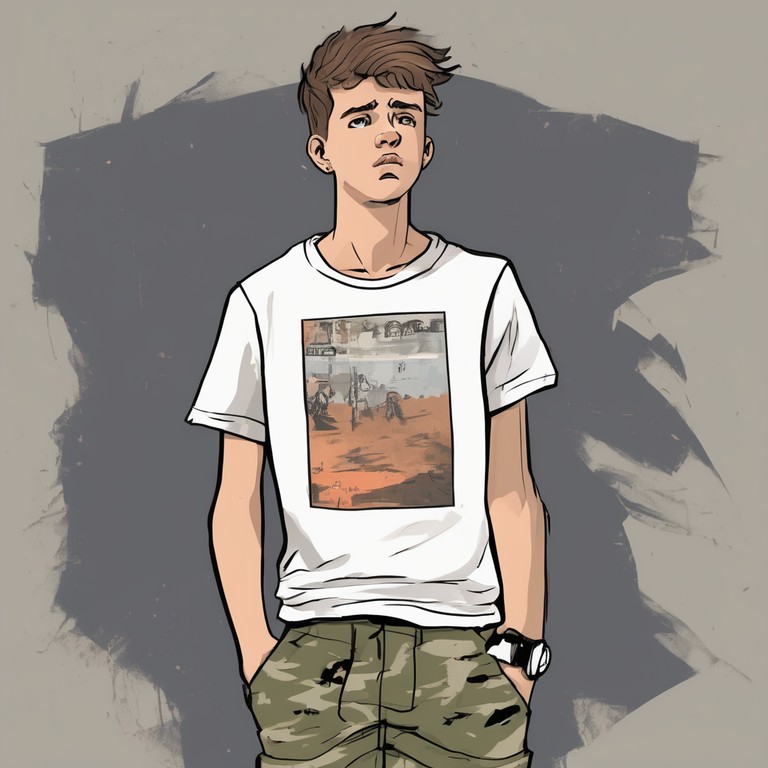} & \includegraphics[width=2cm,height=2cm]{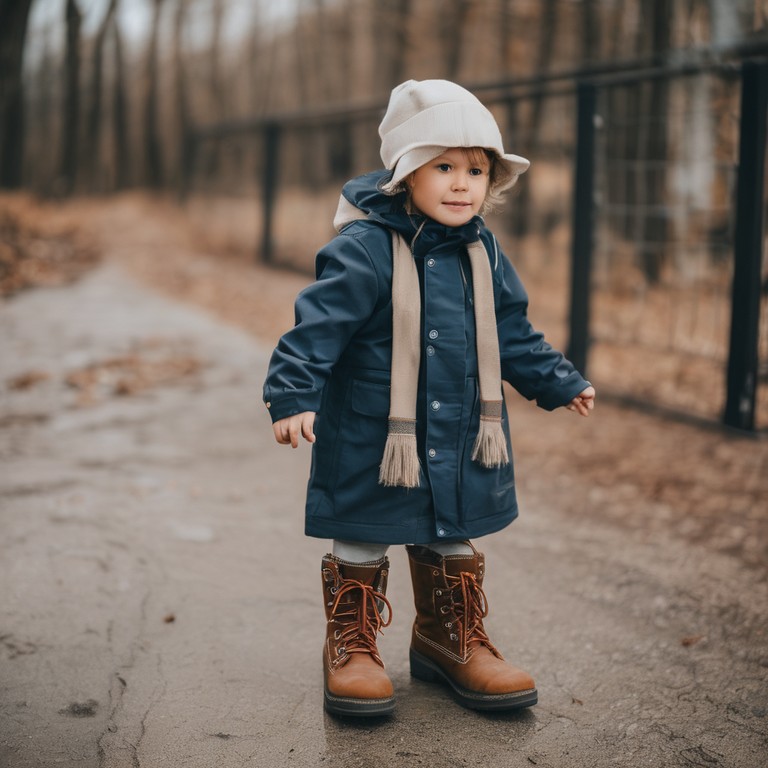} & \includegraphics[width=2cm,height=2cm]{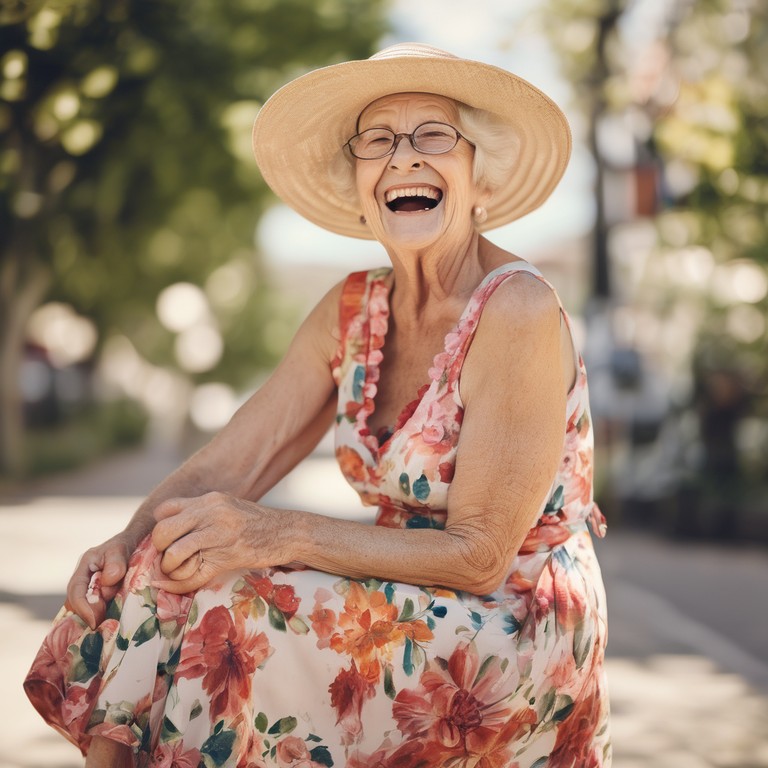} & \includegraphics[width=2cm,height=2cm]{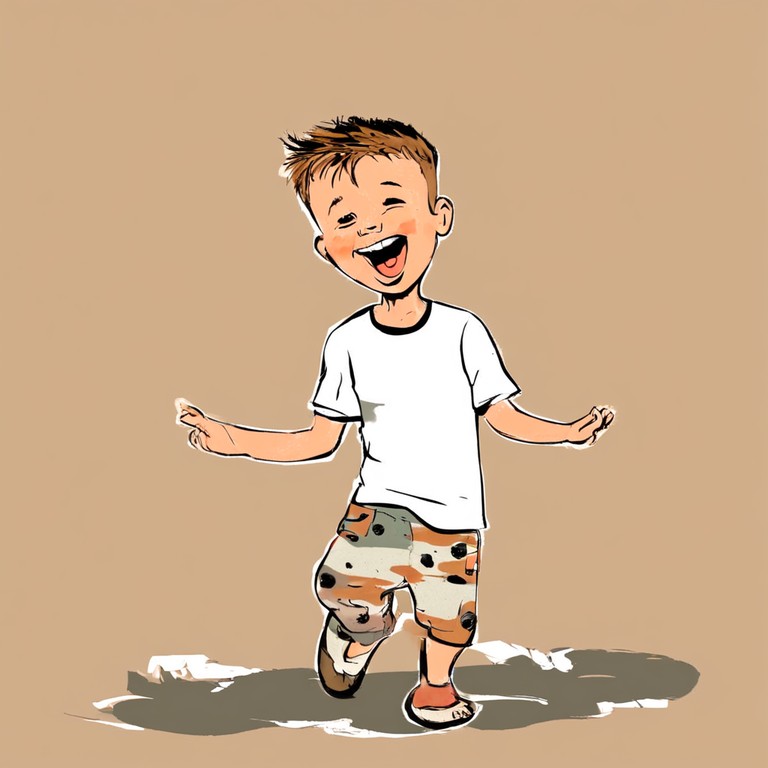} & \includegraphics[width=2cm,height=2cm]{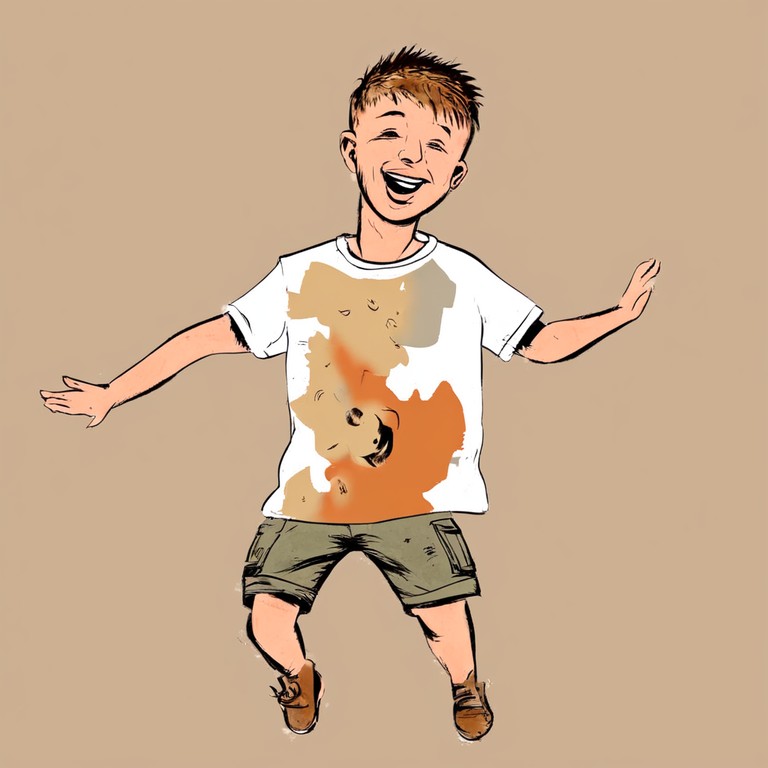} & \includegraphics[width=2cm,height=2cm]{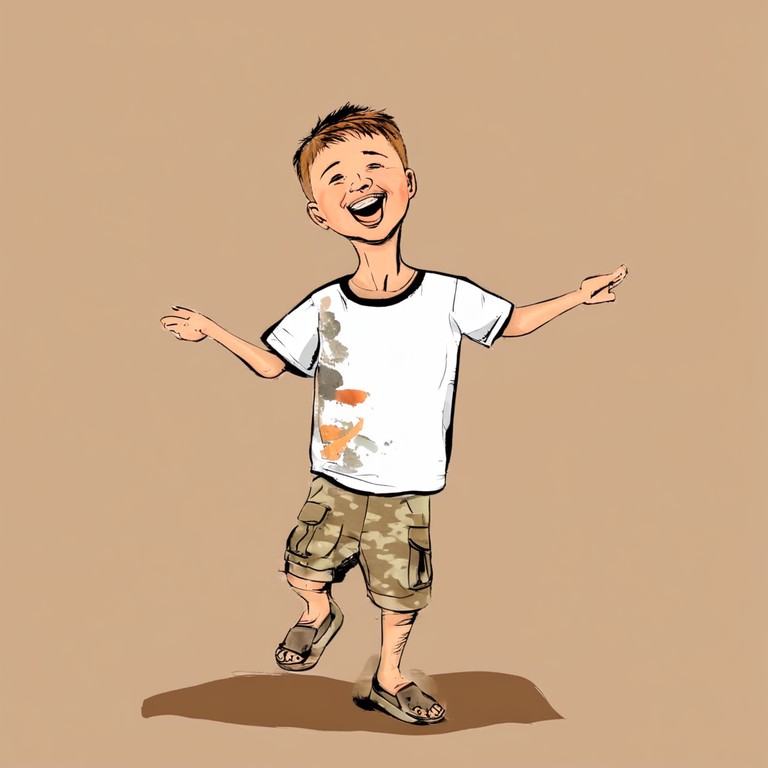} \\
            \includegraphics[width=2cm,height=2cm]{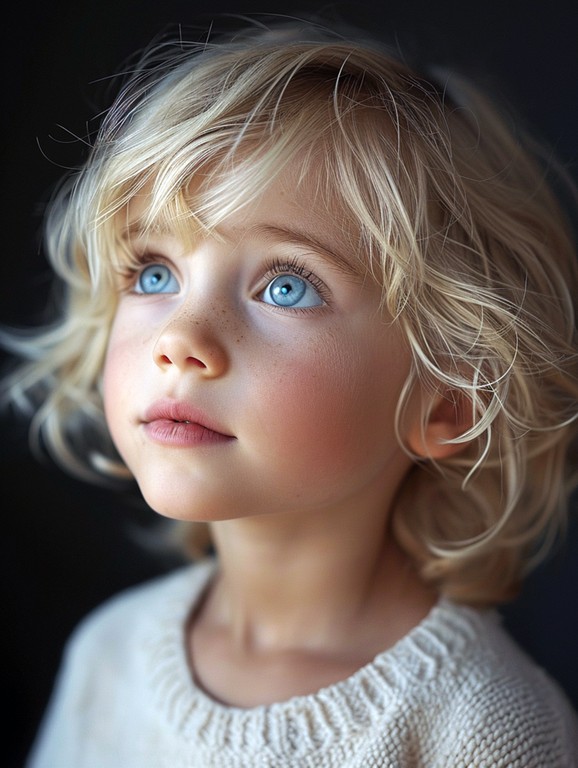} & \includegraphics[width=2cm,height=2cm]{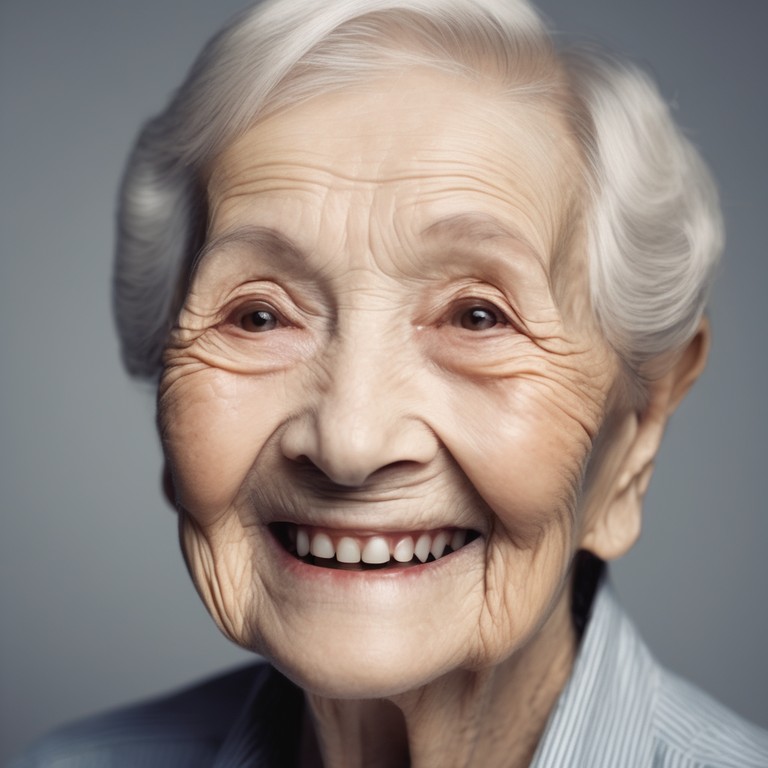} & \includegraphics[width=2cm, height=2cm, trim=250 250 250 250, clip]{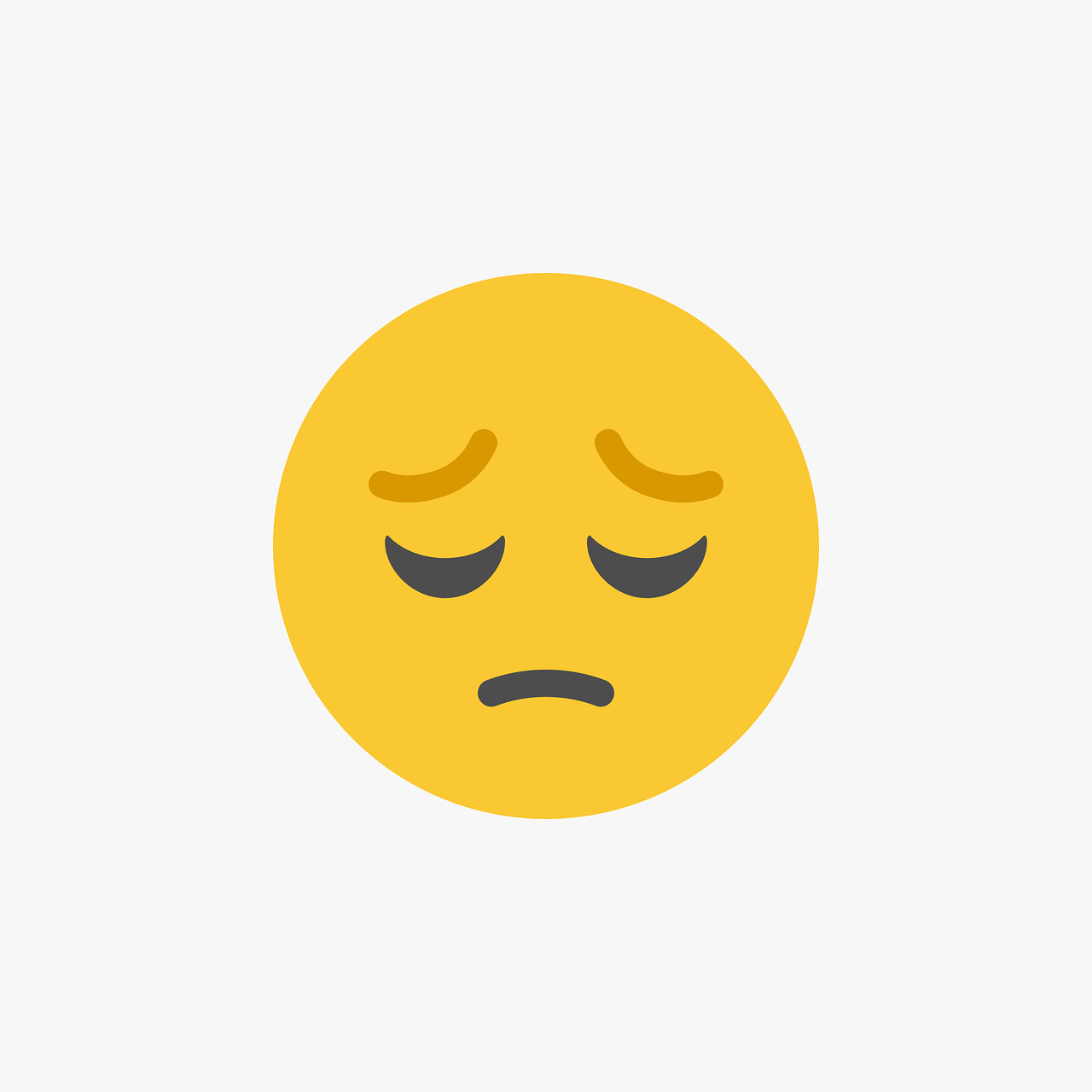} & \includegraphics[width=2cm,height=2cm]{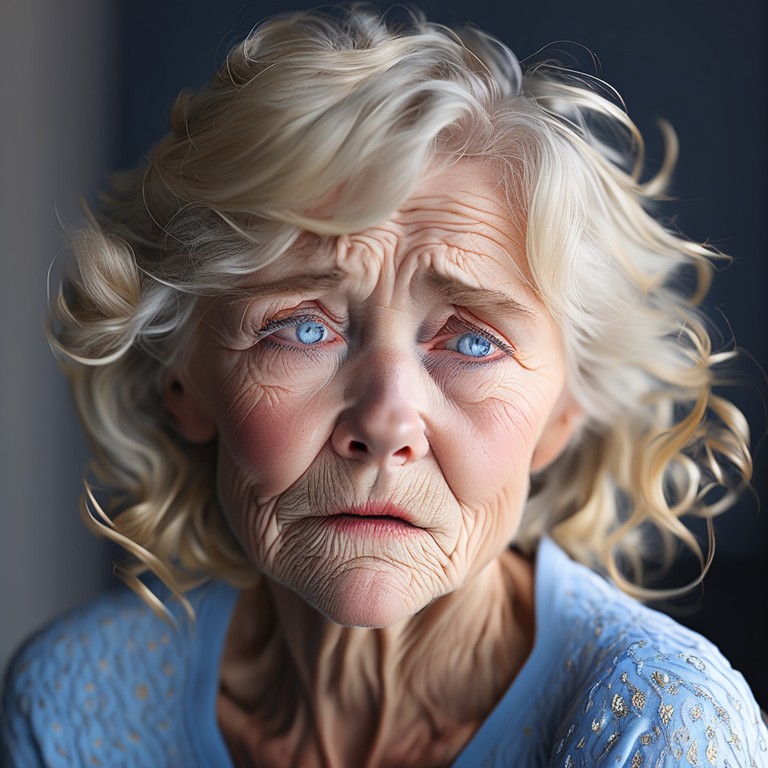} & \includegraphics[width=2cm,height=2cm]{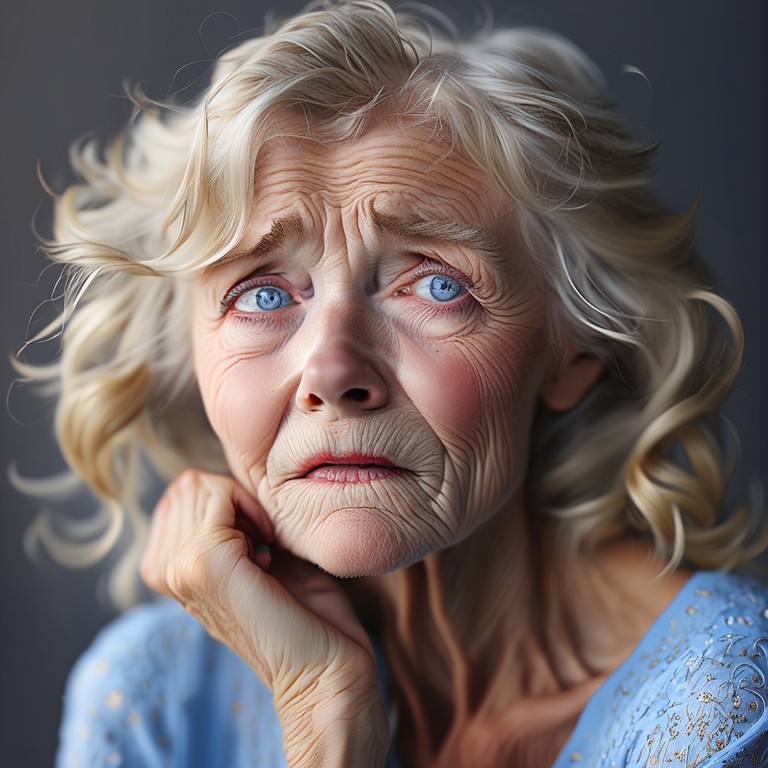} & \includegraphics[width=2cm,height=2cm]{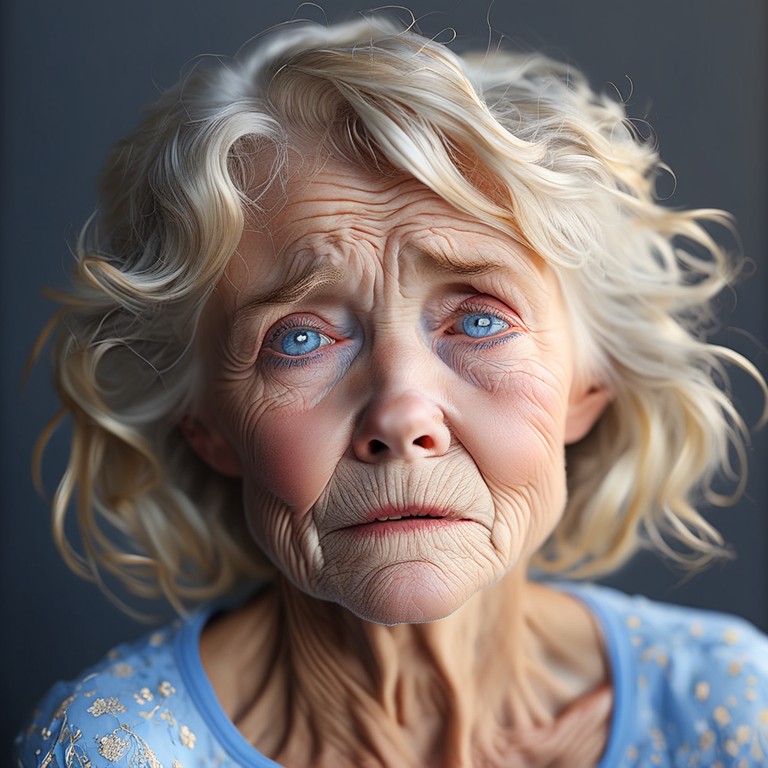} \\
            \includegraphics[width=2cm,height=2cm]{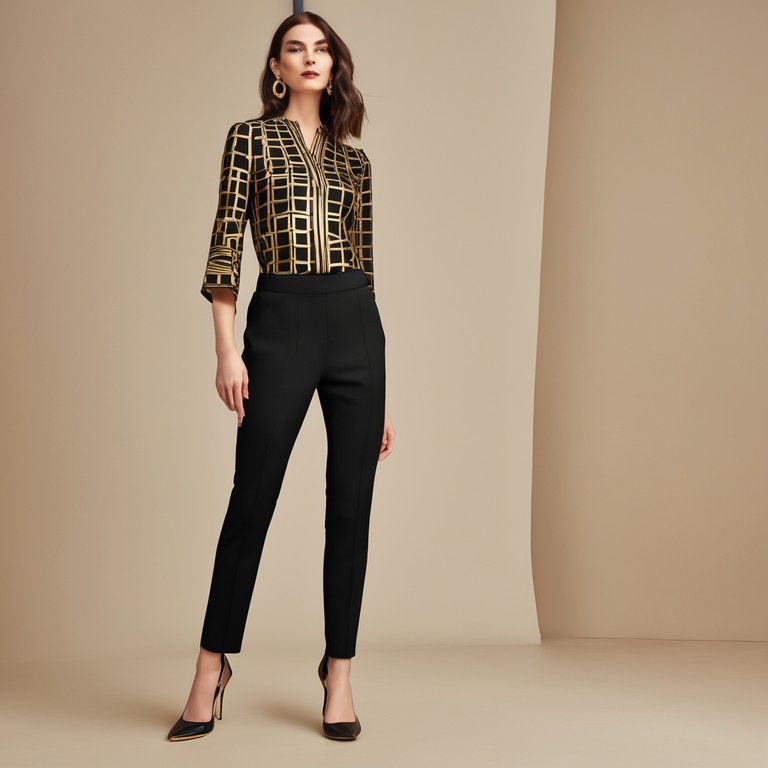} & \includegraphics[width=2cm,height=2cm]{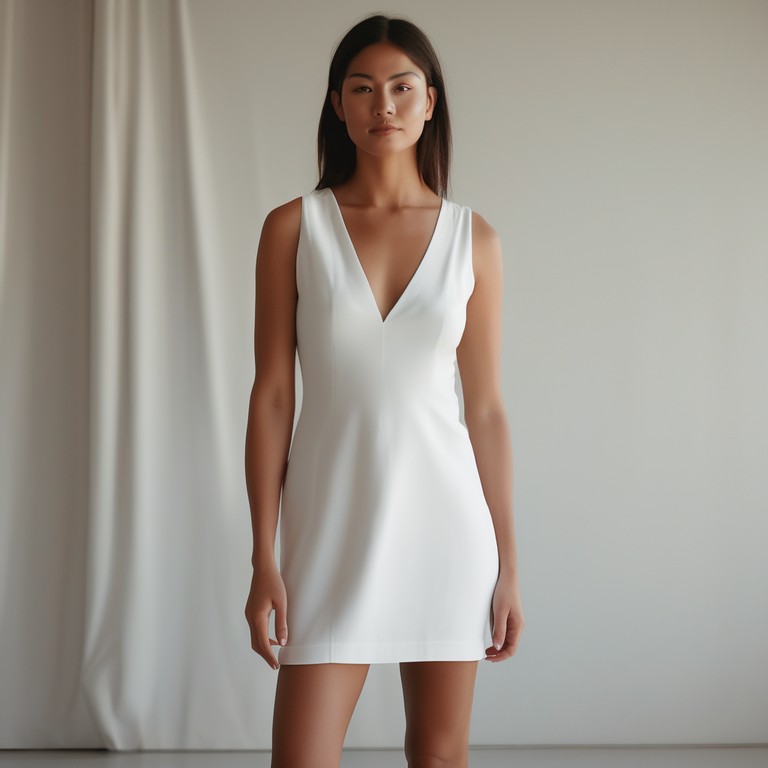} & \includegraphics[width=2cm,height=2cm]{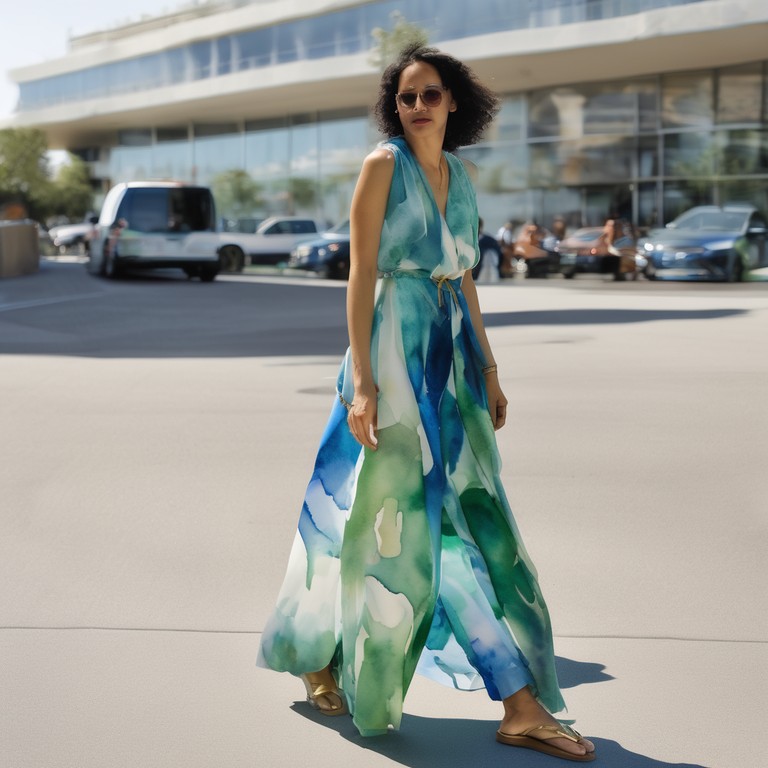} & \includegraphics[width=2cm,height=2cm]{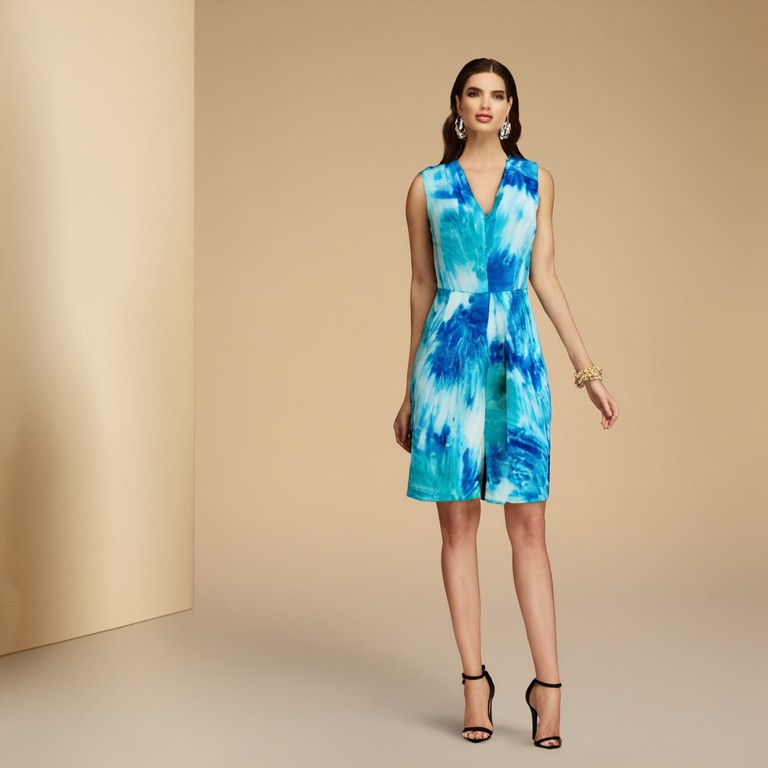} & \includegraphics[width=2cm,height=2cm]{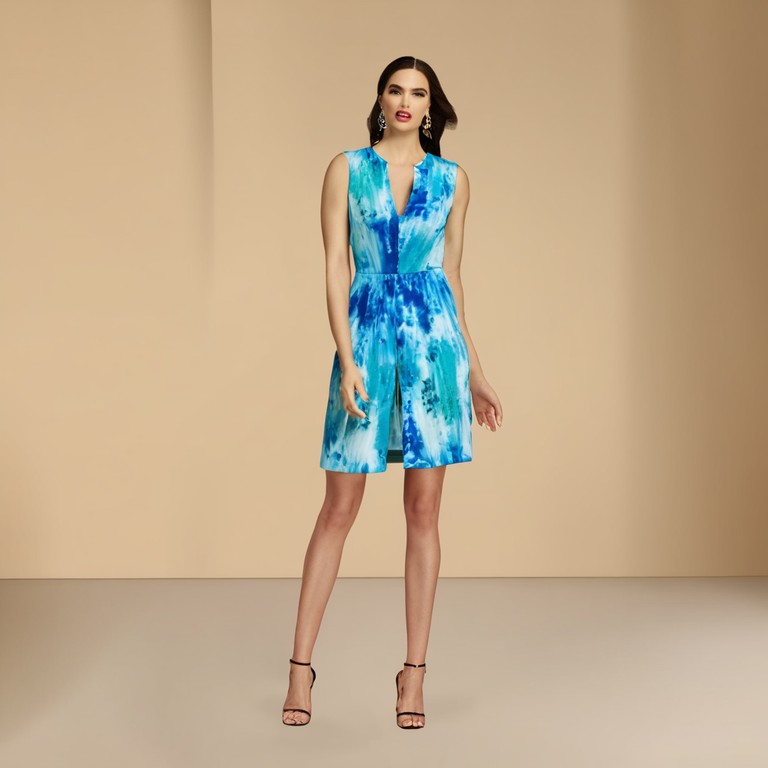} & \includegraphics[width=2cm,height=2cm]{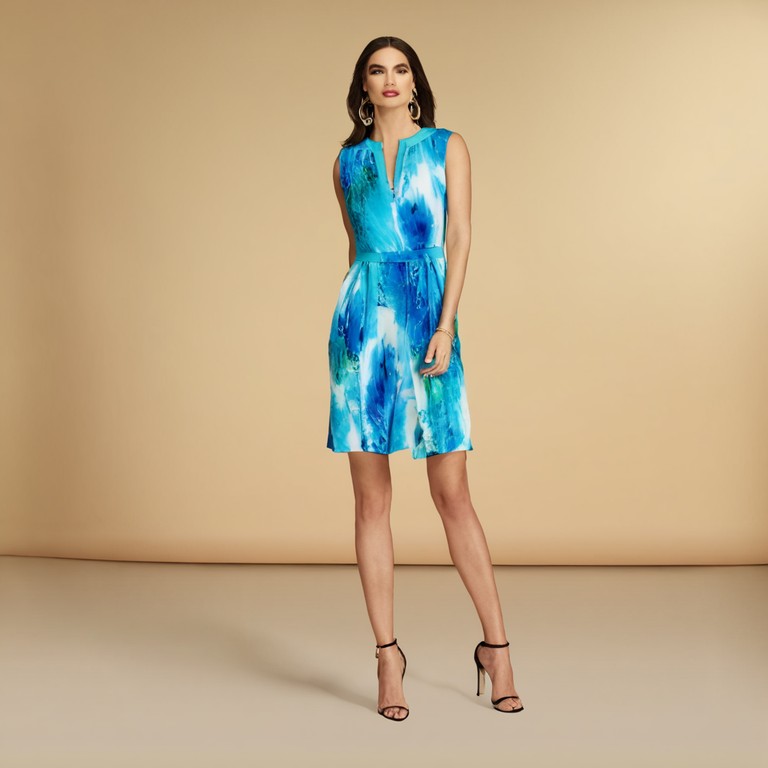} \\
            \includegraphics[width=2cm,height=2cm]{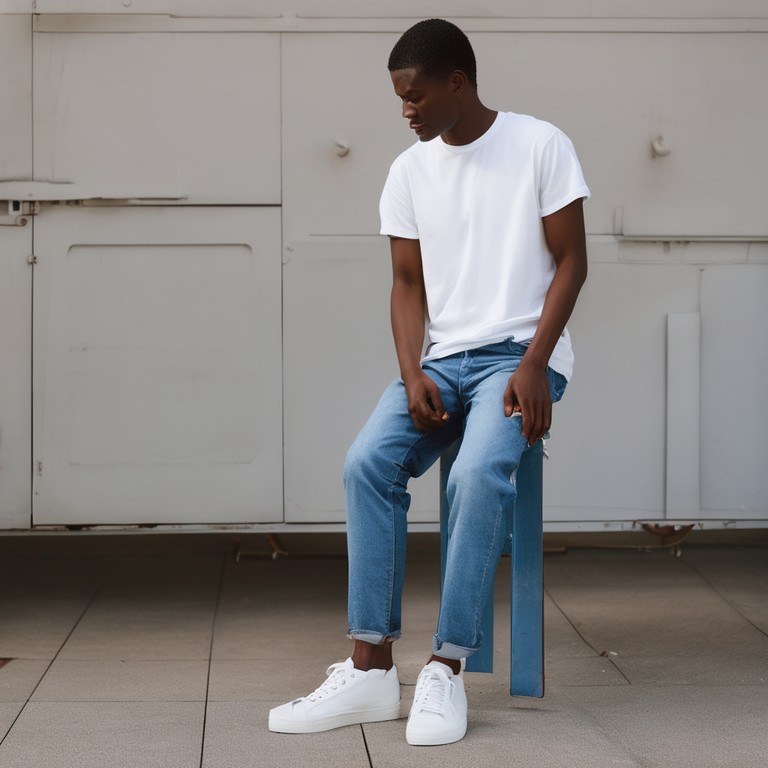} & \includegraphics[width=2cm,height=2cm]{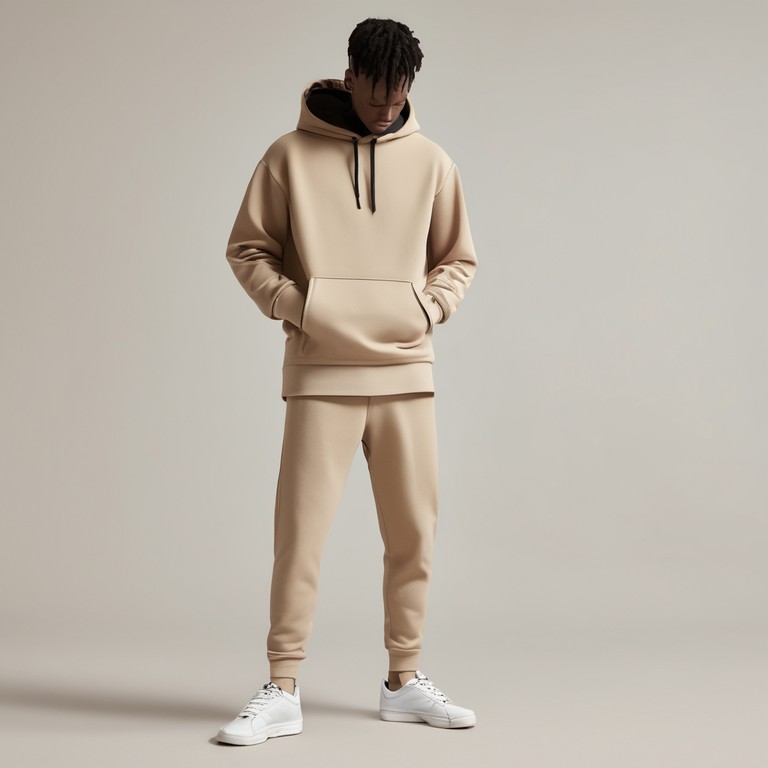} & \includegraphics[width=2cm,height=2cm]{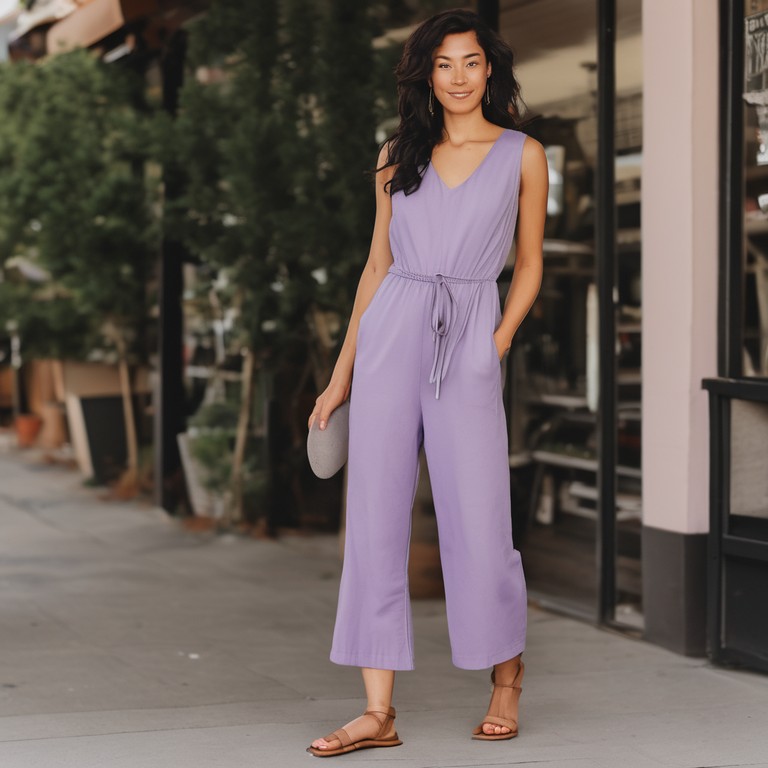} & \includegraphics[width=2cm,height=2cm]{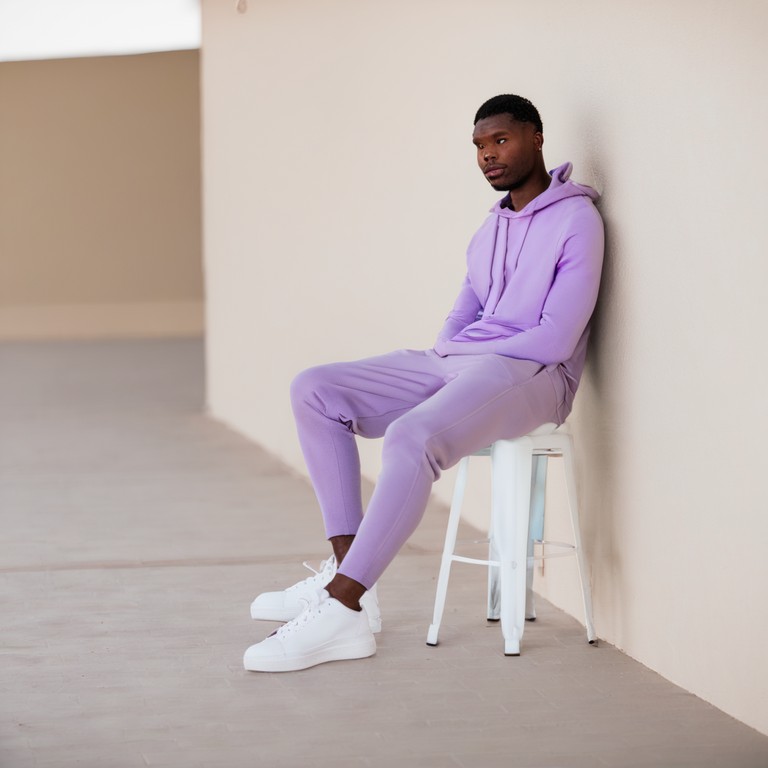} & \includegraphics[width=2cm,height=2cm]{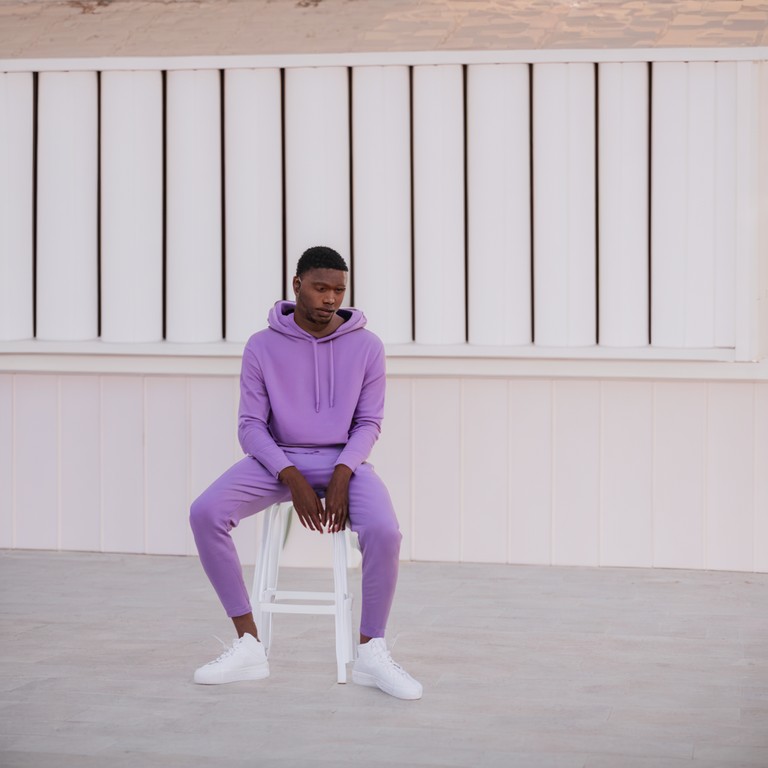} & \includegraphics[width=2cm,height=2cm]{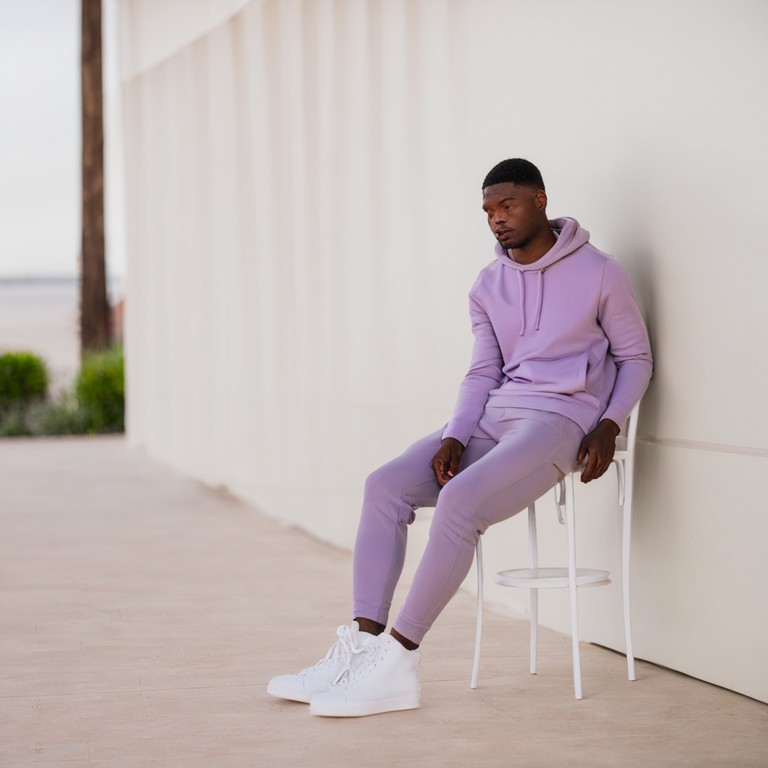} \\
        };

        \path (m1-1-1) -- (m1-1-2) node[midway] {\Large $+$};
        \path (m1-1-2) -- (m1-1-3) node[midway] {\Large $+$};
        \path (m1-1-3) -- (m1-1-4) node[midway] {\Large $=$};

        \path (m1-2-1) -- (m1-2-2) node[midway] {\Large $+$};
        \path (m1-2-2) -- (m1-2-3) node[midway] {\Large $+$};
        \path (m1-2-3) -- (m1-2-4) node[midway] {\Large $=$};

        \path (m1-3-1) -- (m1-3-2) node[midway] {\Large $+$};
        \path (m1-3-2) -- (m1-3-3) node[midway] {\Large $+$};
        \path (m1-3-3) -- (m1-3-4) node[midway] {\Large $=$};

        \path (m1-4-1) -- (m1-4-2) node[midway] {\Large $+$};
        \path (m1-4-2) -- (m1-4-3) node[midway] {\Large $+$};
        \path (m1-4-3) -- (m1-4-4) node[midway] {\Large $=$};

        \node[below=-0.05cm of m1-2-1] {\small ``Person''};
        \node[below=-0.05cm of m1-2-2] {\small ``Age''};
        \node[below=-0.05cm of m1-2-3] {\small ``Expression''};
        \node[below=-0.05cm of m1-2-5] {\small Results};

        \node[below=-0.05cm of m1-4-1] {\small ``Person''};
        \node[below=-0.05cm of m1-4-2] {\small ``Outfit Shape''};
        \node[below=-0.05cm of m1-4-3] {\small ``Outfit Appearance''};
        \node[below=-0.05cm of m1-4-5] {\small Results};
        
        \node[below=0.5cm of m1] (bottom-container) {
            \begin{tikzpicture}
                \matrix (m2) [
                    matrix of nodes,
                    nodes={draw, minimum width=2cm, minimum height=1cm, inner sep=0pt, line width=1.5pt},
                    row sep=0.35cm,
                    column sep=0.3cm
                ] {
                    \includegraphics[width=2cm,height=2cm]{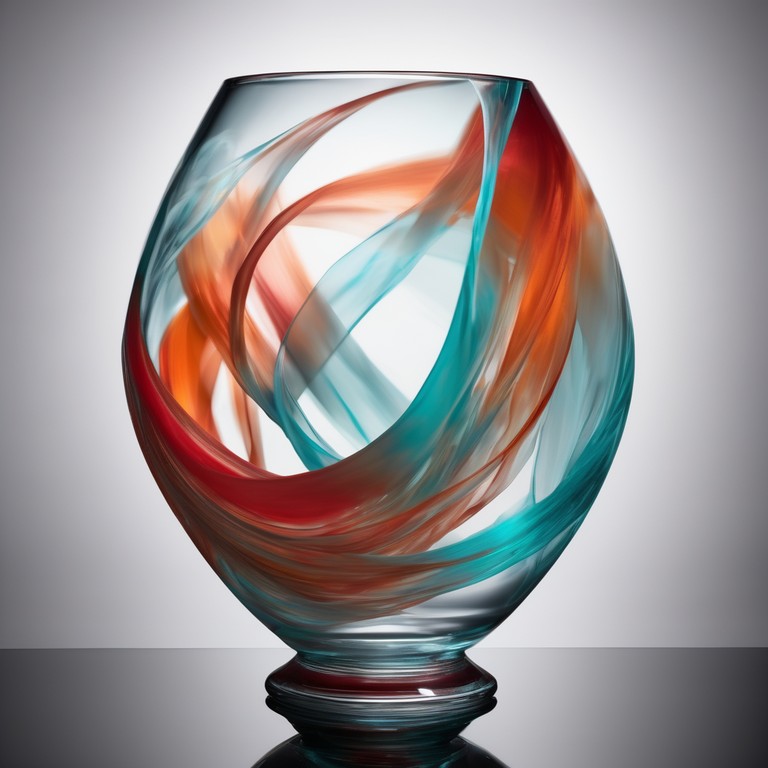} & \includegraphics[width=2cm,height=2cm]{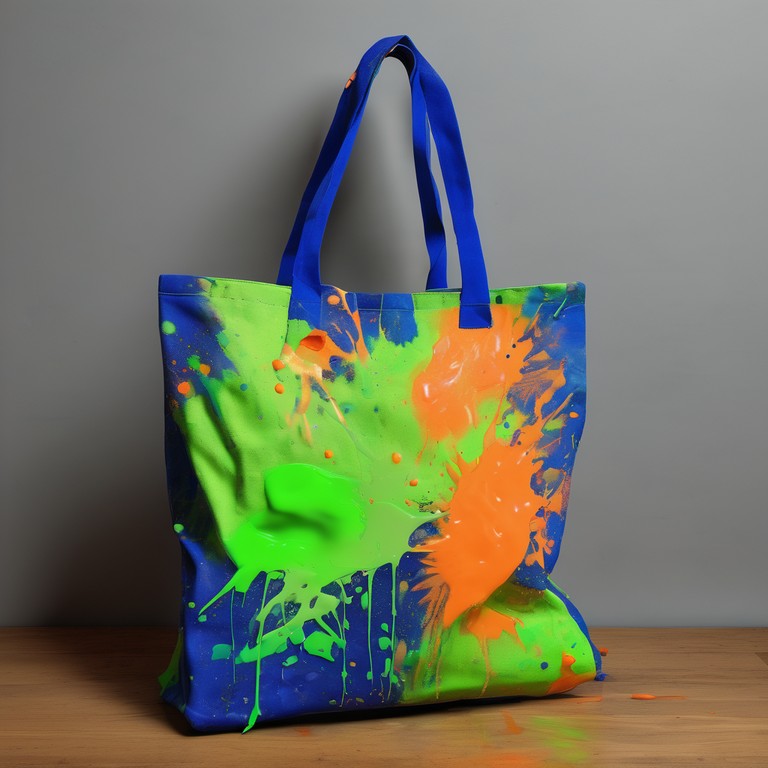} & \includegraphics[width=2cm,height=2cm]{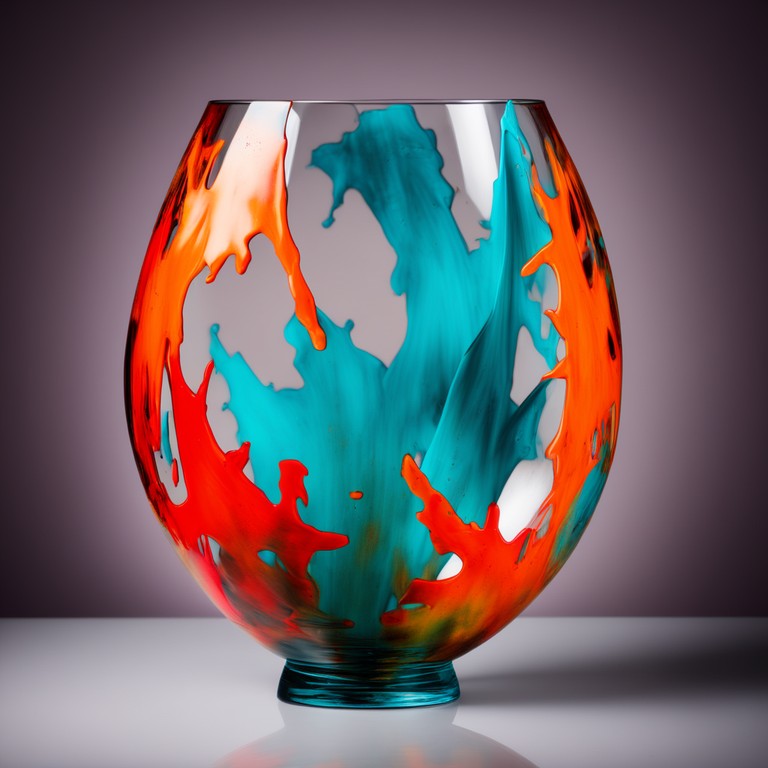} \\
                    \includegraphics[width=2cm,height=2cm]{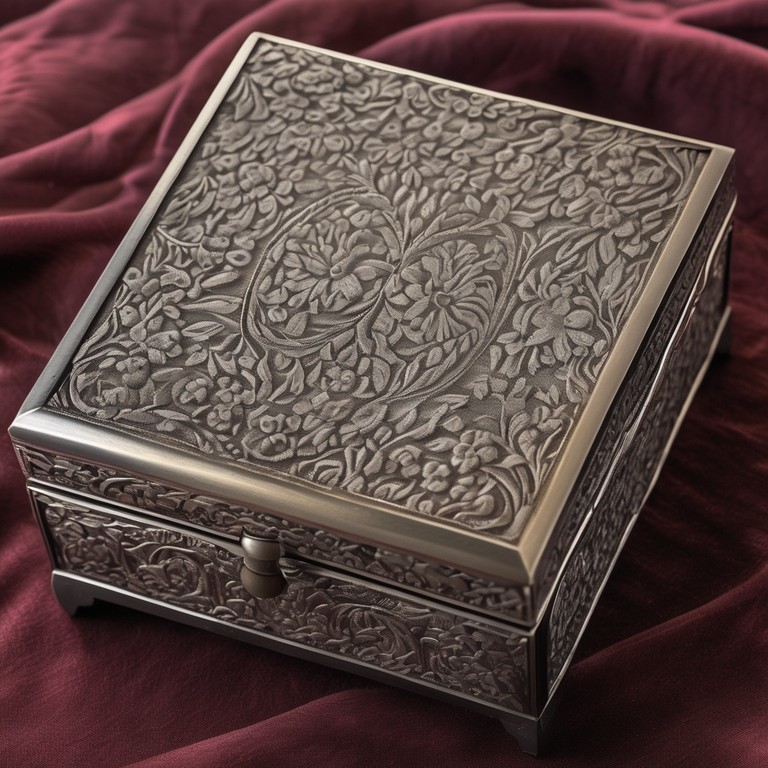} & \includegraphics[width=2cm,height=2cm]{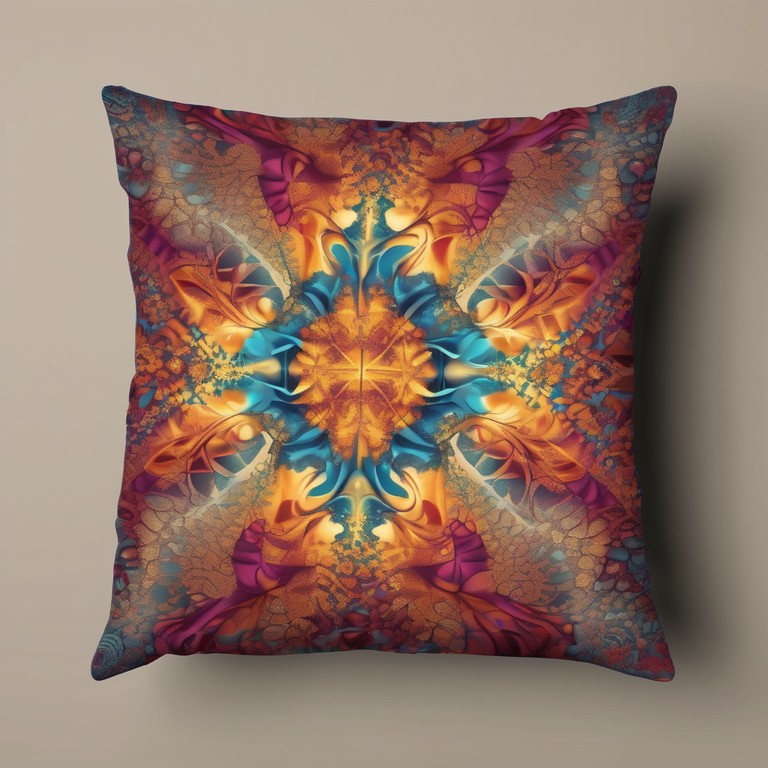} & \includegraphics[width=2cm,height=2cm]{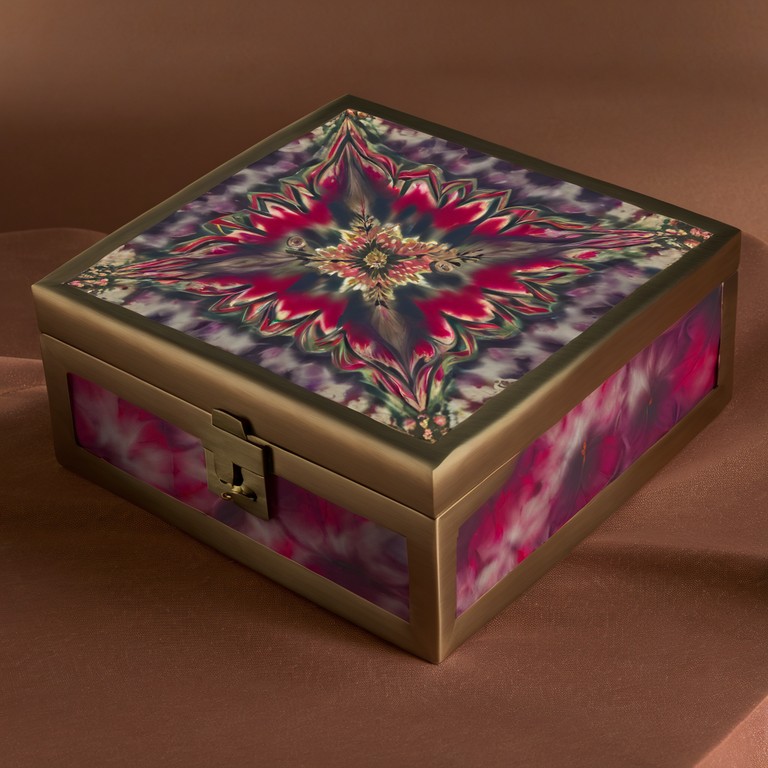} \\
                    \includegraphics[width=2cm,height=2cm]{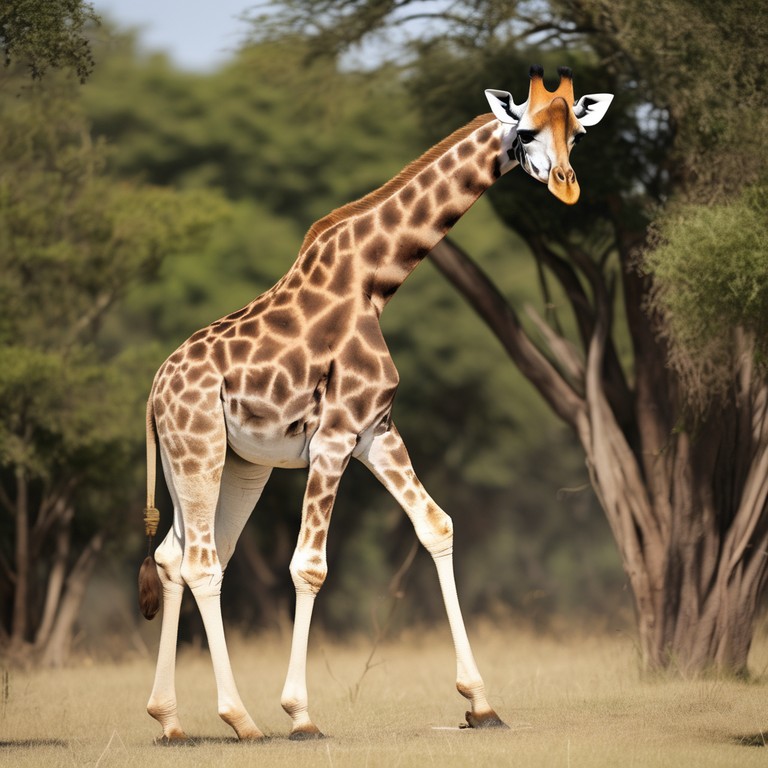} & \includegraphics[width=2cm,height=2cm]{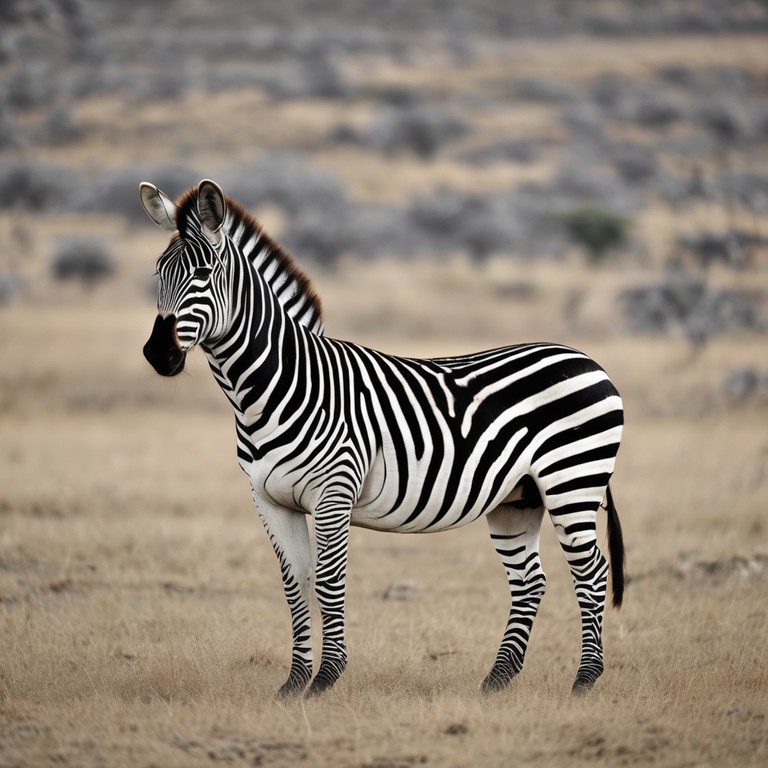} & \includegraphics[width=2cm,height=2cm]{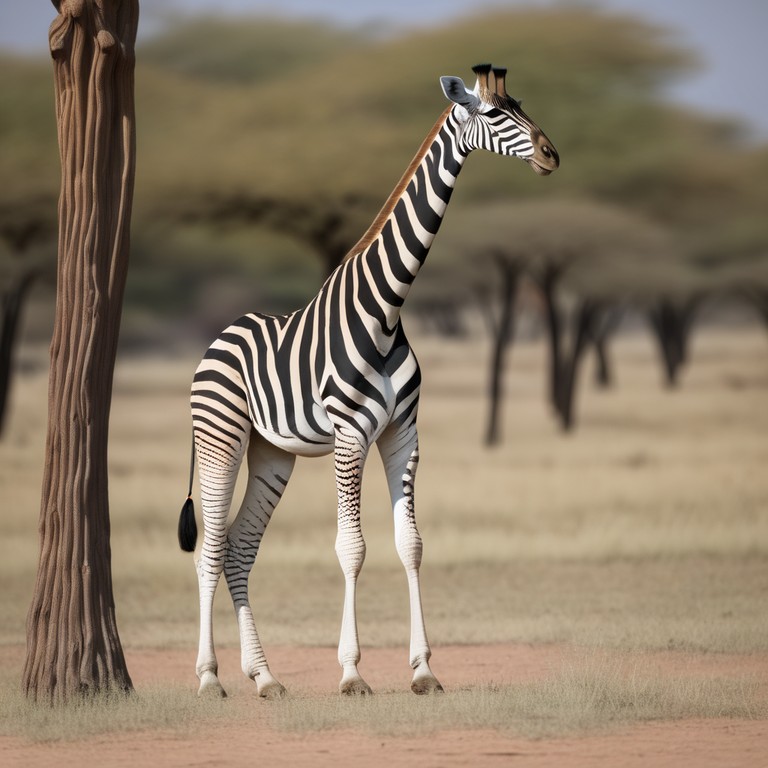} \\
                    \includegraphics[width=2cm,height=2cm]{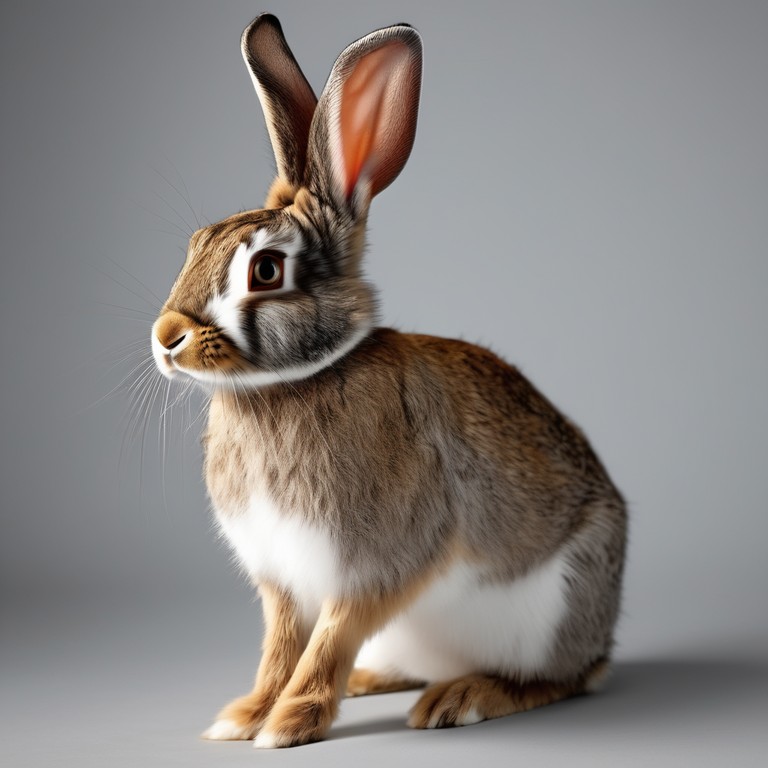} & \includegraphics[width=2cm,height=2cm]{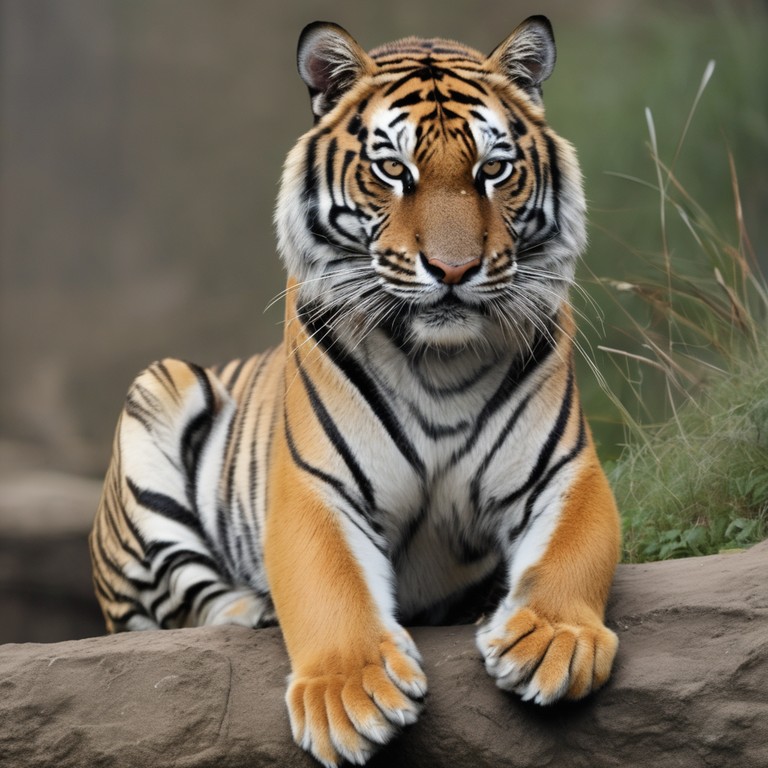} & \includegraphics[width=2cm,height=2cm]{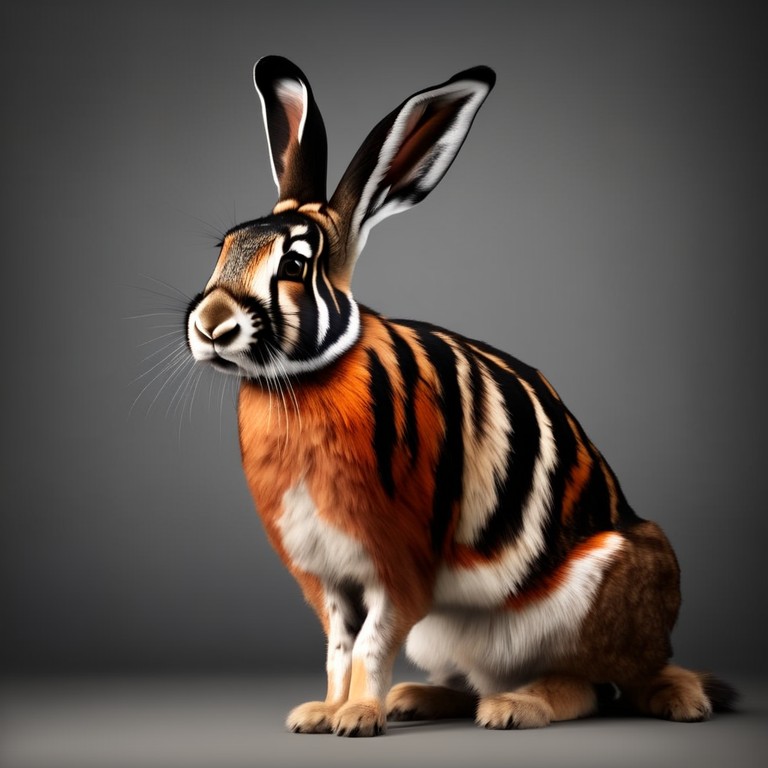} \\
                };

                \node[below=-0.05cm of m2-2-1] {\small ``Object"};
                \node[below=-0.05cm of m2-2-2] {\small ``Pattern"};
                \node[below=-0.05cm of m2-2-3] {\small Result};

                \node[below=-0.05cm of m2-4-1] {\small ``Animal"};
                \node[below=-0.05cm of m2-4-2] {\small ``Fur"};
                \node[below=-0.05cm of m2-4-3] {\small Result};

                \path (m2-1-1) -- (m2-1-2) node[midway, yshift=5.0pt] {\Large $+$};
                \path (m2-1-2) -- (m2-1-3) node[midway, yshift=5.0pt] {\Large $=$};
                \path (m2-2-1) -- (m2-2-2) node[midway, yshift=5.0pt] {\Large $+$};
                \path (m2-2-2) -- (m2-2-3) node[midway, yshift=5.0pt] {\Large $=$};

                \path (m2-3-1) -- (m2-3-2) node[midway, yshift=5.0pt] {\Large $+$};
                \path (m2-3-2) -- (m2-3-3) node[midway, yshift=5.0pt] {\Large $=$};
                \path (m2-4-1) -- (m2-4-2) node[midway, yshift=5.0pt] {\Large $+$};
                \path (m2-4-2) -- (m2-4-3) node[midway, yshift=5.0pt] {\Large $=$};
                
                \matrix (m3) [
                    matrix of nodes,
                    nodes={draw, minimum width=2cm, minimum height=1cm, inner sep=0pt, line width=1.5pt},
                    row sep=0.35cm,
                    column sep=0.3cm,
                    right=0.8cm of m2
                ] {
                    \includegraphics[width=2cm,height=2cm]{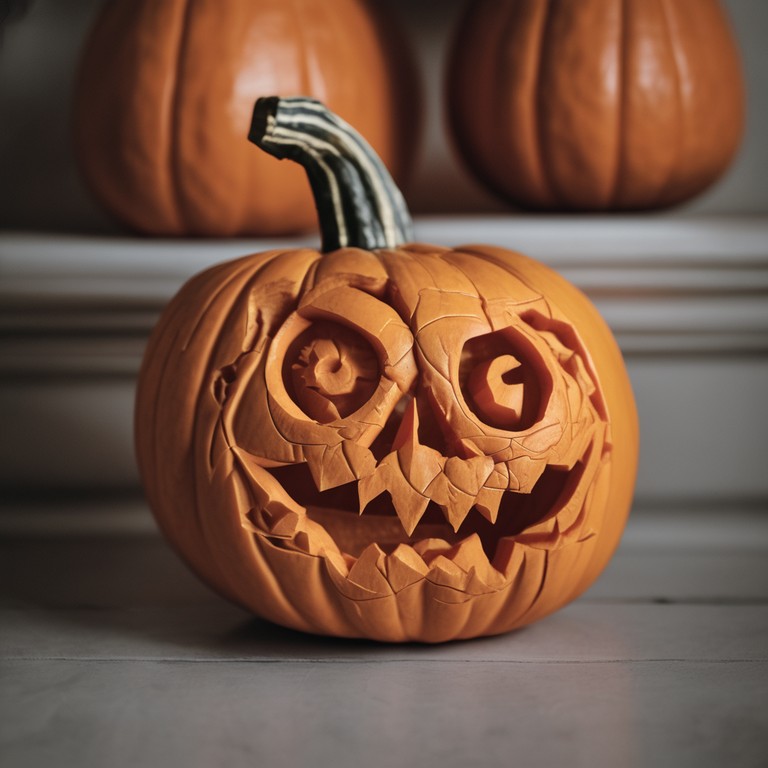} & \includegraphics[width=2cm,height=2cm]{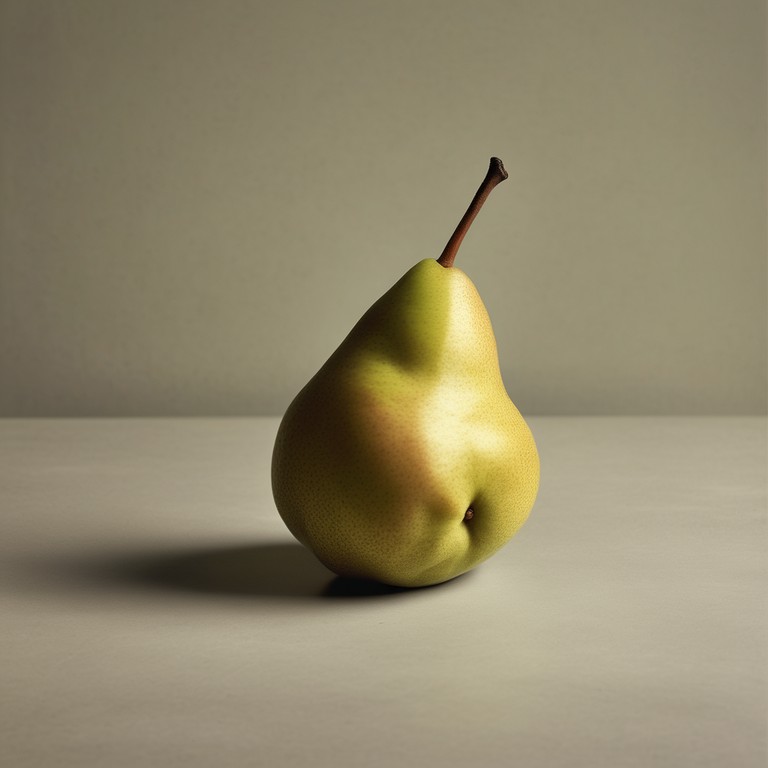} & \includegraphics[width=2cm,height=2cm]{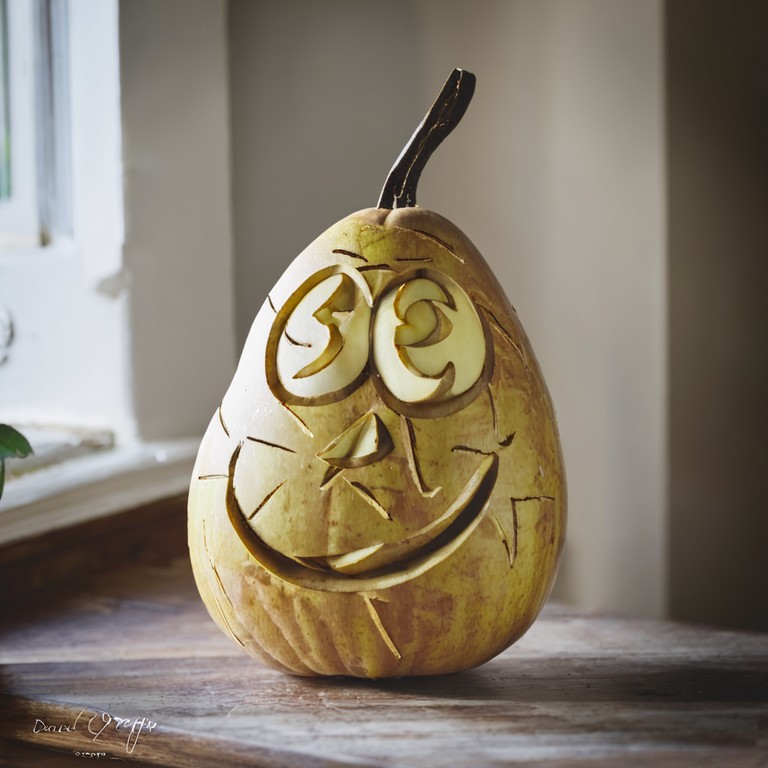} \\
                    \includegraphics[width=2cm,height=2cm]{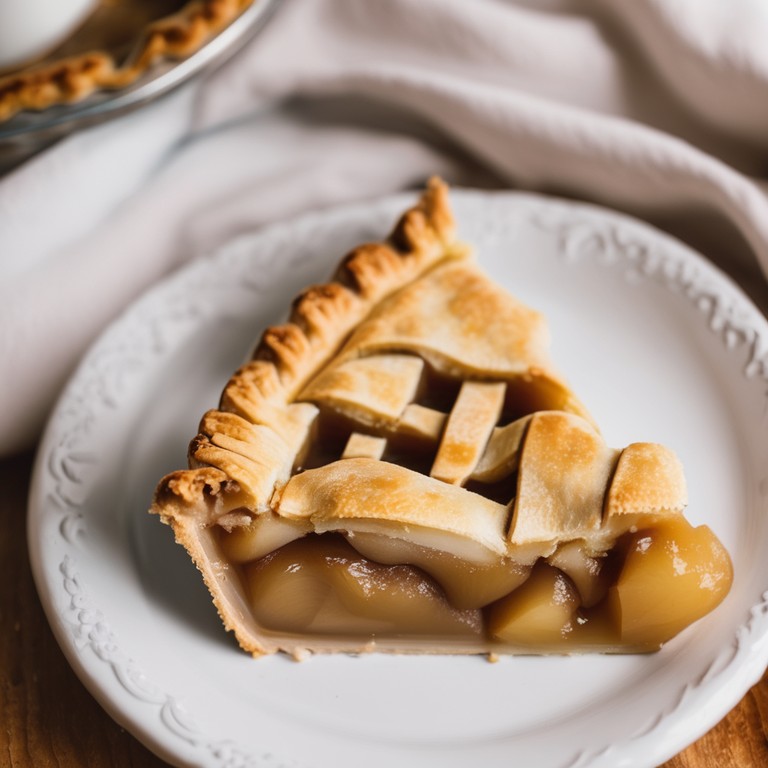} & \includegraphics[width=2cm,height=2cm]{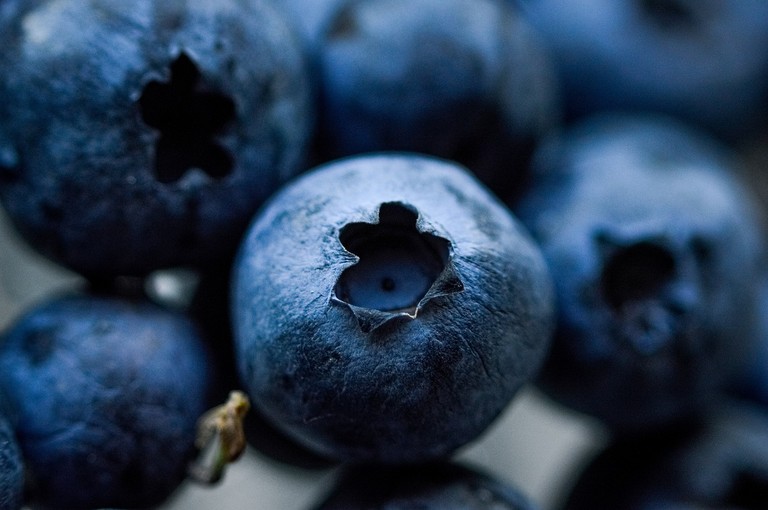} & \includegraphics[width=2cm,height=2cm]{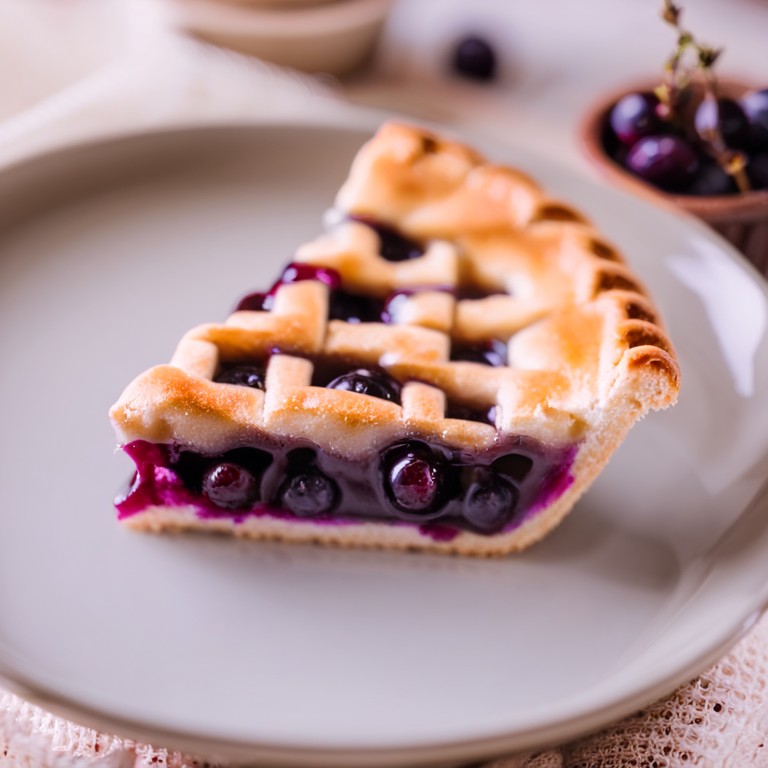} \\
                    \includegraphics[width=2cm,height=2cm]{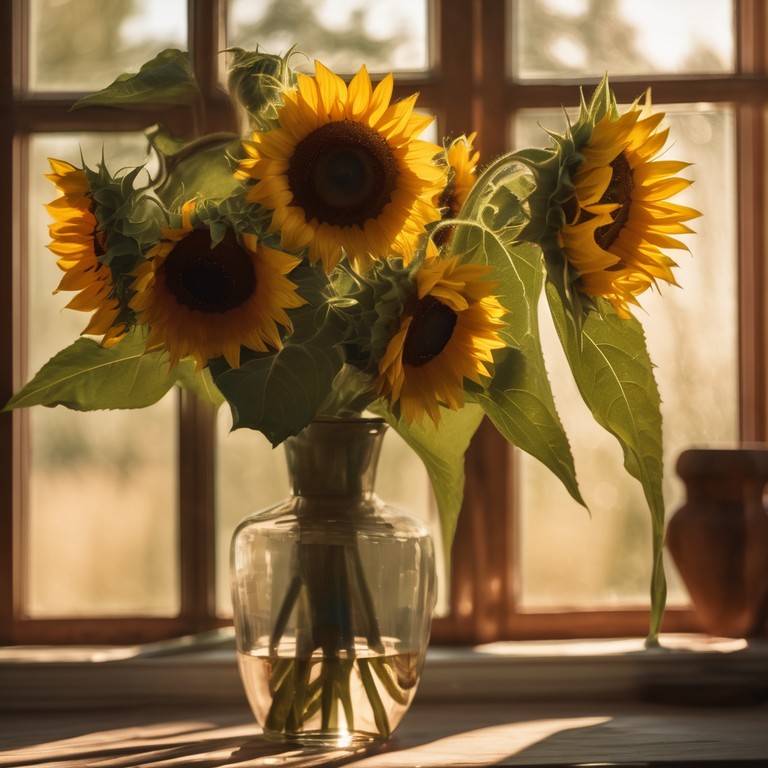} & \includegraphics[width=2cm,height=2cm]{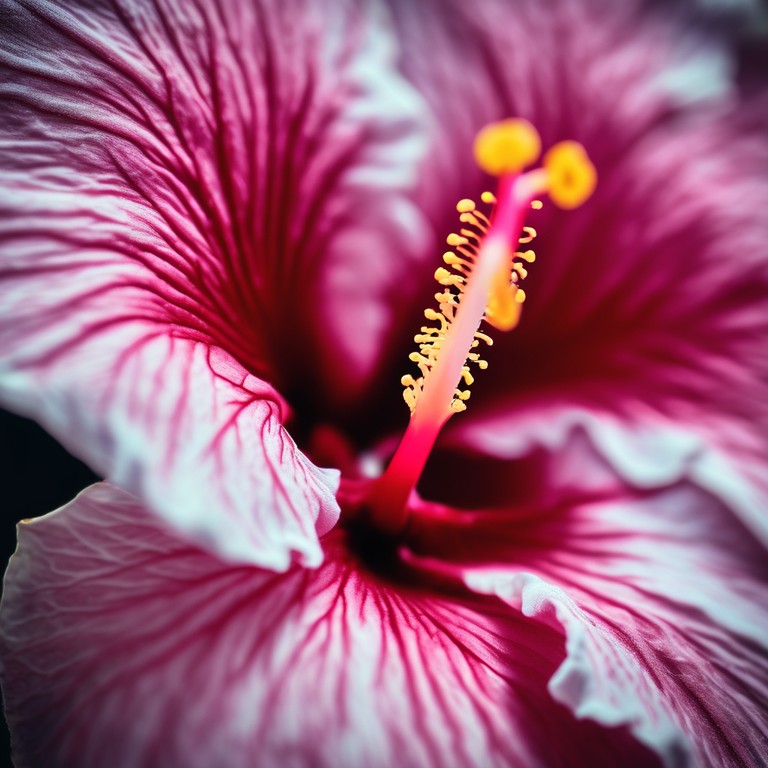} & \includegraphics[width=2cm,height=2cm]{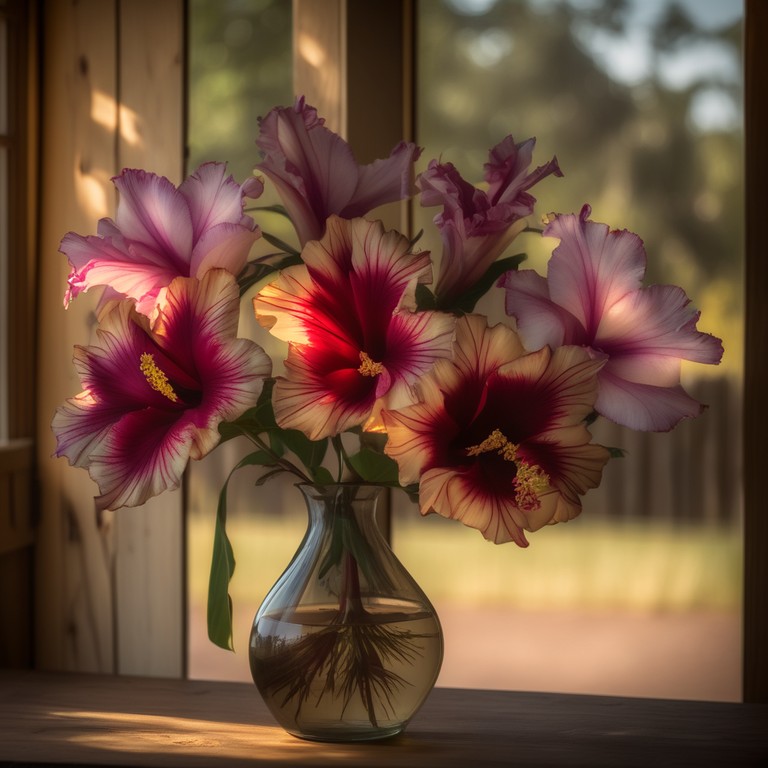} \\
                    \includegraphics[width=2cm,height=2cm]{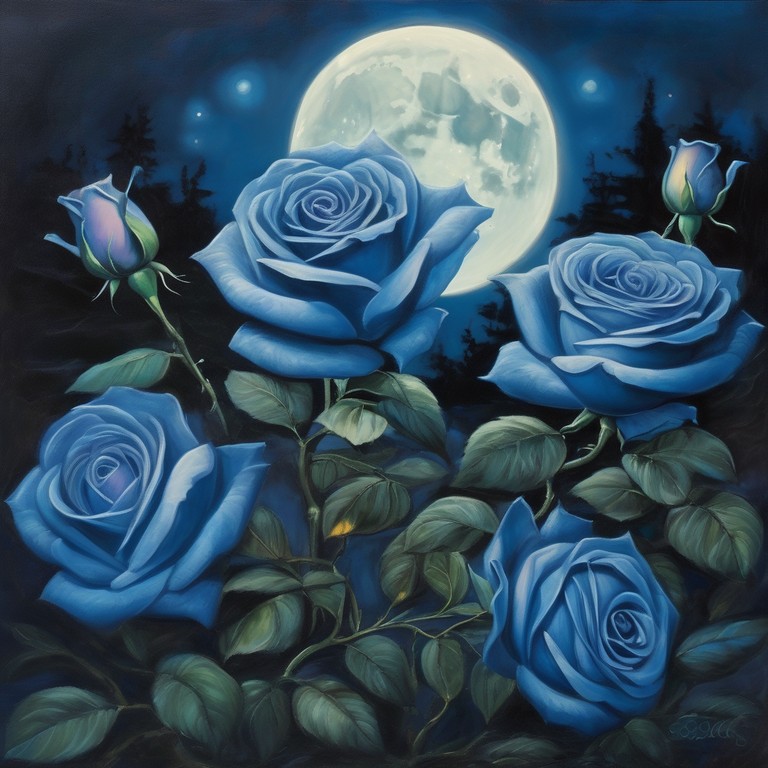} & \includegraphics[width=2cm,height=2cm]{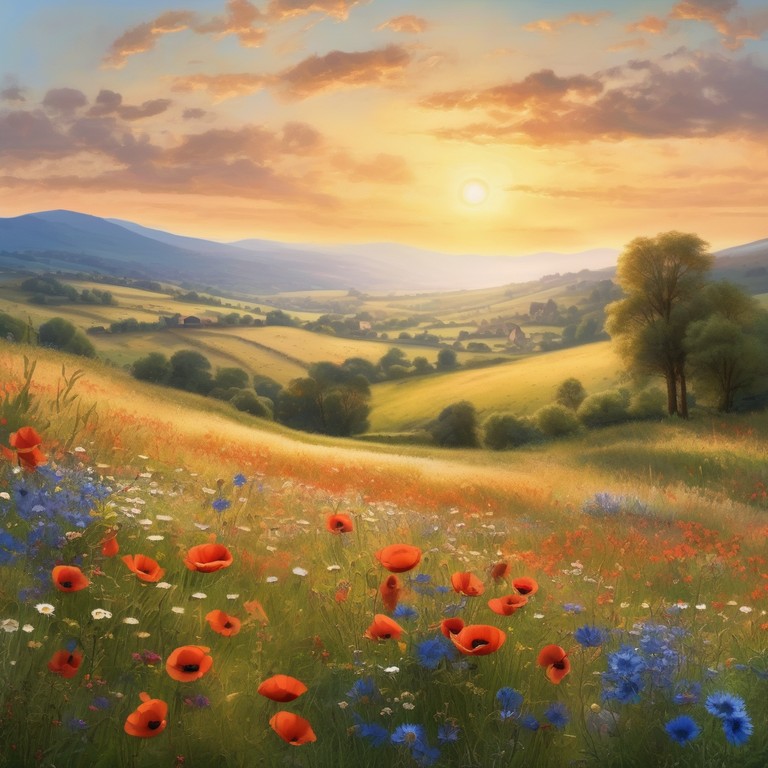} & \includegraphics[width=2cm,height=2cm]{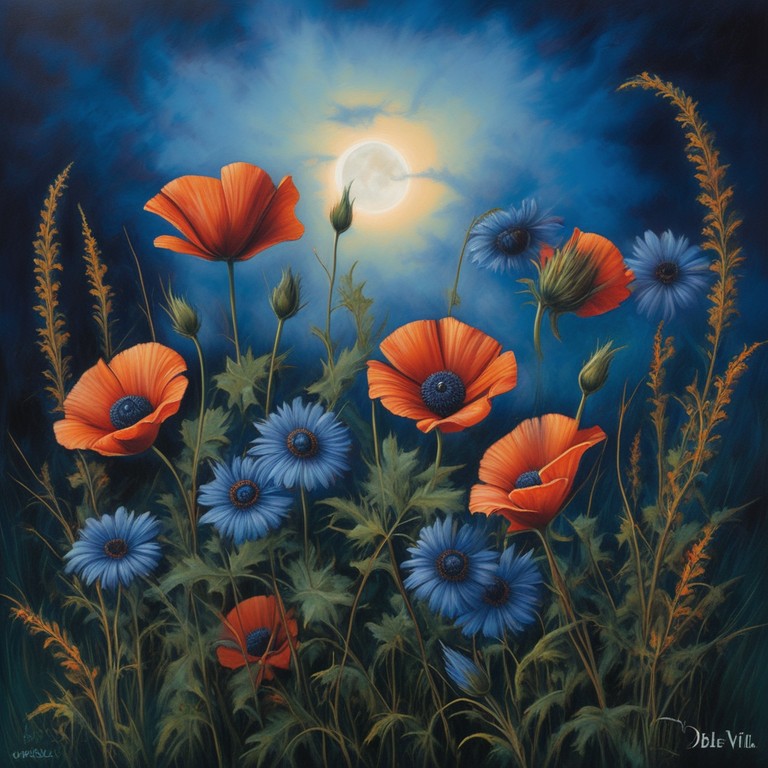} \\
                };

                \node[below=-0.05cm of m3-2-1] {\small ``Scene"};
                \node[below=-0.05cm of m3-2-2] {\small ``Fruit"};
                \node[below=-0.05cm of m3-2-3] {\small Result};

                \node[below=-0.05cm of m3-4-1] {\small ``Scene"};
                \node[below=-0.05cm of m3-4-2] {\small ``Flower"};
                \node[below=-0.05cm of m3-4-3] {\small Result};

                \path (m3-1-1) -- (m3-1-2) node[midway, yshift=5.0pt] {\Large $+$};
                \path (m3-1-2) -- (m3-1-3) node[midway, yshift=5.0pt] {\Large $=$};
                \path (m3-2-1) -- (m3-2-2) node[midway, yshift=5.0pt] {\Large $+$};
                \path (m3-2-2) -- (m3-2-3) node[midway, yshift=5.0pt] {\Large $=$};
                \path (m3-3-1) -- (m3-3-2) node[midway, yshift=5.0pt] {\Large $+$};
                \path (m3-3-2) -- (m3-3-3) node[midway, yshift=5.0pt] {\Large $=$};
                \path (m3-4-1) -- (m3-4-2) node[midway, yshift=5.0pt] {\Large $+$};
                \path (m3-4-2) -- (m3-4-3) node[midway, yshift=5.0pt] {\Large $=$};
                
                \begin{pgfonlayer}{background}
                    \node[draw, rounded corners, inner sep=0.3cm, fit=(m2), line width=1pt] {};
                    \node[draw, rounded corners, inner sep=0.3cm, fit=(m3), line width=1pt] {};
                \end{pgfonlayer}
            \end{tikzpicture}
        };
        
        \begin{pgfonlayer}{background}
            \node[draw, rounded corners, inner sep=0.3cm, fit=(m1), line width=1pt] {};
        \end{pgfonlayer}
    \end{tikzpicture}
    \caption{Additional qualitative results generated using IP-Composer.}
    \label{fig:additional_2}
\end{figure*}